\documentclass[12pt]{article}

\usepackage{geometry}
\usepackage{amsmath}
\usepackage{amssymb}
\usepackage{bbm}

\usepackage{natbib}

\usepackage{booktabs} 

\usepackage{xcolor}

\usepackage[shortlabels]{enumitem}

\usepackage{graphicx}
\graphicspath{ {figures/} }
\usepackage{placeins}

\newtheorem{definition}{Definition}[section]
\newtheorem{lemma}{Lemma}[section]
\newtheorem{proposition}{Proposition}[section]
\newtheorem{theorem}{Theorem}[section]
\newtheorem{remark}{Remark}[section]
\newtheorem{example}{Example}[section]

\newtheorem{assumption}{Assumption}[section]

\makeatletter
\renewcommand\paragraph{%
	\@startsection{paragraph}
	{4}
	{\z@}
	{3.25ex \@plus1ex \@minus.2ex}
	{-1em}
	{\normalfont\normalsize\bfseries\maybe@addperiod}%
}
\newcommand{\maybe@addperiod}[1]{%
	#1\@addpunct{.}%
}
\makeatother

\usepackage[pagebackref]{hyperref}       
\hypersetup{
	colorlinks=true,
	linkcolor=blue,
	filecolor=blue,
	citecolor=blue,      
	urlcolor=blue,
}
\renewcommand*{\backref}[1]{}
\renewcommand*{\backrefalt}[4]{%
	\ifcase #1 Not cited.%
	\or        Cited on page~#2.%
	\else      Cited on pages~#2.%
	\fi}

\usepackage{cleveref}
\crefname{section}{Section}{Sections}
\crefname{figure}{Figure}{Figures}
\crefname{definition}{Definition}{Definitions}
\crefname{lemma}{Lemma}{Lemmas}
\crefname{proposition}{Proposition}{Propositions}
\crefname{theorem}{Theorem}{Theorems}
\crefname{remark}{Remark}{Remarks}
\crefname{corollary}{Corollary}{Corollaries}
\crefname{appendix}{Appendix}{Appendices}
\crefname{assumption}{Assumption}{Assumptions}
\crefname{example}{Example}{Examples}
\crefname{table}{Table}{Tables}

\usepackage[nottoc,numbib]{tocbibind}

\usepackage[parfill]{parskip}

\usepackage[font=footnotesize,labelfont=bf, textfont=it]{caption}

\usepackage{crossreftools}
\DeclareMathOperator*{\argmax}{arg\,max}
\DeclareMathOperator*{\argmin}{arg\,min}
\newcommand{\precprec}{\prec\mathrel{\mkern-5mu}\prec}
\newcommand{\succsucc}{\succ\mathrel{\mkern-5mu}\succ}

\title{Random Pareto front surfaces}
\author{
	Ben Tu\thanks{Imperial College London, United Kingdom}
	\and Nikolas Kantas\footnotemark[1]
	\and Robert M. Lee\thanks{BASF SE, Germany}
	\and Behrang Shafei\footnotemark[2]
}
\date{}

\begin{document}
\maketitle
\begin{abstract}
	The goal of multi-objective optimisation is to identify the Pareto front surface which is the set obtained by connecting the best trade-off points. Typically this surface is computed by evaluating the objectives at different points and then interpolating between the subset of the best evaluated trade-off points. In this work, we propose to parameterise the Pareto front surface using polar coordinates. More precisely, we show that any Pareto front surface can be equivalently represented using a scalar-valued length function which returns the projected length along any positive radial direction. We then use this representation in order to rigorously develop the theory and applications of stochastic Pareto front surfaces. In particular, we derive many Pareto front surface statistics of interest such as the expectation, covariance and quantiles. We then discuss how these can be used in practice within a design of experiments setting, where the goal is to both infer and use the Pareto front surface distribution in order to make effective decisions. Our framework allows for clear uncertainty quantification and we also develop advanced visualisation techniques for this purpose. Finally we discuss the applicability of our ideas within multivariate extreme value theory and illustrate our methodology in a variety of numerical examples, including a case study with a real-world air pollution data set.
\end{abstract}
\section{Introduction}
The Pareto front is often regarded as the natural generalisation of the maximum (or minimum) for vector-valued sets. This generalisation is based on the Pareto partial ordering relation, which gives us a way to compare between any two vectors when possible. Geometrically, the Pareto front can be used to define a surface in the vector space which describes the best possible trade-offs one can obtain between the different components. Capturing this surface is crucial for many real-world applications and machine learning tasks where one is interested in jointly maximising multiple reward criteria. Notably, this problem becomes much more challenging when there is uncertainty in the rewards, which is very common in practice. The purpose of this work is to address this practically important problem directly by providing the tools and methodology required to study and effectively use Pareto front surfaces in the face of uncertainty. To the best of our knowledge, a study of this kind is currently missing in the literature and would be very useful for real-world practitioners who deal with these problems in practice.

At a high level, our novel framework is based on specific polar representation of the Pareto front surface.  More precisely, we show that any Pareto front surface can be equivalently characterised by its projected length function, which is a scalar-valued function that computes the projected lengths of the polar surface along any positive radial direction. The clear benefit of this representation is that it is explicit and gives us an interpretable and efficient way to work over the space of Pareto front surfaces. To demonstrate the value of our framework, we focus our attention on the stochastic setting where the Pareto front surface itself is unknown and random. Under this regime, our polar parameterisation result allows us to treat any random Pareto front surface as a tractable infinite-dimensional random variable. Various statistics such as its expectation, covariance and quantiles are then well defined and straightforward to compute numerically. Importantly, these statistics are especially useful in the context of quantifying uncertainty and making better decisions in the stochastic setting. Overall, we believe that our methodology is very practical as it can be easily integrated into most standard decision making workflows and there are many potential directions to extend our work even further.

\subsection{Structure and contributions} 
The remainder of the paper is organised as follows: In \cref{sec:preliminaries}, we introduce the key definitions and nomenclature that will be used throughout the paper. In \cref{sec:polar_parameterisation}, we present our framework and show that any Pareto front surface can be expressed using polar coordinates. We then use this polar parameterisation in order to define some useful operations and concepts on the space of Pareto front surfaces. Namely we define the concept of an order-preserving transformation (\cref{sec:algebra}), a length-based utility function (\cref{sec:utility}) and a length-based loss function (\cref{sec:loss}). We note that whilst polar representations have been used before as a technical tool for manipulating calculations related to hypervolume indicator \citep{shang2018pgecc,deng2019itec,zhang2020icml}, this is the first time to the best of our knowledge that they have been studied in detail and proposed as a methodological tool.

In \cref{sec:statistics}, we study the setting where the Pareto front surface is random and define many useful statistics of interest such as the expectation (\cref{sec:expectation}), covariance (\cref{sec:covariance}) and quantiles (\cref{sec:quantiles}), among others. We also discuss links with existing works on Pareto front surface distributions based on ideas from random set theory (\cref{sec:vorobev}). We then proceed to \cref{sec:applications}, where we show how these ideas can be used in practice:
\begin{itemize}
	\item In \cref{sec:visualisation}, we present a novel visualisation strategy based our polar parameterisation and we show how it is possible to construct a picture of the whole Pareto front surface by using a family of low-dimensional slices. This can assist in situations where one is interested in actively identifying decisions or inputs which lead to the best trade-offs for the decision maker.
	\item In \cref{sec:uncertainty_quantification}, we illustrate how all of these uncertainty quantification and visualisation ideas can be used within a standard Bayesian experimental design setting. Specifically, we demonstrate how these tools can be used both during an optimisation routine  (\cref{sec:bayesian_optimisation}) and also in the post-decision setting (\cref{sec:input_decision}). The former stage focusses on the problem of identifying the best experiments to run, whilst the latter stage is focussed on identifying the inputs which are most likely going lead to the desired outputs. 
	\item In \cref{sec:extreme_value_theory}, we demonstrate how it is possible to adapt some prominent results from extreme value theory to work in the multivariate setting, where the maximum is defined using the Pareto partial ordering. This is in contrast to most existing work in multivariate extreme value theory, which has largely focussed on the setting where the vector-valued maximum is defined in a component-wise fashion. 
	\item In \cref{sec:air_pollution}, we present a case study on how some of these Pareto front ideas can be used in order to quantify changes in the daily maximum air pollutant levels in a part of west London. 
\end{itemize}

In \cref{sec:conclusion}, we conclude this work and include a discussion on future research directions. Finally, in \cref{app:proofs}, we include the proofs of all of the results that are stated within the paper. The code that is needed to reproduce the figures and numerical experiments are available in our Github repository: \href{https://github.com/benmltu/scalarize}{\texttt{https://github.com/benmltu/scalarize}}.

\section{Preliminaries}
\label{sec:preliminaries}
In this work, we study the properties of Pareto front surfaces, which are the surfaces obtained by interpolating the Pareto optimum of a vector-valued set. Without loss of generality, we assume throughout that the goal of interest is maximisation. Naturally, the minimisation problem can also be treated by simply negating the corresponding set of vectors. Note that there are many subtle differences in the literature when it comes to defining a Pareto front surface. For this reason, we will now carefully define the concept of the truncated interpolated weak Pareto front, which will be the primary focus of this work.
\begin{definition}
	[Pareto domination] The weak, strict and strong Pareto domination is denoted by the binary relations $\succeq, \succ$ and $\succsucc$, respectively. We say a vector $\mathbf{y} \in \mathbb{R}^M$ weakly, strictly or strongly Pareto dominates another vector $\mathbf{y'} \in \mathbb{R}^M$ if
	\begin{align*}
		\mathbf{y} \succeq \mathbf{y'} &\iff \mathbf{y} - \mathbf{y'} \in \mathbb{R}_{\geq 0}^M,
		\\
		\mathbf{y} \succ \mathbf{y'} &\iff \mathbf{y} - \mathbf{y'} \in \mathbb{R}_{\geq 0}^M \setminus \{\mathbf{0}_M\},
		\\
		\mathbf{y} \succsucc \mathbf{y'} &\iff \mathbf{y} - \mathbf{y'} \in \mathbb{R}_{> 0}^M,
	\end{align*}
	respectively, where $\mathbf{0}_M \in \mathbb{R}^M$ denotes the $M$-dimensional vector of zeros.
\end{definition}
\begin{definition}
	[Domination region] The Pareto domination region is defined as the collection of points which dominates (or is dominated) by a particular set of vectors $A \subset \mathbb{R}^M$, that is
	\begin{equation*}
		\mathbb{D}_{\diamond}[A] 
		:= \cup_{\mathbf{a} \in A} \{\mathbf{y} \in \mathbb{R}^M: \mathbf{y} \diamond \mathbf{a}\},
	\end{equation*}
	where $\diamond \in \{\preceq, \succeq, \prec, \succ, \precprec, \succsucc\}$ denotes a partial ordering relation. In addition, we denote the truncated domination region by $\mathbb{D}_{\diamond, \boldsymbol{\eta}}[A] := \mathbb{D}_{\diamond}[A] \cap \mathbb{D}_{\succsucc}[\{\boldsymbol{\eta}\}]$, for any reference vector $\boldsymbol{\eta} \in \mathbb{R}^M$ and denote the complement of any domination region by $\mathbb{D}^{C}_{\diamond}[A] := \mathbb{R}^M \setminus \mathbb{D}_{\diamond}[A]$.
\end{definition}
\begin{definition}
	[Pareto optimality] Given a bounded set of vectors $A \subset \mathbb{R}^M$, a point $\mathbf{a} \in A$ is weakly or strictly Pareto optimal if there does not exist another vector $\mathbf{a}' \in A$ which strongly or strictly Pareto dominates it, respectively. The collection of all weakly or strictly Pareto optimal vectors in this set is called the weak or strict Pareto front, $\mathcal{Y}^{\textnormal{weak}}[A] := A \cap \mathbb{D}^{C}_{\precprec}[A]$ and $\mathcal{Y}^{\textnormal{strict}}[A] := A \cap \mathbb{D}^{C}_{\prec}[A]$, respectively.
\end{definition}
\begin{figure}
	\includegraphics[width=1\linewidth]{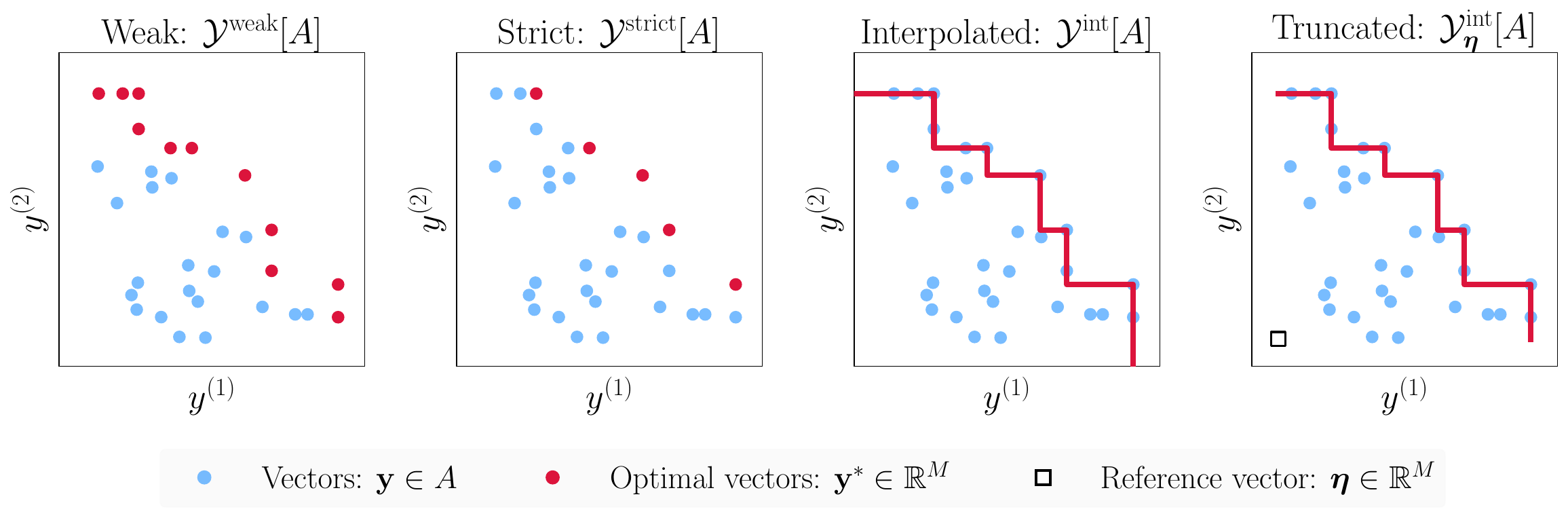}
	\centering
	\caption{An illustration of the different Pareto fronts in $M=2$ dimensions.}
	\label{fig:pareto_front_variants}
\end{figure}
Although strict Pareto optimality is a desirable quality, it is a difficult operation to work with because it returns a subset of the original set, which could contain an arbitrary number of points. To address this problem, we consider working with a more relaxed notion of Pareto optimality, which we refer to as the interpolated weak Pareto optimality (\cref{def:pareto_front_surface}). Conceptually, we define the interpolated weak Pareto front as the interpolation of the strict Pareto front. This interpolation defines a surface in the $M$-dimensional vector space. For practical convenience, we also define the truncated interpolated weak Pareto front, which truncates\footnote{Crudely speaking, we can recover the interpolated weak Pareto front from the truncated interpolated weak Pareto front by sending the reference point to minus infinity: $\boldsymbol{\eta} = (\epsilon, \dots, \epsilon) \in \mathbb{R}^M$ and $\epsilon \rightarrow - \infty$.} this surface at some reference point $\boldsymbol{\eta} \in \mathbb{R}^M$---see \cref{fig:pareto_front_variants} for an illustration.
\begin{definition}
	[Interpolated weak optimality] Given a bounded set of vectors $A \subset \mathbb{R}^M$, we define the interpolated weak Pareto front as the weak Pareto front of its weak domination region after closure, that is: $\mathcal{Y}^{\textnormal{int}}[A] := \mathcal{Y}^{\textnormal{weak}}[\mathbb{D}_{\preceq}[\textsc{Closure}(A)]]$. By truncating this surface at some reference vector $\boldsymbol{\eta} \in \mathbb{R}^M$, we obtain the truncated interpolated weak Pareto front $\mathcal{Y}_{\boldsymbol{\eta}}^{\textnormal{int}}[A] := \mathcal{Y}^{\textnormal{int}}[A] \cap \mathbb{D}_{\succsucc}[\{\boldsymbol{\eta}\}]$. We denote the set of all possible non-empty truncated interpolated weak Pareto fronts by $\mathbb{Y}^*_{\boldsymbol{\eta}} \subset 2^{\mathbb{R}^M}$.
	\label{def:pareto_front_surface}
\end{definition}
In practice, the set of objective vectors that we seek to optimise is often the output of some vector-valued objective function $g: \mathbb{X} \rightarrow \mathbb{R}^M$, where $\mathbb{X}$ denotes the space of feasible inputs. In this setting, the optimal set of objectives is given by the Pareto front of the feasible objective vectors. For example, the strict Pareto front is given by the set $\mathcal{Y}^{\textnormal{strict}}[\{g(\mathbf{x})\}_{\mathbf{x} \in \mathbb{X}}] \subset \mathbb{R}^M$, whilst the strict Pareto set is defined as the corresponding pre-image. Similarly, we can also define the other weak variants of the Pareto front and Pareto set as well.
\begin{remark}
	[Reference vector] Throughout this work, we will assume that the reference vector $\boldsymbol{\eta} \in \mathbb{R}^M$ is known and fixed by the decision maker. Intuitively, this parameter is just used as a way to lower bound the set of vectors that we are interested in targetting. In the multi-objective literature, it is common to see this vector set to an estimate that is equal or close the nadir point, which is the vector comprised of the worst possible values for each objective: $\eta^{(m)} = \min_{\mathbf{x} \in \mathbb{X}} g^{(m)}(\mathbf{x})$ for objectives $m=1,\dots,M$. In fact, some authors have suggested using a reference point that is slightly worse than the nadir when we are interested in evaluating the performance of a multi-objective algorithm \citep{ishibuchi2017pgecc}.
	\label{rem:reference_vector}
\end{remark}
\paragraph{Nomenclature} For convenience, unless otherwise stated, we will from now on refer to a set as being a Pareto front or Pareto front surface if it is a truncated interpolated weak Pareto front (\cref{def:pareto_front_surface}). We will refer to a point as being Pareto optimal if it lies on this truncated interpolated weak Pareto front. In addition, we will refer to the set of all non-empty truncated interpolated weak Pareto front $\mathbb{Y}^*_{\boldsymbol{\eta}}$ as the set of all Pareto fronts or Pareto front surfaces.
\section{Polar parameterisation}
\label{sec:polar_parameterisation}
\begin{figure}
	\includegraphics[width=1\linewidth]{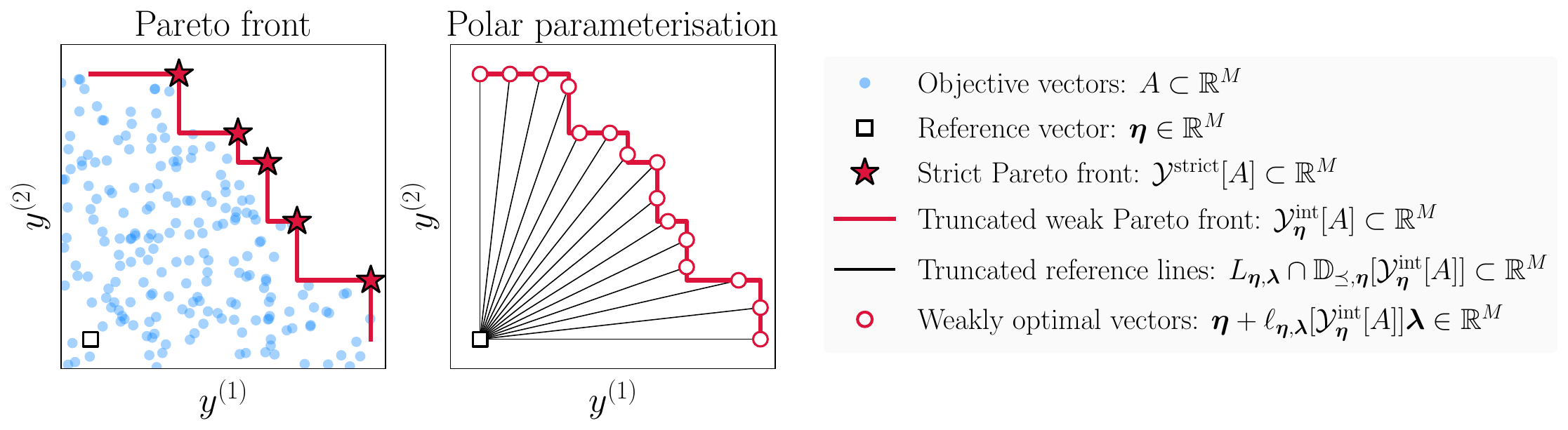}
	\centering
	\caption{An illustration of the polar parameterisation result in $M=2$ dimensions.}
	\label{fig:polar_parameterisation}
\end{figure}

We now present the main result of this paper (\cref{thm:polar_parameterisation}), which states that all non-empty Pareto front surfaces are isomorphic to the set of positive unit vectors
\begin{equation*}
	\mathcal{S}^{M-1}_+: = \{\mathbf{z} \in \mathbb{R}_{> 0} ^M: ||\mathbf{z}||_{L^2} = 1 \} \in \mathbb{Y}^*_{\mathbf{0}_M}.
\end{equation*}
Notably we present an explicit representation for this isomorphism in \eqref{eqn:polar_parameterisation}, which we refer to as the polar parameterisation of a Pareto front surface. The name of this representation is motivated by the fact that we rely on the hyperspherical polar coordinates transformation in order to derive this result. The intuitive idea behind this result is concisely described in \cref{fig:polar_parameterisation}. Informally speaking, we can identify each Pareto optimal point by drawing a line from the reference vector $\boldsymbol{\eta} \in \mathbb{R}^M$ along a positive direction $\boldsymbol{\lambda} \in \mathcal{S}^{M-1}_+$. If we know the length of these lines $\ell_{\boldsymbol{\eta}, \boldsymbol{\lambda}} \in \mathbb{R}$ when it intersects the Pareto front surface, then we can reconstruct the Pareto front surface by using a linear mapping: $\boldsymbol{\eta} + \ell_{\boldsymbol{\eta}, \boldsymbol{\lambda}} \boldsymbol{\lambda} \in \mathbb{R}^M$ for $\boldsymbol{\lambda} \in \mathcal{S}^{M-1}_+$. In the following paragraphs, we will formalise this idea more concretely and give an explicit construction for this length function.

\begin{figure}
	\includegraphics[width=0.8\linewidth]{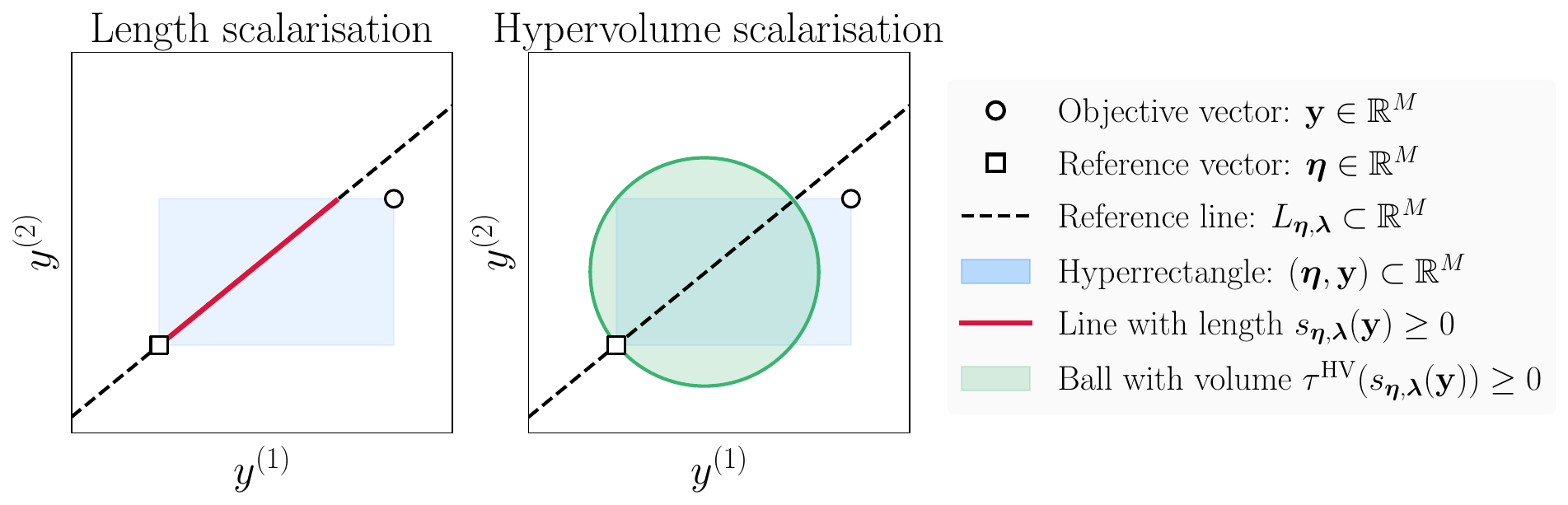}
	\centering
	\caption{An illustration of the length-based scalarisation functions in $M=2$ dimensions.}
	\label{fig:length_based_scalarisations}
\end{figure}

\paragraph{Polar coordinates} To derive our polar parameterisation result, we rely on the following coordinate system transformation $\mathcal{T}_{\boldsymbol{\eta}}: \mathbb{D}_{\succsucc}[\{\boldsymbol{\eta}\}] \rightarrow \mathcal{S}^{M-1}_+ \times \mathbb{R}_{> 0}$,
\begin{align}
	\begin{split}
		\mathcal{T}_{\boldsymbol{\eta}}(\mathbf{y}) 
		&:= (\boldsymbol{\lambda}_{\boldsymbol{\eta}}^*(\mathbf{y}), s_{\boldsymbol{\eta}, \boldsymbol{\lambda}_{\boldsymbol{\eta}}^*(\mathbf{y})}(\mathbf{y}))
		= \biggl(\frac{\mathbf{y} - \boldsymbol{\eta}}{||\mathbf{y} - \boldsymbol{\eta}||_{L^2}}, 
		||\mathbf{y} - \boldsymbol{\eta}||_{L^2} \biggr),
	\end{split}
	\label{eqn:coordinate_transformation}
\end{align}
which maps any vector $\mathbf{y} \in \mathbb{D}_{\succsucc}[\{\boldsymbol{\eta}\}]$ to a positive unit vector $\boldsymbol{\lambda}_{\boldsymbol{\eta}}^*(\mathbf{y}) \in \mathcal{S}^{M-1}_+$ and a positive scalar $s_{\boldsymbol{\eta}, \boldsymbol{\lambda}_{\boldsymbol{\eta}}^*(\mathbf{y})}(\mathbf{y}) > 0$. Intuitively, this mapping can be interpreted as a variant of the hyperspherical polar coordinates transformation centred at the reference vector $\boldsymbol{\eta} \in \mathbb{R}^M$. In our setting, the positive unit vector plays the role of the angle, whilst the positive scalar plays the role of the projected length. The key difference between this transformation and the standard hyperspherical polar coordinates transformation is that here we restrict our attention to just the angles lying in the positive orthant. These positive directions are the only ones which will lead to a Pareto optimal point. Formally, the coordinate transformation in \eqref{eqn:coordinate_transformation} relies on the length\footnote{This scalarisation function has also been referred to in the literature as an achievement scalarisation function \citep{ishibuchi20092icec,deb20122icec}.} scalarisation function $s_{\boldsymbol{\eta}, \boldsymbol{\lambda}}: \mathbb{R}^M \rightarrow \mathbb{R}$,
\begin{equation}
	s_{\boldsymbol{\eta}, \boldsymbol{\lambda}}(\mathbf{y}) 
	:= \textsc{Length}[L_{\boldsymbol{\eta}, \boldsymbol{\lambda}} \cap (\boldsymbol{\eta}, \mathbf{y})]
	= \min_{m=1,\dots,M} \frac{\max(y^{(m)} - \eta^{(m)}, 0)}{\lambda^{(m)}},
	\label{eqn:length_scalarisation}
\end{equation}
which is a non-negative\footnote{The length scalarisation function is positive on the truncated space $\mathbb{D}_{\succsucc}[\{\boldsymbol{\eta}\}]$ and zero everywhere else.} function that is defined for all vectors $\mathbf{y} \in \mathbb{R}^M$. Conceptually, when the objective vector lies in the truncated space $\mathbf{y} \in \mathbb{D}_{\succsucc}[\{\boldsymbol{\eta}\}]$, then the length scalarisation function computes the length of the line $L_{\boldsymbol{\eta}, \boldsymbol{\lambda}} := \{\boldsymbol{\eta} + t \boldsymbol{\lambda}: t \in \mathbb{R}\}$ lying in the open hyper-rectangle $(\boldsymbol{\eta}, \mathbf{y}) \subset \mathbb{R}^M$---see the left of \cref{fig:length_based_scalarisations} for an illustration of this intuition in two dimensions. Consequently, the optimal direction function $\boldsymbol{\lambda}_{\boldsymbol{\eta}}^*: \mathbb{D}_{\succsucc}[\{\boldsymbol{\eta}\}] \rightarrow \mathcal{S}^{M-1}_+$  is the function that returns the largest projected length
\begin{equation}
	\boldsymbol{\lambda}_{\boldsymbol{\eta}}^*(\mathbf{y}) 
	:= \frac{\mathbf{y} - \boldsymbol{\eta}}{||\mathbf{y} - \boldsymbol{\eta}||_{L^2}}
	\in \argmax_{\boldsymbol{\lambda} \in \mathcal{S}_+^{M-1}} s_{\boldsymbol{\eta}, \boldsymbol{\lambda}}(\mathbf{y})
	\label{eqn:optimal_weight}
\end{equation}
for any vector $\mathbf{y} \in \mathbb{D}_{\succsucc}[\{\boldsymbol{\eta}\}]$ lying in the truncated space. Note that to invert this coordinate transformation we can simply apply the linear transformation $\mathcal{T}_{\boldsymbol{\eta}}^{-1}: \mathcal{S}^{M-1}_+ \times \mathbb{R}_{> 0} \rightarrow \mathbb{D}_{\succsucc}[\{\boldsymbol{\eta}\}] $, where $\mathcal{T}_{\boldsymbol{\eta}}^{-1}((\boldsymbol{\lambda}, l)) := \boldsymbol{\eta} + l \boldsymbol{\lambda}$ for any $(\boldsymbol{\lambda}, l) \in \mathcal{S}^{M-1}_+ \times \mathbb{R}_{> 0}$. 

\paragraph{Polar surfaces} By construction, each reference line $L_{\boldsymbol{\eta}, \boldsymbol{\lambda}}$ is pointing in an increasing direction. This implies that each reference line should intersect the Pareto front surface in exactly one point if it is non-empty. If this is not the case, then we get a contradiction to Pareto optimality. Using this observation, we define the set of polar surfaces $\mathbb{L}_{\boldsymbol{\eta}} \subset 2^{\mathbb{R}^M}$, in \cref{def:polar_surfaces}, to be the set of vectors lying in the truncated space\footnote{Note that we have also included the set $\{\boldsymbol{\eta}\} \in \mathbb{L}_{\boldsymbol{\eta}}$ in \cref{def:polar_surfaces}. This subtle inclusion is made in order to accommodate for the degenerate setting---see \cref{rem:singleton}.} which satisfies this intersection constraint.
\begin{definition}
	[Polar surfaces] For any reference vector $\boldsymbol{\eta} \in \mathbb{R}^M$, a set $A \subset \mathbb{R}^M$ is called a polar surface, $A \in \mathbb{L}_{\boldsymbol{\eta}}$, if and only if the following statement holds:	
	\begin{equation*}
		A \subset \mathbb{D}_{\succsucc}[\{\boldsymbol{\eta}\}] \cup \{\boldsymbol{\eta} \} \enskip \text{ and } \enskip |A \cap L_{\boldsymbol{\eta}, \boldsymbol{\lambda}}| = 1 \text{ for all } \boldsymbol{\lambda} \in \mathcal{S}^{M-1}_+,
	\end{equation*}
	where $|\cdot|$ denotes the cardinality of a set. Moreover, for this set of polar surfaces $\mathbb{L}_{\boldsymbol{\eta}}$, we define the projected length function $\ell_{\boldsymbol{\eta}, \boldsymbol{\lambda}}: \mathbb{L}_{\boldsymbol{\eta}} \rightarrow \mathbb{R}_{\geq 0}$ as the function which returns the $L^2$-distance of the intersected point along any positive direction $\boldsymbol{\lambda} \in \mathcal{S}^{M-1}_+$:
	\begin{equation*}
		\ell_{\boldsymbol{\eta}, \boldsymbol{\lambda}}[A] = \max_{\mathbf{a} \in A \cap L_{\boldsymbol{\eta}, \boldsymbol{\lambda}}} ||\mathbf{a}||_{L^2}
	\end{equation*}
	for any polar surface $A \in \mathbb{L}_{\boldsymbol{\eta}}$.
	\label{def:polar_surfaces}
\end{definition}
Note that the projected lengths provide a dual representation of a polar surface, that is 
\begin{equation}
	A \in \mathbb{L}_{\boldsymbol{\eta}} \iff A = \{\boldsymbol{\eta} + \ell_{\boldsymbol{\eta}, \boldsymbol{\lambda}}[A] \boldsymbol{\lambda} \in \mathbb{R}^M: \boldsymbol{\lambda} \in \mathcal{S}_+^{M-1}\}.
	\label{eqn:polar_reconstruction}
\end{equation}
In other words, we can interpret the set of polar surfaces $\mathbb{L}_{\boldsymbol{\eta}}$ as a set of sets of vectors which can be completely characterised by its projected lengths. Geometrically speaking, any polar surface $A \in \mathbb{L}_{\boldsymbol{\eta}}$ can be obtained by translating and then stretching the set of positive unit vectors $\mathcal{S}_+^{M-1}$ along the positive radial directions. Our main result in \cref{thm:polar_parameterisation} states that the set of all Pareto front surfaces is indeed a special subset of this set: $\mathbb{Y}^*_{\boldsymbol{\eta}} \subset \mathbb{L}_{\boldsymbol{\eta}}$. Precisely, it is the subset where the Pareto partial ordering is preserved and the projected length function has an explicit formula given in terms of the length scalarisation function \eqref{eqn:length_scalarisation}. We present the proof of this result in \cref{app:proofs:thm:polar_parameterisation} and an illustration of it in \cref{fig:polar_parameterisation}. 
\begin{theorem}
	[Polar parameterisation] For any bounded set of vectors $A \subset \mathbb{R}^M$ and reference vector $\boldsymbol{\eta} \in \mathbb{R}^M$, if the corresponding Pareto front surface is not empty $\mathcal{Y}_{\boldsymbol{\eta}}^{\textnormal{int}}[A] \neq \emptyset$, then it admits the following polar parameterisation:
	\begin{equation}
		\mathcal{Y}_{\boldsymbol{\eta}}^{\textnormal{int}}[A]
		= \biggl\{\boldsymbol{\eta} + \sup_{\mathbf{a} \in A} s_{\boldsymbol{\eta}, \boldsymbol{\lambda}}(\mathbf{a}) \boldsymbol{\lambda} \in \mathbb{R}^M: \boldsymbol{\lambda} \in \mathcal{S}_+^{M-1} \biggr\}
		\label{eqn:polar_parameterisation}
	\end{equation}
	where $\ell_{\boldsymbol{\eta}, \boldsymbol{\lambda}}[\mathcal{Y}_{\boldsymbol{\eta}}^{\textnormal{int}}[A]] = \sup_{\mathbf{a} \in A} s_{\boldsymbol{\eta}, \boldsymbol{\lambda}}(\mathbf{a})$ is the projected length along $\boldsymbol{\lambda} \in \mathcal{S}_+^{M-1}$.
	\label{thm:polar_parameterisation}
\end{theorem}
\begin{remark}
	[Singleton front] When the Pareto front surface is empty, $\mathcal{Y}_{\boldsymbol{\eta}}^{\textnormal{int}}[A] = \emptyset$, then the right hand side of \eqref{eqn:polar_parameterisation} evaluates to the degenerate polar surface $\{\boldsymbol{\eta}\} \in \mathbb{L}_{\boldsymbol{\eta}}$. This event can happen when the reference vector $\boldsymbol{\eta} \in \mathbb{R}^M$ is set too aggressively in such a way as it weakly dominates the entire feasible objective space.
	\label{rem:singleton}
\end{remark}

\begin{figure}
	\includegraphics[width=1\linewidth]{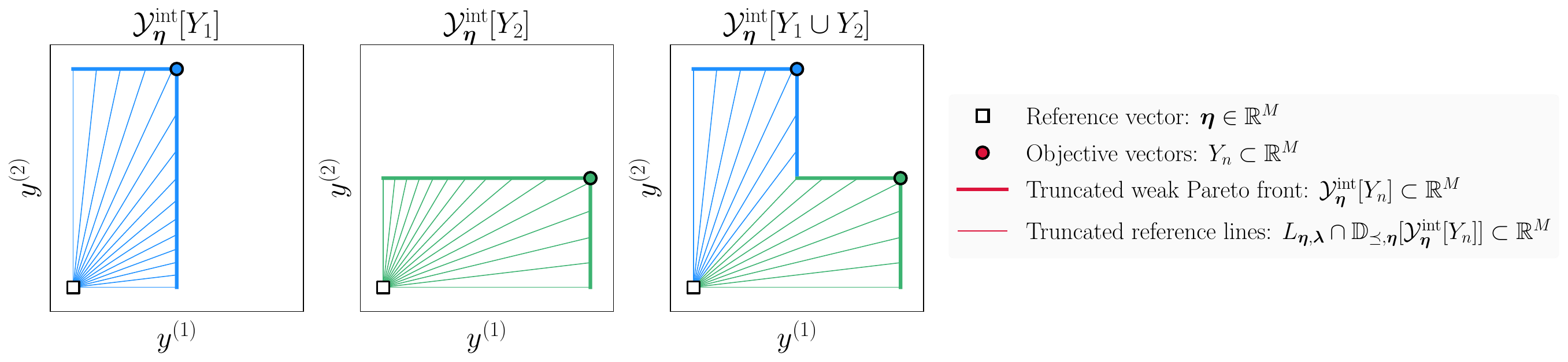}
	\centering
	\caption{An illustration of the polar parameterisation associated with a set of two points.}
	\label{fig:intuition}
\end{figure}

\begin{remark}
	[Conceptual idea] The explicit form of the polar parameterisation \eqref{eqn:polar_parameterisation} can be derived in an constructive manner. Firstly, one can show that the Pareto front surface of a single point $\mathbf{a} \in \mathbb{D}_{\succsucc}[\{\boldsymbol{\eta}\}]$ can be written as the polar surface
	\begin{equation*}
		\mathcal{Y}_{\boldsymbol{\eta}}^{\textnormal{int}}[\{\mathbf{a}\}]
		= \{\boldsymbol{\eta} + s_{\boldsymbol{\eta}, \boldsymbol{\lambda}}(\mathbf{a}) \boldsymbol{\lambda} \in \mathbb{R}^M: \boldsymbol{\lambda} \in \mathcal{S}_+^{M-1} \} \in \mathbb{L}_{\boldsymbol{\eta}}.
	\end{equation*}
	Consequently, we can then determine the Pareto front surface of any set of points $A \subset \mathbb{D}_{\succsucc}[\{\boldsymbol{\eta}\}]$ by simply computing the union of the corresponding individual Pareto front surfaces and then removing all the points which are sub-optimal, that is 
	\begin{align*}
		\mathcal{Y}_{\boldsymbol{\eta}}^{\textnormal{int}}[A]
		&= 
		\mathcal{Y}_{\boldsymbol{\eta}}^{\textnormal{int}}[\cup_{\mathbf{a} \in A} \mathcal{Y}_{\boldsymbol{\eta}}^{\textnormal{int}}[\{\mathbf{a}\}]]
		\\
		&= \mathcal{Y}_{\boldsymbol{\eta}}^{\textnormal{int}}[\{\boldsymbol{\eta} + s_{\boldsymbol{\eta}, \boldsymbol{\lambda}}(\mathbf{a}) \boldsymbol{\lambda} \in \mathbb{R}^M: \mathbf{a} \in A, \boldsymbol{\lambda} \in \mathcal{S}_+^{M-1}\}
		]
		\\
		&= \biggl\{\boldsymbol{\eta} + \sup_{\mathbf{a} \in A} s_{\boldsymbol{\eta}, \boldsymbol{\lambda}}(\mathbf{a}) \boldsymbol{\lambda} \in \mathbb{R}^M: \boldsymbol{\lambda} \in \mathcal{S}_+^{M-1} \biggr\}.
	\end{align*}
	Remarkably, this overall procedure turns out to be equivalent to just retaining the points which achieve the maximal projected lengths along each positive direction $\boldsymbol{\lambda} \in \mathcal{S}^{M-1}_+$. An illustration of this conceptual idea is presented in \cref{fig:intuition}. Note however that the proof of the result in \cref{app:proofs:thm:polar_parameterisation} does not actually follow this intuitive construction. Instead, our proof is derived in a more technical way by taking advantage of a well-known optimisation result regarding the Chebyshev scalarisation function \cite[Part 2, Theorem 3.4.5]{miettinen1998}. The primary reason why we take this more general approach is to ensure that our polar parameterisation result holds for sets lying in the whole objective space $\mathbb{R}^M$ as opposed to just the truncated objective space $\mathbb{D}_{\succsucc}[\{\boldsymbol{\eta}\}]$.
	\label{rem:intuition}
\end{remark}
\begin{remark}
	[Other representations] The coordinate system transformation that we performed in \eqref{eqn:coordinate_transformation} is appealing because it admits a representation for the Pareto front surface, which is simple, explicit and has a linear dependency on the projected length values. Nevertheless, we can easily obtain other nonlinear representations of a Pareto front surface by taking advantage of transformation functions. For example, let $\tau: \mathbb{R}^M \rightarrow \mathbb{R}^M$ denote an invertible and strictly monotonically increasing transformation function such that $\mathbf{y} \succeq \mathbf{y}' \iff \tau(\mathbf{y}) \succeq \tau(\mathbf{y}')$ for any $\mathbf{y}, \mathbf{y}' \in \mathbb{R}^M$. Then, by a simple monotonicity argument, we can show that the following parameterisation of the Pareto front surface is also valid: 
	\begin{equation*}
		\mathcal{Y}_{\boldsymbol{\eta}}^{\textnormal{int}}[A]
		= \biggl\{\tau^{-1} \Bigl( \tau(\boldsymbol{\eta}) + \sup_{\mathbf{a} \in A} 
		s_{\tau(\boldsymbol{\eta}), \boldsymbol{\lambda}}(\tau(\mathbf{a})) \boldsymbol{\lambda} \Bigr) \in \mathbb{R}^M: \boldsymbol{\lambda} \in \mathcal{S}_+^{M-1} \biggr\}.
	\end{equation*}
	Intuitively, this reformulation works by performing the polar parameterisation in the transformed space before inverting the result. Note that strict monotonicity is required here because we want the Pareto front surface to be invariant under these transformations: $\mathcal{Y}_{\boldsymbol{\eta}}^{\textnormal{int}}[A] = \tau^{-1}(\mathcal{Y}_{\tau(\boldsymbol{\eta})}^{\textnormal{int}}[\tau(A)])$.
	\label{rem:representation}
\end{remark}
\begin{remark}
	[Computational cost] For a finite set $A \subset \mathbb{R}^M$, the worst-case cost involved with computing the strictly Pareto optimal front $\mathcal{Y}^{\textnormal{strict}}[A]\subset \mathbb{R}^M$ is $\mathcal{O}(|A|^2 M)$. In contrast, the worst-case cost involved with computing a finite approximation of the polar parameterised Pareto front surface $\mathcal{Y}_{\boldsymbol{\eta}}^{\textnormal{int}}[A]\subset \mathbb{R}^M$ is $\mathcal{O}(|\Lambda||A|M)$, where $\Lambda = \{\boldsymbol{\lambda}_1, \boldsymbol{\lambda}_2, \dots\}$ denotes a finite set of positive unit vectors. 
\end{remark}

\subsection{Order-preserving transformations}
\label{sec:algebra}
\begin{figure}
	\includegraphics[width=1\linewidth]{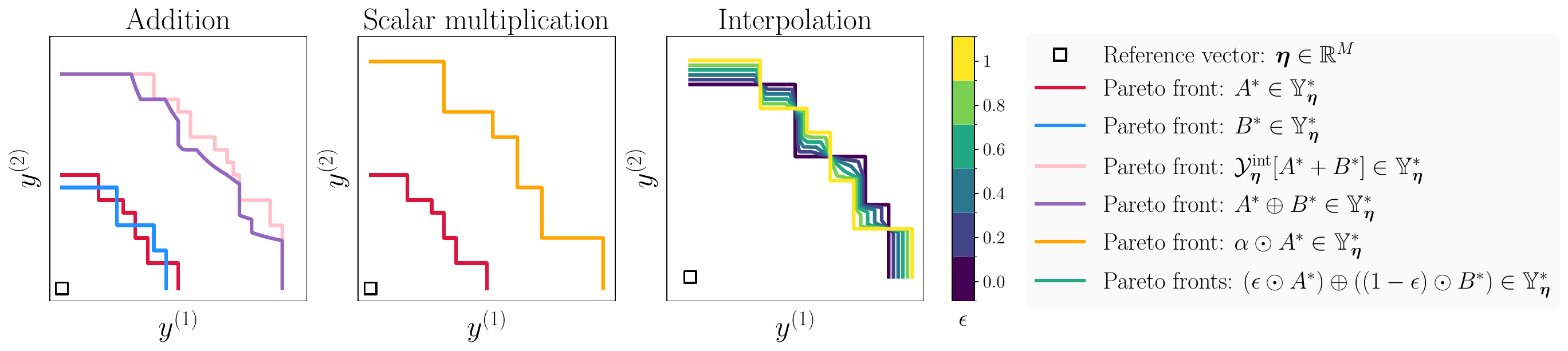}
	\centering
	\caption{An illustration of some algebraic operations on the space of Pareto front surfaces in $M=2$ dimensions. For these examples, we set the scalar multipliers to be $\alpha=2$ and $\epsilon \in \{0.0, 0.2, 0.4, 0.6, 0.8, 1.0\}$.}
	\label{fig:pareto_front_algebra}
\end{figure}

The polar parameterisation result in \cref{thm:polar_parameterisation} tells us that any Pareto front surface is completely defined by its projected lengths. Conceptually, this means that we can define transformations on the space of Pareto front surfaces by simply applying transformations on the projected lengths. As long as these transformation operations preserve the Pareto partial ordering, then we can be assured that resulting set is also a valid Pareto front surface. In \cref{prop:pareto_front_conditions} below, we give two necessary and sufficient conditions for this to happen. The first condition \eqref{eqn:pareto_condition_1} states that the projected lengths have to be positive---this condition ensures that the resulting Pareto front surface is non-empty. The second condition \eqref{eqn:pareto_condition_2} is the Pareto optimality condition---this ensures that we cannot find two vectors in the set such that one strongly dominates the other. The necessity of this latter condition follows from the spirit of \cref{lemma:domination_equivalence}, which presents some equivalent statements for Pareto domination based on the length scalarisation function \eqref{eqn:length_scalarisation}. The proof of both of these result are presented in \cref{app:proofs:prop:pareto_front_conditions} and \cref{app:proofs:lemma:domination_equivalence}, respectively.
\begin{lemma}
	[Domination equivalence] For any Pareto front surface $A^* \in \mathbb{Y}_{\boldsymbol{\eta}}^*$ and vector $\mathbf{y} \in \mathbb{D}_{\succsucc}[\{\boldsymbol{\eta}\}]$, we have the following equivalences:
	\begin{align*}
		\mathbf{y} \in \mathbb{D}_{\preceq, \boldsymbol{\eta}}[A^*] \iff s_{\boldsymbol{\eta}, \boldsymbol{\lambda}_{\boldsymbol{\eta}}^*(\mathbf{y})}(\mathbf{y})
		\leq \ell_{\boldsymbol{\eta}, \boldsymbol{\lambda}_{\boldsymbol{\eta}}^*(\mathbf{y})} [A^*],
		\\
		\mathbf{y} \in \mathbb{D}_{\precprec, \boldsymbol{\eta}}[A^*] \iff s_{\boldsymbol{\eta}, \boldsymbol{\lambda}_{\boldsymbol{\eta}}^*(\mathbf{y})}(\mathbf{y})
		< \ell_{\boldsymbol{\eta}, \boldsymbol{\lambda}_{\boldsymbol{\eta}}^*(\mathbf{y})} [A^*],
		\\
		\mathbf{y} \in \mathbb{D}_{\succsucc, \boldsymbol{\eta}}[A^*] \iff s_{\boldsymbol{\eta}, \boldsymbol{\lambda}_{\boldsymbol{\eta}}^*(\mathbf{y})}(\mathbf{y})
		> \ell_{\boldsymbol{\eta}, \boldsymbol{\lambda}_{\boldsymbol{\eta}}^*(\mathbf{y})} [A^*],
		\\
		\mathbf{y} \in \mathbb{D}_{\succeq, \boldsymbol{\eta}}[A^*] \iff s_{\boldsymbol{\eta}, \boldsymbol{\lambda}_{\boldsymbol{\eta}}^*(\mathbf{y})}(\mathbf{y})
		\geq \ell_{\boldsymbol{\eta}, \boldsymbol{\lambda}_{\boldsymbol{\eta}}^*(\mathbf{y})} [A^*],
	\end{align*}
	where $s_{\boldsymbol{\eta}, \boldsymbol{\lambda}_{\boldsymbol{\eta}}^*(\mathbf{y})}(\mathbf{y}) = ||\mathbf{y}-\boldsymbol{\eta}||_{L^2}$ and $\ell_{\boldsymbol{\eta}, \boldsymbol{\lambda}_{\boldsymbol{\eta}}^*(\mathbf{y})} [A^*] = \max_{\mathbf{a} \in A^*} s_{\boldsymbol{\eta}, \boldsymbol{\lambda}_{\boldsymbol{\eta}}^*(\mathbf{y})}(\mathbf{a})$.
	\label{lemma:domination_equivalence}
\end{lemma}
\begin{proposition}
	[Pareto front conditions] Consider a polar surface $A \in \mathbb{L}_{\boldsymbol{\eta}}$, then this set is a Pareto front surface, that is $A = \mathcal{Y}_{\boldsymbol{\eta}}^{\textnormal{int}}[A]$, if and only if the following two conditions holds:
	\begin{enumerate}[label=C\arabic*., ref=C\arabic*]
		\item The positive lengths condition: $\ell_{\boldsymbol{\eta}, \boldsymbol{\lambda}}[A] > 0$ for all $\boldsymbol{\lambda} \in \mathcal{S}_+^{M-1}$.
		\label{eqn:pareto_condition_1}
		\item The maximum ratio condition: $\max_{m=1,\dots,M} \frac{\ell_{\boldsymbol{\eta}, \boldsymbol{\lambda}}[A] \lambda^{(m)}}{\ell_{\boldsymbol{\eta}, \boldsymbol{\upsilon}}[A] \upsilon^{(m)}} \geq 1$ for all $\boldsymbol{\lambda}, \boldsymbol{\upsilon} \in \mathcal{S}_+^{M-1}$.
		\label{eqn:pareto_condition_2}
	\end{enumerate}
	\label{prop:pareto_front_conditions}
\end{proposition}
\cref{prop:pareto_front_conditions} gives us two simple conditions on the projected length function which can be easily checked whenever we want to determine whether a Pareto front surface is valid. For convenience, we will say a transformation (or an operation) acting on the space of Pareto front surfaces is order-preserving if the resulting projected length function satisfies the two conditions in \cref{prop:pareto_front_conditions}. Below, we present some notable order-preserving operations which are defined over the general space of polar surfaces. Firstly, we present the union operation in \cref{eg:union}, which formalises the idea described in \cref{rem:intuition}. 
\begin{example}
	[Union] Consider two Pareto front surfaces $A^*, B^* \in \mathbb{Y}_{\boldsymbol{\eta}}^*$, by \cref{thm:polar_parameterisation}, we can compute the Pareto front surface of the union of these two sets by simply taking the maximum of the projected lengths, that is 
	\label{eg:union}
	\begin{equation*}
		\mathcal{Y}_{\boldsymbol{\eta}}^{\textnormal{int}}[A^* \cup B^*]
		= \{\boldsymbol{\eta} + \max(\ell_{\boldsymbol{\eta}, \boldsymbol{\lambda}}[A^*], \ell_{\boldsymbol{\eta}, \boldsymbol{\lambda}}[B^*]) \boldsymbol{\lambda} \in \mathbb{R}^M: \boldsymbol{\lambda} \in \mathcal{S}_+^{M-1}\} \in \mathbb{Y}^*_{\boldsymbol{\eta}}.
	\end{equation*}
\end{example}
Besides this, we also define the addition and scalar multiplication operation over the space of polar surfaces in  \cref{eg:addition} and \cref{eg:scalar_multiplication}, respectively. These two concepts are very useful because they give us a simple way to perform standard linear operations over the space of polar surfaces. 
Later on, in \cref{sec:statistics}, we will implicitly use these linear operations in order to define the concept of integration over the space of Pareto front surfaces. Explicitly, we used these ideas in \cref{sec:expectation} when we define the expected Pareto front surface. 
\begin{example}
	[Addition] Consider two Pareto front surfaces $A^*, B^* \in \mathbb{Y}^*_{\boldsymbol{\eta}}$, we define the sum of these two Pareto front surfaces by the equation
	\begin{equation*}
		A^* \oplus B^* := \{\boldsymbol{\eta} + (\ell_{\boldsymbol{\eta}, \boldsymbol{\lambda}}[A^*] + \ell_{\boldsymbol{\eta}, \boldsymbol{\lambda}}[B^*]) \boldsymbol{\lambda} \in \mathbb{R}^M: \boldsymbol{\lambda} \in \mathcal{S}_+^{M-1}\} \in \mathbb{Y}^*_{\boldsymbol{\eta}}.
	\end{equation*}
	This set is a Pareto front surface because it satisfies the conditions in \cref{prop:pareto_front_conditions}. Note that this sum is not necessarily equivalent to the Pareto front surface obtained by adding the two sets together: $\mathcal{Y}_{\boldsymbol{\eta}}^{\textnormal{int}}[A^* + B^*] \neq A^* \oplus B^* $, where $A + B := \{\mathbf{a} + \mathbf{b} \in \mathbb{R}^M: \mathbf{a} \in A, \mathbf{b} \in B\}$ denotes the Minkowski sum for any set of vectors $A, B \subset \mathbb{R}^M$. We illustrate this subtle distinction in the left plot of \cref{fig:pareto_front_algebra}.
	\label{eg:addition}
\end{example}

\begin{example}
	[Scalar multiplication] Consider a positive scalar $\epsilon > 0$ and a Pareto front surface $A^* \in \mathbb{Y}_{\boldsymbol{\eta}}^*$, we define the scalar multiplication operation by the equation
	\begin{equation*}
		\epsilon \odot A^* := \{\boldsymbol{\eta} + \epsilon \ell_{\boldsymbol{\eta}, \boldsymbol{\lambda}}[A^*] \boldsymbol{\lambda} \in \mathbb{R}^M: \boldsymbol{\lambda} \in \mathcal{S}_+^{M-1}\} \in \mathbb{Y}^*_{\boldsymbol{\eta}}.
	\end{equation*}
	This set is a Pareto front surface because it satisfies the conditions in \cref{prop:pareto_front_conditions}. We illustrate an example of this operation in the middle plot of \cref{fig:pareto_front_algebra}.
	\label{eg:scalar_multiplication}
\end{example}

\subsection{Utility functions}
\label{sec:utility}
Utility functions are often used in multi-objective optimisation in order to assess the quality of a Pareto front approximation that is comprised of a finite set of points. In this section, we introduce a family of utility functions based on the length scalarisation function \eqref{eqn:length_scalarisation}. We show that this family of utility functions are both interpretable and satisfy many desirable properties such as strict Pareto compliancy (\cref{prop:strict_pareto_compliancy}). 

\paragraph{Length-based R2 utilities} The R2 utilities \citep{hansen1998trimm} are a special family of utility functions that are constructed using scalarisation functions. They possess many general and desirable properties \citep{tu2024a} which makes them very appealing to work with. By using the length scalarisation function \eqref{eqn:length_scalarisation}, we define a special subfamily of the R2 utilities which we call the length-based R2 utilities. Precisely, a utility function is a length-based R2 utility if it can be written in the form
\begin{align}
	\begin{split}
		U_{\boldsymbol{\eta}, \tau}[Y] 
		:= \mathbb{E}_{\boldsymbol{\lambda} \sim \text{Uniform}(\mathcal{S}_+^{M-1})}
		\biggl[
		\max_{\mathbf{y} \in Y} \tau(s_{\boldsymbol{\eta}, \boldsymbol{\lambda}}(\mathbf{y}))
		\biggr]
		\\
		= \mathbb{E}_{\boldsymbol{\lambda} \sim \text{Uniform}(\mathcal{S}_+^{M-1})}
		\bigl[
		\tau(\ell_{\boldsymbol{\eta}, \boldsymbol{\lambda}}[\mathcal{Y}_{\boldsymbol{\eta}}^{\text{int}}[Y]])
		\bigr],
	\end{split}
	\label{eqn:length_based_utility}
\end{align}
for any finite set of objective vectors $Y \subset \mathbb{R}^M$, where $\tau: \mathbb{R}_{\geq 0} \rightarrow \mathbb{R}$ is any strictly monotonically increasing transformation that is defined over the space of non-negative scalars. In our case, this family of utility functions assesses the quality of a Pareto front approximation based on some notion of the average length away from the reference vector $\boldsymbol{\eta} \in \mathbb{R}^M$. In this setting, a greater utility is more desirable because it means that the set is much further away, on average, from the reference vector.

\paragraph{Hypervolume indicator} A special case of these length-based R2 utilities is the hypervolume indicator \citep{zitzler1998ppsn}, which is a popular performance metric in multi-objective optimisation. The hypervolume indicator of a finite set $Y \subset \mathbb{R}^M$ is defined as the volume of its truncated domination region:
\begin{equation}
	U_{\boldsymbol{\eta}}^{\text{HV}}[Y] 
	:= \nu[\mathbb{D}_{\preceq, \boldsymbol{\eta}}[Y]]
	= \mathbb{E}_{\boldsymbol{\lambda} \sim \text{Uniform}(\mathcal{S}_+^{M-1})}
	\biggl[\max_{\mathbf{y} \in Y} c_M (s_{\boldsymbol{\eta}, \boldsymbol{\lambda}}(\mathbf{y}))^M \biggr]
	\label{eqn:hypervolume_indicator}
\end{equation}
where $\nu[A] := \int_{\mathbb{R}^M} \mathbbm{1}[\mathbf{y} \in A] d \mathbf{y}$ is the Lebesgue measure on $\mathbb{R}^M$ with $A \subset \mathbb{R}^M$ denoting a measurable set and $\mathbbm{1}$ denoting the indicator function. The second equality above in \eqref{eqn:hypervolume_indicator} shows that the hypervolume indicator can be written in terms of a transformation of the length scalarisation function $\tau^{\text{HV}}(x) = c_M x^M$, where $c_M = \pi^{M/2}2^{-M} \Gamma(M/2 + 1)^{-1}$ is a positive scalar depending on the Gamma function $\Gamma(z) = \int_0^\infty t^{z-1} e^{-t} dt$. This equality has been proved in many earlier works \citep{shang2018pgecc,deng2019itec,zhang2020icml}. In the left of \cref{fig:hypervolume_utility_and_distance}, we present a simple illustration of the hypervolume indicator in two dimensions. Geometrically, the hypervolume scalarised value can also be interpreted as being equal to the volume of the hypersphere with diameter $s_{\boldsymbol{\eta}, \boldsymbol{\lambda}}(\mathbf{y}) \geq 0$, that is $\tau^{\textnormal{HV}}(s_{\boldsymbol{\eta}, \boldsymbol{\lambda}}(\mathbf{y})) = \nu[\{\mathbf{z} \in \mathbb{R}^M: ||\mathbf{z}||_{L^2} \leq s_{\boldsymbol{\eta}, \boldsymbol{\lambda}}(\mathbf{y}) / 2\}]$ for any objective vector $\mathbf{y} \in \mathbb{R}^M$---see the right of \cref{fig:length_based_scalarisations} for a visualisation of this interpretation in two dimensions.

\paragraph{General properties} By design, the length-based R2 utilities \eqref{eqn:length_based_utility} inherit all of the standard properties that any R2 utility satisfies such as the monotone and submodularity property \citep[Proposition 3.1]{tu2024a}. In the following, we will additionally show that this special subfamily of utility functions also satisfies the strict Pareto compliancy property over the truncated space (\cref{prop:strict_pareto_compliancy}). Loosely speaking, the strict Pareto compliancy property (\cref{def:strict_pareto_compliancy}) states that if a set of vectors is strictly better than another, then it should have a higher utility. This result follows from the fact that the hypervolume indicator is strictly Pareto compliant over the truncated space \citep{zitzler2003itec}. The full proof of this result is presented in \cref{app:proofs:prop:strict_pareto_compliancy}.

\begin{definition}
	[Set domination] Consider the subsets $A, B \subset \mathbb{R}^M$, we say that the set $A$ weakly or strictly
	dominates the set $B$ if and only if its weak Pareto domination region weakly or strictly contains the other, respectively:
	\label{def:set_domination}
	\begin{align*}
		A \succeq B &\iff \mathbb{D}_{\preceq}[A] \supseteq \mathbb{D}_{\preceq}[B],
		\\
		A \succ B &\iff \mathbb{D}_{\preceq}[A] \supset \mathbb{D}_{\preceq}[B].
	\end{align*}
\end{definition}

\begin{definition}
	[Strict Pareto compliancy] A utility function $U: 2^{\mathbb{R}^M} \rightarrow \mathbb{R}$ is strictly Pareto compliant over a set $Y \subseteq \mathbb{R}^M$, if for all finite sets $A, B \subseteq Y$, we have that $A \succ B \implies U(A) > U(B)$.
	\label{def:strict_pareto_compliancy}
\end{definition}

\begin{proposition}
	For any reference vector $\boldsymbol{\eta} \in \mathbb{R}^M$ and strictly monotonically increasing transformation $\tau: \mathbb{R}_{\geq 0} \rightarrow \mathbb{R}$, the corresponding length-based R2 utility \eqref{eqn:length_based_utility} satisfies the strict Pareto compliancy property over the truncated space $\mathbb{D}_{\succsucc}[\{\boldsymbol{\eta}\}] \subset \mathbb{R}^M$.
	\label{prop:strict_pareto_compliancy}
\end{proposition}
\subsection{Loss functions}
\label{sec:loss}
Loss functions are an important aspect of decision theory because they quantify the cost associated with performing a particular decision or action. In this section, we define the frontier loss functions, which is a family of loss functions acting on the space of Pareto front surfaces, or more generally the space of polar surfaces $\mathbb{L}_{\boldsymbol{\eta}}$. These loss functions are useful because they give us a way to compare between any two Pareto front surfaces. Later on in \cref{sec:decision_theory}, we will show how these loss functions can be used in order to define various probabilistic concepts on the space of random Pareto front surfaces such as the expectation (\cref{sec:expectation}) and quantiles (\cref{sec:quantiles}).
	
\paragraph{Frontier loss} We define the frontier loss function $\mathcal{D}_{\boldsymbol{\eta}, S}: \mathbb{L}_{\boldsymbol{\eta}} \times \mathbb{L}_{\boldsymbol{\eta}} \rightarrow \mathbb{R}_{\geq 0}$ as the average loss incurred along each reference line, that is
\begin{equation}
	\mathcal{D}_{\boldsymbol{\eta}, S}[A, B] := \mathbb{E}_{\boldsymbol{\lambda} \sim \text{Uniform}(\mathcal{S}_+^{M-1})}[
	S(
	\ell_{\boldsymbol{\eta}, \boldsymbol{\lambda}}[A], 
	\ell_{\boldsymbol{\eta}, \boldsymbol{\lambda}}[B]
	)
	]
	\label{eqn:frontier_distance_function}
\end{equation}
for any two polar surfaces $A, B \in \mathbb{L}_{\boldsymbol{\eta}}$, where $S: \mathbb{R} \times \mathbb{R} \rightarrow \mathbb{R}_{\geq 0}$ denotes a non-negative scoring function (or loss function). Clearly, different choices of scoring functions will lead to different notions of loss. For example, later on in \cref{tab:scoring_functions}, we present some popular scoring functions which are commonly used in the field of probabilistic forecasting---see \cite{gneiting2011jasa} for a discussion. 

\paragraph{Frontier distance} If the scoring function $S$ is a metric in $\mathbb{R}$, then the resulting frontier loss function can be interpreted as a type of distance function over the space of polar surfaces $\mathbb{L}_{\boldsymbol{\eta}}$. Note though that this frontier distance function will only be a pseudometric in $\mathbb{L}_{\boldsymbol{\eta}}$. That is, unlike a metric, the distance between any two polar surfaces $A, B \in \mathbb{L}_{\boldsymbol{\eta}}$ will be zero, $\mathcal{D}_{\boldsymbol{\eta}, S}[A, B] = 0$, if and only if $\nu[\{\ell_{\boldsymbol{\eta}, \boldsymbol{\lambda}}[A] \neq \ell_{\boldsymbol{\eta}, \boldsymbol{\lambda}}[B]: \boldsymbol{\lambda} \in \mathcal{S}_+^{M-1}\}] = 0$. In other words, the polar surfaces $A, B \in \mathbb{L}_{\boldsymbol{\eta}}$ are equivalent under this frontier distance if and only if their projected lengths disagree on a set of measure zero.

\begin{figure}
	\includegraphics[width=0.8\linewidth]{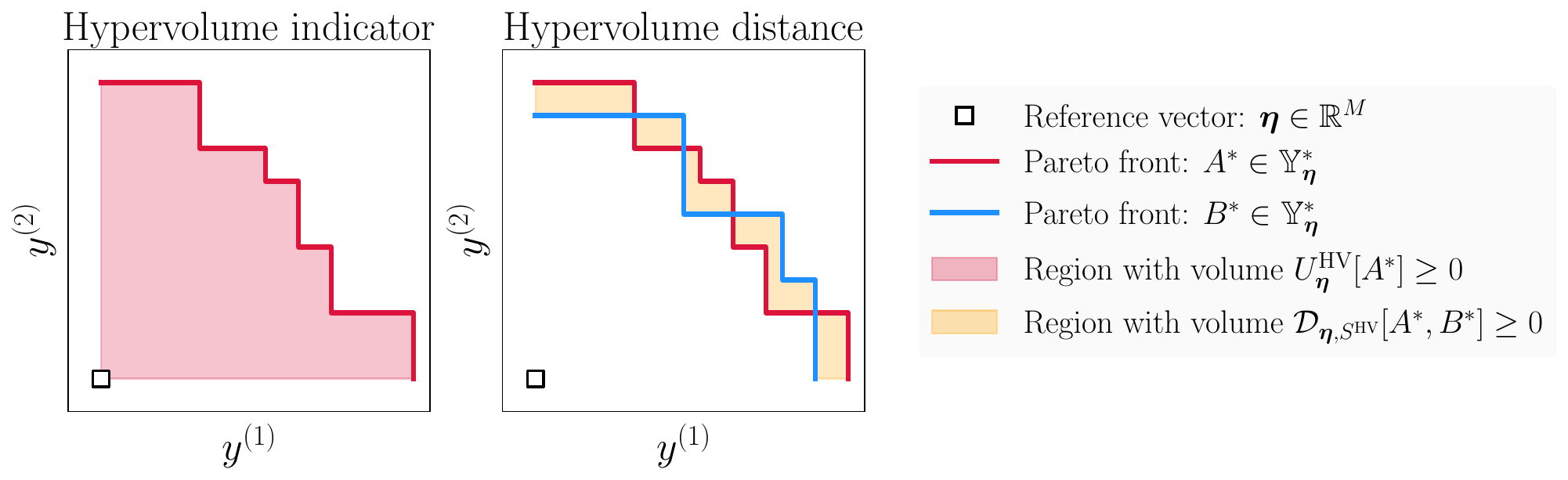}
	\centering
	\caption{An illustration of the hypervolume indicator and hypervolume distance in $M=2$ dimensions.}
	\label{fig:hypervolume_utility_and_distance}
\end{figure}

\paragraph{Hypervolume distance} A special and interpretable case of the frontier distance function is the hypervolume distance. The hypervolume distance between any two Pareto front surfaces is the volume of the symmetric difference between their corresponding truncated dominated regions. Specifically, if we restrict the distance function to the space of Pareto front surfaces $\mathbb{Y}^*_{\boldsymbol{\eta}}$ and set $S^{\text{HV}}(x, y) := |\tau^{\text{HV}}(x) - \tau^{\text{HV}}(y)|$, then we recover the hypervolume distance
\begin{align}
	\begin{split}
		\mathcal{D}_{\boldsymbol{\eta}, S^{\text{HV}}}[A^*, B^*] 
		&:=\nu[\mathbb{D}_{\preceq, \boldsymbol{\eta}}[A^*] \triangle \mathbb{D}_{\preceq, \boldsymbol{\eta}}[B^*]]
		= 2 U_{\boldsymbol{\eta}}^{\text{HV}}[A^* \cup B^*]
		- U_{\boldsymbol{\eta}}^{\text{HV}}[A^*]
		- U_{\boldsymbol{\eta}}^{\text{HV}}[B^*]
	\end{split}
	\label{eqn:hypervolume_distance}
\end{align}
for any Pareto front surfaces $A^*, B^* \in \mathbb{Y}^*_{\boldsymbol{\eta}}$, where $A \triangle B = (A \setminus B) \cup (B \setminus A)$ denotes the symmetric difference between the sets $A, B \subset \mathbb{R}^M$---see the right of \cref{fig:hypervolume_utility_and_distance} for an illustration of this distance function on a two-dimensional example.

\begin{remark}
	[Generalised frontier loss] The frontier loss in \eqref{eqn:frontier_distance_function} is defined using the length scalarisation function \eqref{eqn:length_scalarisation} and a uniform distribution over the scalarisation parameters. Naturally, we can generalise this construction further by considering an arbitrary scalarisation function and parameter distribution. This generalisation bares some similarity with the construction of the R2 utilities \citep{hansen1998trimm,tu2024a}.
\end{remark}

\section{Pareto front surface statistics}
\label{sec:statistics}
Real-world problems are typically noisy and subject to many sources of uncertainty. In order to quantify this uncertainty, decision makers often appeal to ideas from Probability and Statistics. In this section, we will generalise these existing ideas to function over the space of Pareto front surfaces $\mathbb{Y}^*_{\boldsymbol{\eta}}$. Precisely, we will study the concept of a random Pareto front surface and showcase how our polar parameterisation result (\cref{thm:polar_parameterisation}) can be easily used in order to define many statistical quantities of interest. Intuitively, we will treat the Pareto front surface as an infinite-dimensional random variable \eqref{eqn:stochastic_pareto_front} and define the statistics according to its corresponding finite-dimensional distributions. Before we introduce these ideas in more detail, we will first describe the general set up and assumptions that we will be using throughout this section.

\paragraph{Formulation} Consider a standard probability space $(\Omega, \mathcal{F}, \mathbb{P})$, where $\Omega$ denotes the sample space, $\mathcal{F}$ denotes a $\sigma$-algebra and $\mathbb{P}$ denotes a probability measure. In this section, we will be interested in studying the Pareto front surface associated with some vector-valued random function $f: \mathbb{X} \times \Omega \rightarrow \mathbb{R}^M$, which we have indexed with the inputs $\mathbf{x} \in \mathbb{X}$. For a fixed and known reference vector $\boldsymbol{\eta} \in \mathbb{R}^M$, we denote the corresponding Pareto front surface by
\begin{align}
	\begin{split}
		Y_{\boldsymbol{\eta}, f}^* (\boldsymbol{\omega})
		&:= \mathcal{Y}_{\boldsymbol{\eta}}^{\textnormal{int}}[\{f(\mathbf{x}, \boldsymbol{\omega})\}_{\mathbf{x} \in \mathbb{X}}]
		= \biggl\{
		\boldsymbol{\eta} + 
		\sup_{\mathbf{x} \in \mathbb{X}} s_{\boldsymbol{\eta}, \boldsymbol{\lambda}}(f(\mathbf{x}, \boldsymbol{\omega}))
		\boldsymbol{\lambda}
		\in \mathbb{R}^M: \boldsymbol{\lambda} \in \mathcal{S}_+^{M-1} \biggr\},
		\label{eqn:stochastic_pareto_front}
	\end{split}
\end{align}
for $\boldsymbol{\omega} \in \Omega$, where $\ell_{\boldsymbol{\eta}, \boldsymbol{\lambda}}[Y_{\boldsymbol{\eta}, f}^* (\boldsymbol{\omega})] = \sup_{\mathbf{x} \in \mathbb{X}} s_{\boldsymbol{\eta}, \boldsymbol{\lambda}}(f(\mathbf{x}, \boldsymbol{\omega})) \geq 0$ are the corresponding the projected lengths along the positive directions $\boldsymbol{\lambda} \in \mathcal{S}_+^{M-1}$. 

\paragraph{Projected lengths} By exploiting the explicit form of the random Pareto front surface in \eqref{eqn:stochastic_pareto_front}, we see that the only dependence on $\boldsymbol{\omega} \in \Omega$ arises solely in the projected lengths $\ell_{\boldsymbol{\eta}, \boldsymbol{\lambda}}[Y_{\boldsymbol{\eta}, f}^* (\boldsymbol{\omega})] \geq 0$ for $\boldsymbol{\lambda} \in \mathcal{S}_+^{M-1}$. Therefore all of the distributional information about the Pareto front surface is completely characterised by its projected lengths
\begin{equation}
	\mathcal{L}_{\boldsymbol{\eta}}(\boldsymbol{\omega}) := \{\ell_{\boldsymbol{\eta}, \boldsymbol{\lambda}}[Y_{\boldsymbol{\eta}, f}^* (\boldsymbol{\omega})] \geq 0: \boldsymbol{\lambda} \in \mathcal{S}_+^{M-1}\}.
	\label{eqn:length_process}
\end{equation}
Note that if we used a nonlinear representation for the Pareto front surface instead (\cref{rem:representation}), then it is unlikely that such a simplification would have been possible.

\paragraph{Assumptions} To avoid dealing with some unnecessary complications, we will make the following two simplifying assumptions (\cref{ass:positive_lengths} and \cref{ass:bounded_lengths}). Together, both of these assumptions ensures that the polar parameterisation of the Pareto front surface, described on the right of \eqref{eqn:stochastic_pareto_front}, holds almost surely. Moreover, \cref{ass:bounded_lengths} also ensures that the projected length $\ell_{\boldsymbol{\eta}, \boldsymbol{\lambda}}[Y_{\boldsymbol{\eta}, f}^*] \geq 0$, along any positive direction $\boldsymbol{\lambda} \in \mathcal{S}_+^{M-1}$, is bounded almost surely. Consequently, this means that the moments and quantiles of the projected lengths exist and are finite.

\begin{assumption}
	[Positive lengths] The Pareto front surface $Y_{\boldsymbol{\eta}, f}^*$ is non-empty\footnote{Note that we could easily lift this non-empty assumption by simply treating every empty Pareto front surface as being equivalent to the degenerate singleton set $\{\boldsymbol{\eta}\} \in \mathbb{L}_{\boldsymbol{\eta}}$.} almost surely.
	\label{ass:positive_lengths}
\end{assumption}

\begin{assumption}
	[Bounded lengths] The random function $f$ is bounded almost surely.
	\label{ass:bounded_lengths}
\end{assumption}

\begin{remark}
	[Finite input space] All of the results in this section hold for any input space $\mathbb{X}$. Nevertheless, for computational convenience, most of the examples in this section focus on the practical setting where the input space is finite and $|\mathbb{X}|< \infty$. This implies that the projected lengths can be computed exactly because the supremum reduces down to a maximum: $\ell_{\boldsymbol{\eta}, \boldsymbol{\lambda}}[Y_{\boldsymbol{\eta}, f}^* (\boldsymbol{\omega})] = \max_{\mathbf{x} \in \mathbb{X}} s_{\boldsymbol{\eta}, \boldsymbol{\lambda}}(f(\mathbf{x}, \boldsymbol{\omega}))$ for $\boldsymbol{\lambda} \in \mathcal{S}_+^{M-1}$. 
\end{remark}

\begin{remark}
	[Sensitivity to transformations] Most of the probabilistic concepts which we describe in this section depend on the choice of reference vector (\cref{rem:reference_vector}) and any nonlinear transformations of the objective space (\cref{rem:representation}). For example, in \cref{sec:expectation}, we define the expected Pareto front surface in the original objective space: $\mathbb{E}_{\boldsymbol{\omega}}[Y_{\boldsymbol{\eta}, f}^*(\boldsymbol{\omega})] \in \mathbb{Y}^*_{\boldsymbol{\eta}}$. But we could have quite easily defined the same expectation in the transformed space, before inverting it back to the original objective space: $\tau^{-1}(\mathbb{E}_{\boldsymbol{\omega}}[Y_{\tau(\boldsymbol{\eta}), \tau \circ f}^*(\boldsymbol{\omega})])$, where $\tau: \mathbb{R}^M \rightarrow \mathbb{R}^M$ is an invertible monotonically increasing function. Clearly these two quantities are not necessarily equal and therefore some care has to be taken when using one or the other. Note that this issue also arises for the scalar-valued setting as well.
	\label{rem:sensitive}
\end{remark}

\begin{figure}
	\includegraphics[width=1\linewidth]{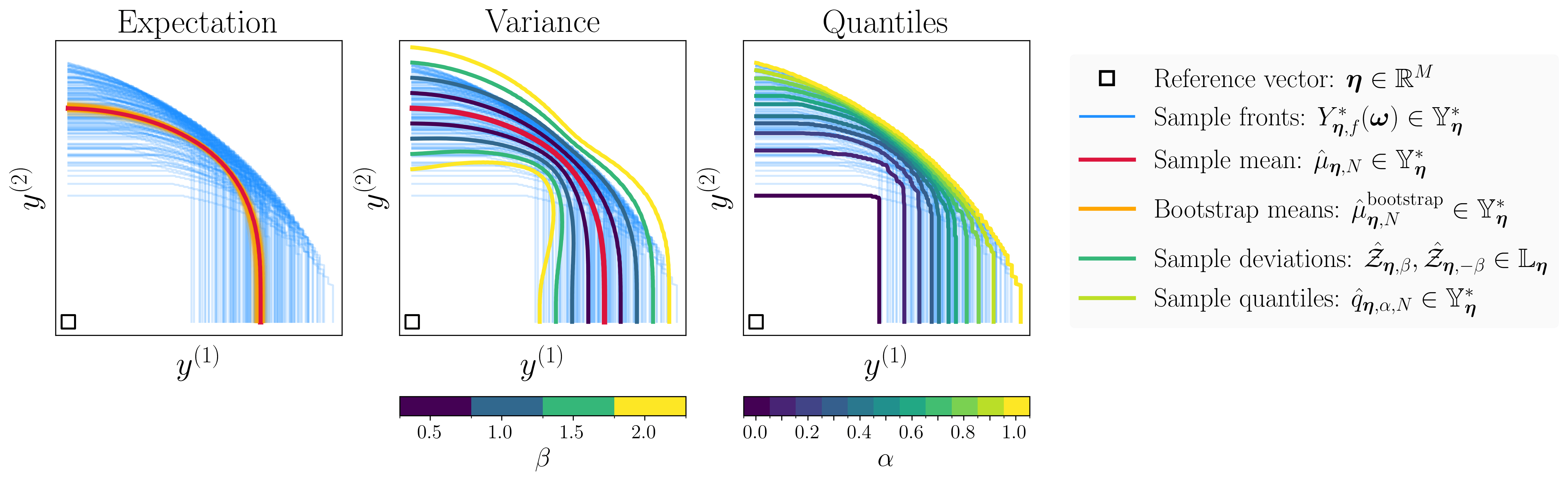}
	\centering
	\caption{An illustration of the estimated Pareto front surface statistics for a finite set of samples in $M=2$ dimensions.}
	\label{fig:pareto_front_statistics}
\end{figure}
\subsection{Expectation}
\label{sec:expectation}
We define the expected value of the Pareto front surface distribution \eqref{eqn:stochastic_pareto_front} by the equation
\begin{align}
	\mathbb{E}_{\boldsymbol{\omega}}[Y_{\boldsymbol{\eta}, f}^*(\boldsymbol{\omega})]
	:= \{\boldsymbol{\eta} + 
	\mathbb{E}_{\boldsymbol{\omega}}[
	\ell_{\boldsymbol{\eta}, \boldsymbol{\lambda}}[Y_{\boldsymbol{\eta}, f}^* (\boldsymbol{\omega})]
	]
	\boldsymbol{\lambda} \in \mathbb{R}^M
	: \boldsymbol{\lambda} \in \mathcal{S}_+^{M-1}\},
	\label{eqn:expectation}
\end{align}
where $\mathbb{E}_{\boldsymbol{\omega}}[\cdot]$ denotes the expectation operator under $\mathbb{P}$. It can be easily shown that under the given assumptions, the expectation above satisfies the conditions in \cref{prop:pareto_front_conditions} and therefore is a valid Pareto front surface. For completeness, we state this result in \cref{prop:expected_front} and prove it in \cref{app:proofs:prop:expected_front}.

\begin{proposition}
	Under \cref{ass:positive_lengths,ass:bounded_lengths}, the expectation of a Pareto front surface distribution \eqref{eqn:expectation} is a Pareto front surface $\mathbb{E}_{\boldsymbol{\omega}}[Y_{\boldsymbol{\eta}, f}^*(\boldsymbol{\omega})] \in \mathbb{Y}^*_{\boldsymbol{\eta}}$.
	\label{prop:expected_front}
\end{proposition}

\paragraph{Sample-based estimate} In general, the expected projected lengths
is an intractable quantity that cannot be computed exactly. Nevertheless, if we can sample a collection of random Pareto front surfaces, then we can easily estimate this quantity using a sample average
\begin{equation}
	\hat{\mu}_{\boldsymbol{\eta}, \boldsymbol{\lambda}, N} = \frac{1}{N} \sum_{n=1}^N 
	\ell_{\boldsymbol{\eta}, \boldsymbol{\lambda}}[Y_{\boldsymbol{\eta}, f}^* (\boldsymbol{\omega}_n)]
	\label{eqn:mean_length}
\end{equation}
for $\boldsymbol{\lambda} \in \mathcal{S}_+^{M-1}$, where $\{\boldsymbol{\omega}_n\}_{n=1}^N \subset \Omega$ denotes i.i.d. samples of the random parameter. Consequently, we can then define the sample mean Pareto front surface $\hat{\mu}_{\boldsymbol{\eta}, N} \in \mathbb{Y}^*_{\boldsymbol{\eta}}$ using the projected length estimates given by \eqref{eqn:mean_length}. The fact that this estimated front is also a valid Pareto front surface is a direct consequence of \cref{prop:expected_front}. We illustrate an example of the estimated mean Pareto front surface in the left plot of \cref{fig:pareto_front_statistics}.

\begin{example}
	[Bayesian bootstrap] To quantify some uncertainty about the mean estimate, we can appeal to the use many traditional techniques from scalar-valued statistics. For example, we can construct a Bayesian bootstrap \citep{rubin1981as} of the estimated means
	\begin{equation*}
		\hat{\mu}^{\textnormal{bootstrap}}_{\boldsymbol{\eta}, \boldsymbol{\lambda}, N}(\mathbf{w}) 
		= \sum_{n=1}^N w^{(n)}
		\ell_{\boldsymbol{\eta}, \boldsymbol{\lambda}}[Y_{\boldsymbol{\eta}, f}^* (\boldsymbol{\omega}_n)],
	\end{equation*}
	where $\mathbf{w} = (w^{(1)},\dots,w^{(N)}) \sim \text{Uniform}(\Delta^{M-1})$ denotes a weight vector sampled from a uniform distribution over the probability simplex $\Delta^{M-1} := \{\mathbf{w} \in \mathbb{R}^M_{\geq 0}: ||\mathbf{w}||_{L^1}=1\}$. We demonstrate an example of these bootstrap estimates in the left plot of \cref{fig:pareto_front_statistics}. Note that a similar type of bootstrap estimate can also be computed for the other Pareto front surface statistics which we will introduce in the upcoming sections. 
\end{example}

\subsection{Covariance}
\label{sec:covariance}
We define the covariance of the Pareto front surface distribution \eqref{eqn:stochastic_pareto_front} as the corresponding collection of covariance matrices with
\begin{align}
	\begin{split}
		\text{Cov}_{\boldsymbol{\omega}}
		[
		Y_{\boldsymbol{\eta}, f}^*(\boldsymbol{\omega}), 
		Y_{\boldsymbol{\eta}, f}^*(\boldsymbol{\omega})
		]_{\boldsymbol{\lambda}_i,\boldsymbol{\lambda}_j}
		&:= \boldsymbol{\lambda}_i \boldsymbol{\lambda}_j^T 
		\text{Cov}_{\boldsymbol{\omega}}[
		\ell_{\boldsymbol{\eta}, \boldsymbol{\lambda}_i}[Y_{\boldsymbol{\eta}, f}^* (\boldsymbol{\omega})],
		\ell_{\boldsymbol{\eta}, \boldsymbol{\lambda}_j}[Y_{\boldsymbol{\eta}, f}^* (\boldsymbol{\omega})]
		] \in \mathbb{R}^{M \times M}
	\end{split}
	\label{eqn:covariance}
\end{align}
for any two positive unit vectors $\boldsymbol{\lambda}_i, \boldsymbol{\lambda}_j \in \mathcal{S}_+^{M-1}$, where $\text{Cov}_{\boldsymbol{\omega}}[\cdot,\cdot]$ denotes the covariance operator under $\mathbb{P}$.

\paragraph{Sample-based estimate} In general, the covariance of the projected lengths
is an intractable quantity that cannot be computed exactly. Nevertheless, if we can sample a collection of random Pareto front surfaces, then we can easily estimate these terms by using the standard sample-based estimate
\begin{align*}
	\hat{\sigma}^2_{\boldsymbol{\eta}, \boldsymbol{\lambda}_i, \boldsymbol{\lambda}_i, N}
	= \frac{1}{N-1} \sum_{n=1}^N 
	(\ell_{\boldsymbol{\eta}, \boldsymbol{\lambda}_i}[Y_{\boldsymbol{\eta}, f}^* (\boldsymbol{\omega}_n)]
	- \hat{\mu}_{\boldsymbol{\eta}, \boldsymbol{\lambda}_i, N})
	(\ell_{\boldsymbol{\eta}, \boldsymbol{\lambda}_j}[Y_{\boldsymbol{\eta}, f}^* (\boldsymbol{\omega}_n)]
	- \hat{\mu}_{\boldsymbol{\eta}, \boldsymbol{\lambda}_j, N})
\end{align*}
for any two positive unit vectors $\boldsymbol{\lambda}_i, \boldsymbol{\lambda}_j \in \mathcal{S}_+^{M-1}$, where $\{\boldsymbol{\omega}_n\}_{n=1}^N \subset \Omega$ denotes i.i.d. samples of the random parameter with $N>1$.

\paragraph{Marginal deviation surfaces} The marginal variances can be used in order to define an uncertainty region around the mean. That is, we can define the upper deviation surface $\mathcal{Z}_{\boldsymbol{\eta}, \beta}$ and lower deviation surface $\mathcal{Z}_{\boldsymbol{\eta}, -\beta}$ for some $\beta \geq 0$ by
\begin{equation*}
	\mathcal{Z}_{\boldsymbol{\eta}, \beta} := 
	\{\boldsymbol{\eta} + (\mu_{\boldsymbol{\eta}, \boldsymbol{\lambda}} + \beta \sigma_{\boldsymbol{\eta}, \boldsymbol{\lambda}, \boldsymbol{\lambda}})_+ \boldsymbol{\lambda} \in \mathbb{R}^M
	: \boldsymbol{\lambda} \in \mathcal{S}_+^{M-1}\} \in \mathbb{L}_{\boldsymbol{\eta}}
	\label{eqn:deviation_surfaces}
\end{equation*}
where $\sigma_{\boldsymbol{\eta}, \boldsymbol{\lambda}, \boldsymbol{\lambda}} := (\sigma^2_{\boldsymbol{\eta}, \boldsymbol{\lambda}, \boldsymbol{\lambda}})^{1/2}$ denotes the marginal standard deviation in the direction $\boldsymbol{\lambda} \in \mathcal{S}^{M-1}_+$ and $(x)_+ := \max(x, 0)$, for $x \in \mathbb{R}$, denotes the truncation function which is used to ensure that the lengths are non-negative. Note that these polar surfaces are not necessarily Pareto front surfaces because they do not necessarily satisfy the maximum ratio condition in \cref{prop:pareto_front_conditions}. We present an illustration of these deviation surfaces in the middle plot of \cref{fig:pareto_front_statistics} for a simple two-dimensional problem.
\subsection{Quantiles}
\label{sec:quantiles}
We also define the quantiles of the Pareto front surface distribution \eqref{eqn:stochastic_pareto_front} as the collection of marginal quantiles
\begin{align}
	\mathcal{Q}_{\boldsymbol{\omega}}[Y_{\boldsymbol{\eta}, f}^*(\boldsymbol{\omega}), \alpha]
	&:= \{\boldsymbol{\eta} + 
	\mathcal{Q}_{\boldsymbol{\omega}}[
	\ell_{\boldsymbol{\eta}, \boldsymbol{\lambda}}[Y_{\boldsymbol{\eta}, f}^* (\boldsymbol{\omega})], \alpha]
	\boldsymbol{\lambda} \in \mathbb{R}^M
	: \boldsymbol{\lambda} \in \mathcal{S}_+^{M-1}\},
	\label{eqn:quantile}
\end{align}
where $\mathcal{Q}_{\boldsymbol{\omega}}[\cdot, \alpha]$ denotes the the $\alpha$-level quantile operator under $\mathbb{P}$ for any $\alpha \in (0, 1)$. It can be easily shown that under the given assumptions, the quantiles above satisfies the conditions in \cref{prop:pareto_front_conditions} and therefore is a valid Pareto front surface. For completeness, we state this result in \cref{prop:quantile_front} and prove it in \cref{app:proofs:prop:quantile_front}.

\begin{proposition}
	Under \cref{ass:positive_lengths,ass:bounded_lengths}, for any $\alpha \in (0, 1)$, the $\alpha$-quantile of a Pareto front surface distribution \eqref{eqn:quantile} is a Pareto front surface $\mathcal{Q}_{\boldsymbol{\omega}}[Y_{\boldsymbol{\eta}, f}^*(\boldsymbol{\omega}), \alpha] \in \mathbb{Y}^*_{\boldsymbol{\eta}}$.
	\label{prop:quantile_front}
\end{proposition}

\paragraph{Interpretation} Note that the $\alpha$-level quantile \eqref{eqn:quantile} does not imply that 100$\alpha$\% of the possible Pareto front surfaces are weakly dominated by this quantile Pareto front surface. Instead, this definition of the quantile can be interpreted as a surface which divides the truncated objective space $\mathbb{D}_{\succsucc}[\{\boldsymbol{\eta}\}]$ into two regions. One region is the space dominated by the $\alpha$-level quantile, which contains at least $\alpha$-probability. Whilst the other region, its complement, contains at most ($1-\alpha$)-probability. This interpretation follows immediately from the alternative formulation of the $\alpha$-quantiles described below in \eqref{eqn:quantile_equivalence}, which are based on the concept of domination probabilities (\cref{sec:domination_probabilities}).

\paragraph{Sample-based estimates} In general, the quantile of the projected lengths
is an intractable quantity that cannot be computed exactly. Nevertheless, if we can sample a collection of random Pareto front surfaces, then we can easily estimate these quantiles by using an empirical estimate
\begin{equation}
	\hat{q}_{\boldsymbol{\eta}, \boldsymbol{\lambda}, \alpha, N}
	:= \mathcal{Q}_{\boldsymbol{\omega} \sim \text{Uniform}(\{\boldsymbol{\omega}_n\}_{n=1}^N)}[
	\ell_{\boldsymbol{\eta}, \boldsymbol{\lambda}}[Y_{\boldsymbol{\eta}, f}^* (\boldsymbol{\omega})], \alpha
	]
	\label{eqn:quantile_length}
\end{equation}
for $\boldsymbol{\lambda} \in \mathcal{S}_+^{M-1}$, where $\{\boldsymbol{\omega}_n\}_{n=1}^N \subset \Omega$ denotes i.i.d. samples of the random parameter. Consequently, we can then define the empirical quantile Pareto front surface $\hat{q}_{\boldsymbol{\eta}, \alpha, N} \in \mathbb{Y}^*_{\boldsymbol{\eta}}$ using the projected lengths estimates given by \eqref{eqn:quantile_length}. By \cref{prop:quantile_front}, this empirical Pareto front surface is indeed a valid Pareto front surface. We illustrate an example of this empirical quantile front in the right plot of \cref{fig:pareto_front_statistics}.


\subsection{Probability of domination}
\label{sec:domination_probabilities}
The probability that a set of vectors is Pareto optimal is given by the probability of domination. This probability can also be interpreted as a generalisation of the survival function for Pareto front surface distributions. By \cref{lemma:domination_equivalence}, we can write the probability of domination in term of the projected lengths
\begin{align}
	\begin{split}
	\mathbb{P}[Y \subseteq \mathbb{D}_{\preceq, \boldsymbol{\eta}}[Y^*_{\boldsymbol{\eta}, f}]]
	&= \mathbb{P}[
	\land_{\mathbf{y} \in Y} 
	(s_{\boldsymbol{\eta}, \boldsymbol{\lambda}_{\boldsymbol{\eta}}^*(\mathbf{y})}(\mathbf{y}) 
	\leq 
	\ell_{\boldsymbol{\eta}, \boldsymbol{\lambda}_{\boldsymbol{\eta}}^*(\mathbf{y})}[Y^*_{\boldsymbol{\eta}, f}])
	]
	\\
	&= \mathbb{P}\biggl[
	\inf_{\mathbf{y} \in Y} 
	\frac{\ell_{\boldsymbol{\eta}, \boldsymbol{\lambda}_{\boldsymbol{\eta}}^*(\mathbf{y})}[Y^*_{\boldsymbol{\eta}, f}]}
	{||\mathbf{y}-\boldsymbol{\eta}||_{L^2}}
	\geq 1
	\biggr],
	\end{split}
	\label{eqn:domination_probability_y}
\end{align}
for any set of vectors $Y \subseteq \mathbb{D}_{\succsucc}[\{\boldsymbol{\eta}\}]$, where $\land$ denotes the logical \textsc{and} operation. In practice, this probability is an intractable quantity. Nevertheless, if we can sample a collection of random Pareto front surfaces, then we can easily estimate it by using a simple Monte Carlo average. In the left plot of \cref{fig:pareto_front_statistics_vorobev}, we visualise the contours associated with the estimated marginal domination probabilities for the Pareto front surface distribution described in \cref{fig:pareto_front_statistics}.

\subsection{Probability of deviation} 
\label{sec:deviation_probabilities}
In some cases, we might be interested in computing the probability that a set of vectors lies between two potentially random Pareto front surfaces. We refer to this quantity as the probability of deviation. By using \cref{lemma:domination_equivalence}, we can write the probability of deviation in term of the projected lengths
\begin{align}
	\begin{split}
		&\mathbb{P}[Y \subseteq
		(\mathbb{D}_{\preceq, \boldsymbol{\eta}}[A^*]
		\triangle 
		\mathbb{D}_{\preceq, \boldsymbol{\eta}}[B^*]) \setminus 
		\mathcal{Y}^{\textnormal{int}}_{\boldsymbol{\eta}}[A^* \cup B^*]
		]
		\\
		&= \mathbb{P}\biggl[\land_{\mathbf{y} \in Y}\Bigl(
		\min_{Y^* \in \{A^*, B^*\}}
		\ell_{\boldsymbol{\eta}, \boldsymbol{\lambda}_{\boldsymbol{\eta}}^*(\mathbf{y})}[Y^*]
		< s_{\boldsymbol{\eta}, \boldsymbol{\lambda}_{\boldsymbol{\eta}}^*(\mathbf{y})}(\mathbf{y})
		<
		\max_{Y^* \in \{A^*, B^*\}} 
		\ell_{\boldsymbol{\eta}, \boldsymbol{\lambda}_{\boldsymbol{\eta}}^*(\mathbf{y})}[Y^*]
		\Bigr)
		\biggr]
		\\
		&= \mathbb{P}\biggl[\sup_{\mathbf{y} \in Y}
		\biggl(\frac{\ell_{\boldsymbol{\eta}, \boldsymbol{\lambda}_{\boldsymbol{\eta}}^*(\mathbf{y})}[A^*]}{||\mathbf{y}-\boldsymbol{\eta}||_{L^2}} - 1 \biggr)
		\biggl(\frac{\ell_{\boldsymbol{\eta}, \boldsymbol{\lambda}_{\boldsymbol{\eta}}^*(\mathbf{y})}[B^*]}{||\mathbf{y}-\boldsymbol{\eta}||_{L^2}} - 1 \biggr)
		< 0
		\biggr],
	\end{split}
	\label{eqn:deviation_probability_y}
\end{align}
for any random Pareto front surfaces $A^*(\boldsymbol{\omega}), B^*(\boldsymbol{\omega}) \in \mathbb{Y}^*_{\boldsymbol{\eta}}$ and any set vectors $Y \subseteq \mathbb{D}_{\succsucc}[\{\boldsymbol{\eta}\}]$. As with the domination probabilities in \eqref{eqn:domination_probability_y}, this quantity can be easily estimated using a Monte Carlo average. In \cref{fig:pareto_front_statistics_vorobev}, we give an illustration of the estimated marginal deviation probabilities when $A^*(\boldsymbol{\omega}) = Y^*_{\boldsymbol{\eta}, f}(\boldsymbol{\omega})$ is a random Pareto front surface and $B^*(\boldsymbol{\omega})$ is a candidate for the expectation.
\subsection{Vorob'ev statistics}
\label{sec:vorobev}
Random set theory \citep{molchanov2005} extends standard concepts from Probability and Statistics to work over the space of closed sets. Some key ideas from this field have been used in earlier work in order to define some Pareto front surface statistics \citep{grunertdafonseca2010emaoa,binois2015ejoor}. These works have focussed on defining probabilistic concepts over the space of domination regions $\mathcal{D}_{\preceq} := \{\mathbb{D}_{\preceq}[A] \subset \mathbb{R}^M: A \subset \mathbb{R}^M\}$. In the following, we will review the core ideas within these works and show how they relate to our framework. To increase the generality of our discussion, we will consider working in the more general space of truncated domination regions $\mathcal{D}_{\preceq, \boldsymbol{\eta}} := \{\mathbb{D}_{\preceq, \boldsymbol{\eta}}[A] \subset \mathbb{R}^M: A \subset \mathbb{R}^M\}$. 

\paragraph{Coverage function} In random set theory, one often relies on the concept of a coverage function, which gives us the probability that an element lies within a random closed set. When working over the space of truncated domination regions $\mathcal{D}_{\preceq, \boldsymbol{\eta}}$, this coverage function is simply given by the probability of domination \eqref{eqn:domination_probability_y}: $\mathbb{P}[\mathbf{y} \in \mathbb{D}_{\preceq, \boldsymbol{\eta}}[Y^*_{\boldsymbol{\eta}, f}]]$ for any vector $\mathbf{y} \in \mathbb{R}^M$. Equipped with this coverage function, we can now define many different types of probabilistic concepts by leveraging ideas from random set theory. Notably, \cite{grunertdafonseca2010emaoa} and \cite{binois2015ejoor} have primarily focussed on extending the Vorob'ev concepts of random set theory \cite[Section 2.2]{molchanov2005}, which we describe below.

\paragraph{Vorob'ev quantiles} The Vorob'ev $\alpha$-quantile is a possible generalisation of the quantile function which is defined using the coverage function. Mathematically, the Vorob'ev $\alpha$-quantile is defined as the $\alpha$-level excursion set associated with the coverage function
\begin{align}
	\begin{split}
	\mathcal{Q}^{\text{Vorob'ev}}_{\boldsymbol{\omega}}[Y^*_{\boldsymbol{\eta}, f}(\boldsymbol{\omega}), \alpha]
	&:= \{\mathbf{y} \in \mathbb{R}^M: 
	\mathbb{P}[\mathbf{y} \in \mathbb{D}_{\preceq, \boldsymbol{\eta}}[Y^*_{\boldsymbol{\eta}, f}]] \geq \alpha
	\}
	\end{split}
	\label{eqn:quantile_vorobev}
\end{align}
for $\alpha \in (0, 1)$. By \cref{lemma:domination_equivalence}, we see that the Vorob'ev $\alpha$-quantile \eqref{eqn:quantile_vorobev} is equivalent to the truncated domination region of the $(1-\alpha)$-quantile front in \eqref{eqn:quantile}, that is
\begin{equation}
	\mathcal{Q}^{\text{Vorob'ev}}_{\boldsymbol{\omega}}[Y^*_{\boldsymbol{\eta}, f}(\boldsymbol{\omega}), \alpha]
	= \mathbb{D}_{\preceq, \boldsymbol{\eta}}[\mathcal{Q}_{\boldsymbol{\omega}}[Y^*_{\boldsymbol{\eta}, f}(\boldsymbol{\omega}), 1-\alpha]].
	\label{eqn:quantile_equivalence}
\end{equation}
In other words, the $\alpha$-level quantile in \eqref{eqn:quantile} is equal to the Pareto front surface or the upper isoline of the $(1-\alpha)$-level Vorob'ev quantile. Note that in the original work by \cite{grunertdafonseca2010emaoa}, they worked in the space of domination regions $\mathcal{D}_{\preceq}$ and therefore their result does not depend on the reference vector. Crudely speaking, we can recover this special case by letting the reference vector tend to negative infinity: $\boldsymbol{\eta} = (\epsilon, \dots, \epsilon) \in \mathbb{R}^M$ with $\epsilon \rightarrow -\infty$.

\paragraph{Vorob'ev mean} In addition to proposing the Vorob'ev $\alpha$-quantile, \cite{grunertdafonseca2010emaoa} proposed using the median quantile ($\alpha=0.5$) as a potential candidate for the mean Pareto front surface. In contrast \cite{binois2015ejoor} proposed using the Vorob'ev expectation as a candidate for the mean. The Vorob'ev expectation is defined as the $\alpha^*$-level Vorob'ev quantile whose hypervolume is the closest to the expected hypervolume of the random Pareto front surface. More precisely,
\begin{equation*}
	\mathbb{E}^{\text{Vorob'ev}}_{\boldsymbol{\omega}}[Y^*_{\boldsymbol{\eta}, f}(\boldsymbol{\omega})] := \mathcal{Q}_{\boldsymbol{\omega}}^{\text{Vorob'ev}}[Y^*_{\boldsymbol{\eta}, f}(\boldsymbol{\omega}), \alpha^*],
\end{equation*}
where $\alpha^*$ satisfies the inequality
\begin{equation}
	\nu[\mathcal{Q}^{\text{Vorob'ev}}_{\boldsymbol{\omega}}[
	Y^*_{\boldsymbol{\eta}, f}(\boldsymbol{\omega}), \alpha]
	]
	\leq 
	V_{\boldsymbol{\eta}, f}
	\leq \nu[\mathcal{Q}^{\text{Vorob'ev}}_{\boldsymbol{\omega}}[
	Y^*_{\boldsymbol{\eta}, f}(\boldsymbol{\omega}), \alpha^*]
	],
	\label{eqn:alpha_star}
\end{equation}
for any $\alpha > \alpha^*$, where $V_{\boldsymbol{\eta}, f} := \mathbb{E}_{\boldsymbol{\omega}} [\nu[\mathbb{D}_{\preceq, \boldsymbol{\eta}}[Y^*_{\boldsymbol{\eta}, f}(\boldsymbol{\omega})]] ] = \mathbb{E}_{\boldsymbol{\omega}} [U^{\text{HV}}_{\boldsymbol{\eta}}[Y^*_{\boldsymbol{\eta}, f}(\boldsymbol{\omega})] ]$ is the expected hypervolume of the Pareto front surface distribution. Note that the Vorob'ev mean is a Vorob'ev quantile and therefore it is a truncated domination region and not a Pareto front surface. As a result, when we later refer to the Vorob'ev mean front, we are actually referring to the Pareto front surface of the Vorob'ev mean, that is the $(1-\alpha^*)$-quantile front.

\paragraph{Vorob'ev deviation} Analogous to how the traditional scalar-valued expectation minimises the variance, the Vorob'ev mean is known to minimise a quantity known as the Vorob'ev deviation. \cite{binois2015ejoor} defined the Vorob'ev deviation of a Pareto front surface distribution as the expected hypervolume of the symmetric difference between the Pareto front surface and the considered set. Equivalently, the quantity can be written as the expected hypervolume distance \eqref{eqn:hypervolume_distance} between these two sets, that is
\begin{align}
	\textsc{Vorob'evDeviation}_{\boldsymbol{\eta}, f}[A] 
	&:=
	\mathbb{E}_{\boldsymbol{\omega}}
	[
	\nu [A \triangle \mathbb{D}_{\preceq, \boldsymbol{\eta}}[Y^*_{\boldsymbol{\eta}, f}(\boldsymbol{\omega})]]
	]
	\label{eqn:vorobev_deviation}
\end{align}
for any measurable set $A \in \mathcal{M}(\mathbb{D}_{\succsucc}[\{\boldsymbol{\eta}\}])$ lying in the truncated space. As shown by \citet[Section 2.2]{molchanov2005}, the Vorob'ev mean is a minimiser to the Vorob'ev deviation over the space of measurable sets $A \in \mathcal{M}(\mathbb{D}_{\succsucc}[\{\boldsymbol{\eta}\}])$ subject to the condition that $\nu[A] = V_{\boldsymbol{\eta}, f}$. That is,
\begin{equation*}
	\mathbb{E}^{\text{Vorob'ev}}_{\boldsymbol{\omega}}[Y^*_{\boldsymbol{\eta}, f}(\boldsymbol{\omega})]
	\in \argmin_{A \in \mathcal{M}(\mathbb{D}_{\succsucc}[\{\boldsymbol{\eta}\}]): \nu[A]=V_{\boldsymbol{\eta}, f}} \textsc{Vorob'evDeviation}_{\boldsymbol{\eta}, f}[A].
	\label{eqn:vorobev_mean}
\end{equation*}

\paragraph{Validity of the Pareto front surfaces} As a direct consequence of \cref{prop:quantile_front}, the positive isolines of the Vorob'ev based Pareto front surfaces are all valid Pareto front surfaces because they are just instances of the quantile front \eqref{eqn:quantile}. This result was not proven and just taken for granted in the original works, but we have shown that it follows quite naturally from our polar parameterisation and \cref{prop:quantile_front}.

\begin{figure}
	\includegraphics[width=0.8\linewidth]{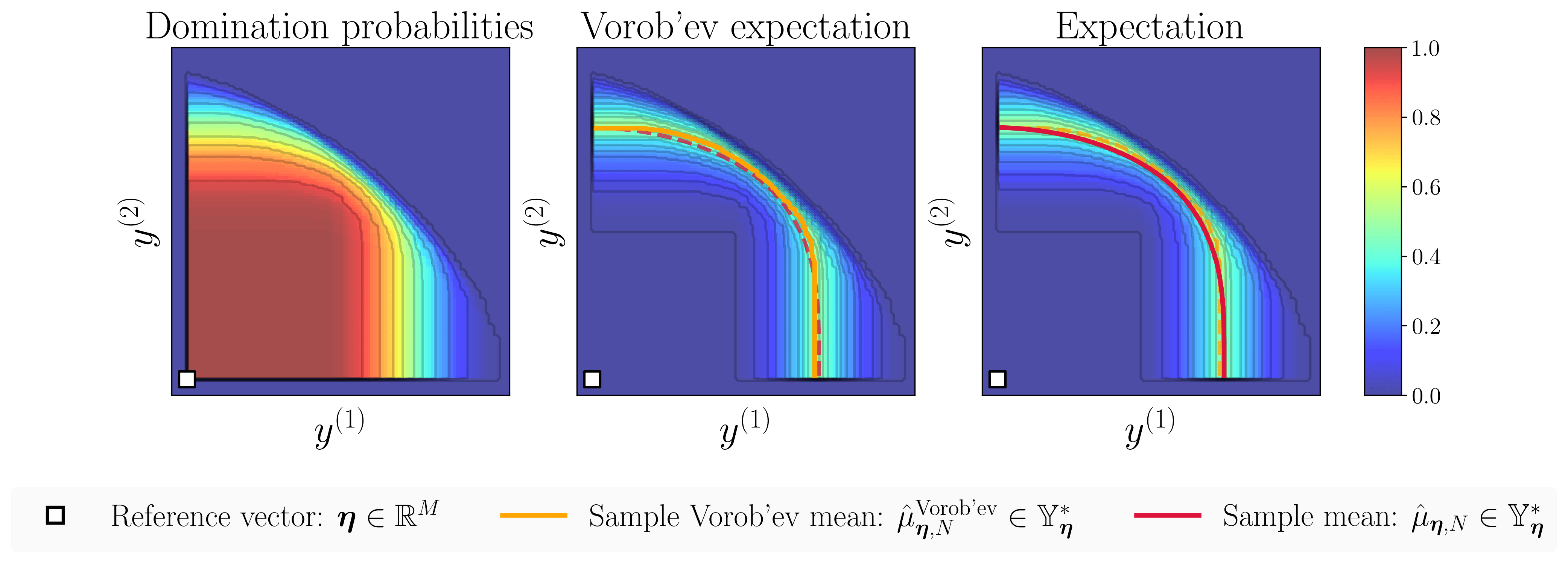}
	\centering
	\caption{An illustration of the Vorob'ev statistics on an $M=2$ dimensional example. On the left plot, we illustrate the contours of the estimated marginal probability of domination \eqref{eqn:domination_probability_y}. 
	In the middle and right plot, we illustrate the estimated Vorob'ev mean front and expected front, respectively. For both of these fronts, we include the contours of the estimated marginal probability of deviation \eqref{eqn:deviation_probability_y} between the random Pareto front surface and the mean estimate.}
	\label{fig:pareto_front_statistics_vorobev}
\end{figure}

\paragraph{Sample-based estimates} The Vorob'ev statistics above can be estimated using the empirical quantiles as we described in \cref{sec:quantiles}. For the Vorob'ev mean front, we have to additionally compute the parameter $\alpha^*$ which satisfies the hypervolume condition \eqref{eqn:alpha_star}. To approximately solve this problem, \cite{binois2015ejoor} proposed using an iterative bisection scheme. Notably, this procedure can be very expensive because it requires evaluating the hypervolume of the quantiles several times. On the other hand, our definition of the expectation is much cheaper and simpler to estimate. In \cref{fig:pareto_front_statistics_vorobev}, we present a visual comparison between the estimated Vorob'ev mean front and our sample mean front for a simple two-dimensional example. In this example, there does not appear to be any meaningful difference between these two Pareto front surfaces.

\subsection{A decision-theoretic construction}
\label{sec:decision_theory}
Many standard statistics of real-valued random variables can be constructed under the decision-theoretic perspective in which we optimise a certain scoring rule \citep{gneiting2011jasa}. Our polar parameterisation of the Pareto front surface gives us an easy way to generalise this decision-theoretic construction to the space of Pareto front surface distributions. In particular, we can recover many of the Pareto front surface statistics introduced in the previous sections as being equal to a functional $F_{\boldsymbol{\omega}}$, which minimises some expected frontier loss \eqref{eqn:frontier_distance_function},
\begin{equation}
	F_{\boldsymbol{\omega}}[Y_{\boldsymbol{\eta}, f}^*(\boldsymbol{\omega})]
	\in \argmin_{A \in \mathbb{L}_{\boldsymbol{\eta}}} \mathbb{E}_{\boldsymbol{\omega}}[
	\mathcal{D}_{\boldsymbol{\eta}, S}[A, Y_{\boldsymbol{\eta}, f}^*(\boldsymbol{\omega})]
	],
	\label{eqn:decision_theory}
\end{equation}
where $S: \mathbb{R} \times \mathbb{R} \rightarrow \mathbb{R}_{\geq 0}$ denotes some scoring function. In \cref{tab:scoring_functions}, we present the explicit choices of $S$ which leads to the different Pareto front surface statistics that we introduced earlier. Note that all of the results in this table follow immediately from the standard construction of these functionals in the univariate setting. To see this, we first note that we can rewrite the expected frontier loss  as an expectation over the space of positive unit vectors by interchanging the order of integration, that is
\begin{equation*}
	\mathbb{E}_{\boldsymbol{\omega}}[
	\mathcal{D}_{\boldsymbol{\eta}, S}[A, Y_{\boldsymbol{\eta}, f}^*(\boldsymbol{\omega})]
	]
	= \mathbb{E}_{\boldsymbol{\lambda} \sim \text{Uniform}(\mathcal{S}^{M-1}_+)}[
	\mathbb{E}_{\boldsymbol{\omega}}[
	S(\ell_{\boldsymbol{\eta}, \boldsymbol{\lambda}}[A], \ell_{\boldsymbol{\eta}, \boldsymbol{\lambda}}[Y_{\boldsymbol{\eta}, f}^*(\boldsymbol{\omega})])]
	].
\end{equation*}
As there are no inter-dependencies between the different unit vectors in this expression, we see that the optimisation problem in \eqref{eqn:decision_theory} can equivalently be solved by independently solving for the projected lengths along each positive direction, namely
\begin{equation}
	\ell_{\boldsymbol{\eta}, \boldsymbol{\lambda}}[F_{\boldsymbol{\omega}}[Y_{\boldsymbol{\eta}, f}^*(\boldsymbol{\omega})]] 
	\in \argmin_{a \geq 0} \mathbb{E}_{\boldsymbol{\omega}}[
	S(a, \ell_{\boldsymbol{\eta}, \boldsymbol{\lambda}}[Y_{\boldsymbol{\eta}, f}^* (\boldsymbol{\omega})])
	]
	\label{eqn:decision_theory_scalar}
\end{equation}
for $\boldsymbol{\lambda} \in \mathcal{S}^{M-1}_+$. Evidently, once we have solved this collection of optimisation problems, we can easily reconstruct the polar surface $F_{\boldsymbol{\omega}}[Y_{\boldsymbol{\eta}, f}^*(\boldsymbol{\omega})] \in \mathbb{L}_{\boldsymbol{\eta}}$ by using the linear mapping described earlier in \eqref{eqn:polar_reconstruction}. In essence, the scalar-valued optimisation problems in \eqref{eqn:decision_theory_scalar} are equivalent to the standard decision-theoretic optimisation problems which appear in the univariate setting. Consequently, this means that we can just take advantage of existing results from univariate statistics in order to define Pareto front surface statistics. For instance, in the previous sections, we set the scoring function to be the squared error loss and the pinball loss in order to recover the expected value \citep{savage1971jasa} and the $\alpha$-quantiles \citep{raiffa1968,ferguson1967}, respectively. Naturally, we can also use this strategy to generalise any other standard univariate statistic of interest such as the mode, $\alpha$-expectiles and so on.

\begin{table*}[!htb]
	\centering
	\begin{tabular}{lll}
		\toprule
		Pareto front surface statistic & $F_{\boldsymbol{\omega}}[\cdot]$ & Scoring function: $S(x, y)$\\
		\midrule
		Mean front & $\mathbb{E}_{\boldsymbol{\omega}}[\cdot]$ & $(x-y)^2$ \\
		Quantile front &  $\mathcal{Q}_{\boldsymbol{\omega}}[\cdot, \alpha]$ & $(\mathbbm{1}[x \leq y] - \alpha)(x - y)$ \\
		Vorob'ev quantile front &  $\mathcal{Y}_{\boldsymbol{\eta}}^{\text{int}}[\mathcal{Q}^{\text{Vorob'ev}}_{\boldsymbol{\omega}}[\cdot, \alpha]]$ & $(\mathbbm{1}[x \leq y] - (1-\alpha))(x - y)$ \\
		Vorob'ev mean front &  $\mathcal{Y}_{\boldsymbol{\eta}}^{\text{int}}[\mathbb{E}^{\text{Vorob'ev}}_{\boldsymbol{\omega}}[\cdot]]$ & $(\mathbbm{1}[x \leq y] - (1-\alpha^*))(x - y)$\\
		\bottomrule
	\end{tabular}
	\bigskip
	\caption{A list of Pareto front surface statistics and their scoring functions \eqref{eqn:decision_theory}.}
	\label{tab:scoring_functions}
\end{table*}
\section{Applications}
\label{sec:applications}
The concepts and results which we describe in this work are general and can be applied in any scenario where we want to infer or quantify properties about a random Pareto front surface. In this section, we identify a few concrete applications of these ideas which might be of interest for practitioners. Firstly, we begin in \cref{sec:visualisation} by looking at the problem of Pareto front visualisation, which is a core component in many modern decision making workflows. We showcase how it is possible to adapt many existing visualisation strategies from the deterministic setting into the stochastic setting by appealing to our polar parameterisation result. Secondly, in \cref{sec:uncertainty_quantification}, we illustrate how all of these Pareto front surface statistics and visualisation ideas can be used within a standard Bayesian experimental design setting. Subsequently, in \cref{sec:extreme_value_theory}, we apply our framework to extend multivariate extreme value theory to the Pareto partially ordered setting. Lastly, in \cref{sec:air_pollution}, we present a Pareto front analysis of a real-world time series data set concerning the air pollution levels in North Kensington, London.
\subsection{Visualisation}
\label{sec:visualisation}
Visualisation is a key element in many modern multi-objective decision making pipelines. We can easily visualise the Pareto front surface for low-dimensional problems ($M \leq 3$) by using a simple two-dimensional line plot or a three-dimensional surface plot. For higher dimensional problems ($M > 3$), these simple approaches no longer work and we have to become more creative when it comes to visualising the Pareto front---see the work by \cite{tusar2015itec} for a survey on these visualisation strategies. Notably, this problem of visualising the Pareto front surface is exacerbated when we also want to include uncertainty information about the Pareto front surface distribution as well. The majority of existing work in Pareto front visualisation has largely focussed on the deterministic setting. To the best of our knowledge, no work has focussed on the practically important problem of visualising a random Pareto front surface. In this section, we address this gap in the literature by proposing a general visualisation approach which works even in the stochastic setting. We develop a projection mapping strategy which gives us a way to slice a high-dimensional Pareto front surface into a collection of lower-dimensional Pareto front surfaces. Conceptually, we build a picture of the overall Pareto front surface by navigating the space of low-dimensional slices. To illustrate the usefulness of our new method, we built two simple dashboard applications which can be used to visualise and navigate the space of one or two-dimensional Pareto front surface slices, respectively---see \cref{fig:projection_one,fig:projection_two} for a snapshot of these applications.
\subsubsection{Projected Pareto front surfaces}
\label{sec:projection}
The polar parameterisation result tells us that any $M$-dimensional Pareto front is isomorphic to the set of positive unit vectors $\mathcal{S}_+^{M-1}$. Therefore, in theory, any strategy that can be used to visualise the set of positive unit vectors can also be used to visualise a Pareto front surface. In this section, we propose a novel idea to visualise any Pareto front surface by appealing to the following partitioning of the space of positive unit vectors:
\begin{equation}
	\mathcal{S}_+^{M-1} 
	= \bigcup_{\mathbf{v} \in \mathcal{V}_+^{M-P}} \{\boldsymbol{\lambda} \in \mathcal{S}_+^{M-1}:
	\mathcal{P}_{[M] \setminus I}(\boldsymbol{\lambda}) = \mathbf{v}\}
	\label{eqn:directions_partition}
\end{equation}
where $I = \{i_1, \dots, i_{P}\} \subset [M] := \{1,\dots,M\}$ is a non-empty set of $|I| = P$ ordered indices with $i_1<\cdots<i_{P}$; $\mathcal{P}_{I}: \mathbb{R}^M \rightarrow \mathbb{R}^{P}$ is the projection mapping with $\mathcal{P}_I(\mathbf{a}) = (a^{(i_1)}, \dots, a^{(i_{P})}) \in \mathbb{R}^{P}$ for any vector $\mathbf{a} \in \mathbb{R}^M$; and $\mathcal{V}_+^{M-P} := \{\mathbf{z} \in \mathbb{R}^{M-P}_{>0}: ||\mathbf{z}||_{L^2} < 1\}$ is the set of vectors lying within the positive orthant of the $(M-P)$-dimensional sphere. Intuitively, the partition in \eqref{eqn:directions_partition} gives us a way to slice up the set of positive unit vectors by intersecting it with the hyperplanes $\{\mathbf{y} \in \mathbb{R}^M: \mathcal{P}_{[M] \setminus I}(\mathbf{y}) = \mathbf{v}\}$ for $\mathbf{v} \in \mathcal{V}_+^{M-P}$. By construction, if we project each slice onto the remaining indices, then we obtain a lower-dimensional Pareto front surface. We summarise this result in \cref{lemma:projected_slice} and prove it in \cref{app:proofs:lemma:projected_slice}. 
\begin{lemma}
	[Projected slice] For any non-empty set of $P>0$ indices $I \subset [M]$ and vector $\mathbf{v} \in \mathcal{V}_+^{M-P}$, the polar surface $\mathcal{P}_{I,\mathbf{v}}[\mathcal{S}^{M-1}_+] :=\{\mathcal{P}_I(\boldsymbol{\lambda}) \in \mathbb{R}^P: \boldsymbol{\lambda} \in \mathcal{S}_{+,\mathbf{v}}^{M-1}\} \in \mathbb{L}_{\mathbf{0}_P}$ is a $P$-dimensional Pareto front surface with the zero reference vector, that is $\mathcal{P}_{I,\mathbf{v}}[\mathcal{S}^{M-1}_+] \in \mathbb{Y}_{\mathbf{0}_P}^*$.
	\label{lemma:projected_slice}
\end{lemma}
Notably, this result gives us a way to visualise the set of positive unit vectors $\mathcal{S}_+^{M-1}$ through its lower-dimensional projections. In the following, we will extend this visualisation strategy to work for any arbitrary Pareto front surface $A^* \in \mathbb{Y}_{\boldsymbol{\eta}}^*$.

\begin{figure}
	\includegraphics[width=0.8\linewidth]{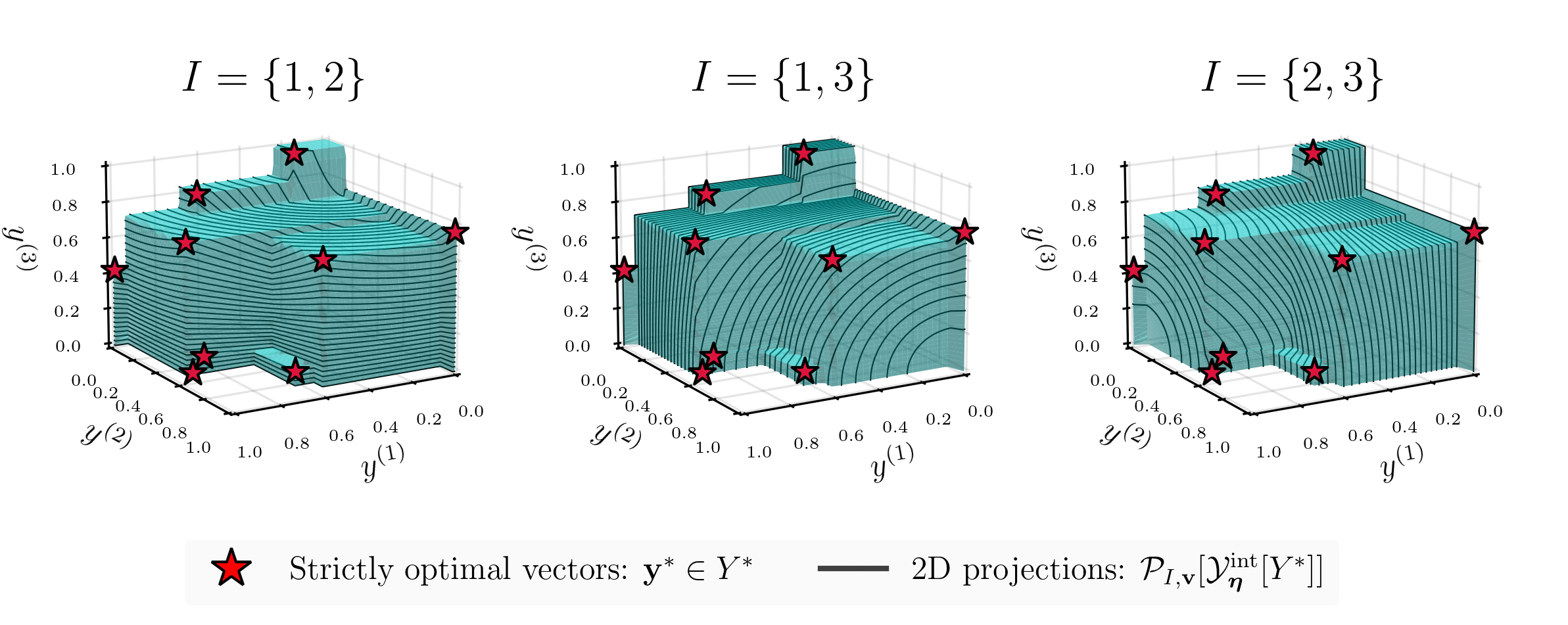}
	\centering
	\caption{An illustration of the three different ways that we can generate two-dimensional slices of a three-dimensional Pareto front surface.}
	\label{fig:pareto_front_slices}
\end{figure}

\paragraph{Projection mapping} We will now formalise the concept of the projected Pareto front surface, which is the natural extension to the ideas described above. To accomplish this, we define the reconstruction function $\phi_I: \mathcal{V}_+^{M-P} \times \mathcal{S}_+^{P-1} \rightarrow \mathcal{S}_+^{M-1}$, which creates a positive unit vector whose projected values are given by
\begin{align*}
	\mathcal{P}_I(\phi_I(\mathbf{v}, \boldsymbol{\lambda})) = \sqrt{1-||\mathbf{v}||_{L^2}^2} \boldsymbol{\lambda}
	\quad \text{ and } \quad 
	\mathcal{P}_{[M] \setminus I}(\phi_I(\mathbf{v}, \boldsymbol{\lambda})) = \mathbf{v},
\end{align*}
for any non-empty set of $P$ ordered indices $I = \{i_1,\dots,i_P\} \subset [M]$, any vector $\mathbf{v} \in \mathcal{V}_+^{M-P}$ and any positive unit vector $\boldsymbol{\lambda} \in \mathcal{S}_+^{P-1}$. Explicitly, the reconstruction function takes the form
\begin{equation}
	\phi_I^{(m)}(\mathbf{v}, \boldsymbol{\lambda})
	:=
	\begin{cases}
		\sqrt{1-||\mathbf{v}||_{L^2}^2} \lambda^{(p)}, & \text{if } m \in I \text{ and } m = i_p, \\
		v^{(p)}, & \text{if } m \in J \text{ and } m = j_p,
	\end{cases}
	\label{eqn:reconstruction_function}
\end{equation}
for $m = 1,\dots,M$, where $J = [M] \setminus I = \{j_1,\dots,j_{M-P}\}$ denotes the complement of $I$ containing $M-P$ ordered indices. Using this function, we define the projection mapping $\mathcal{P}_{I, \mathbf{v}}: \mathbb{L}_{\boldsymbol{\eta}} \rightarrow \mathbb{L}_{\mathcal{P}_I(\boldsymbol{\eta})}$ to satisfy the equation
\begin{align}
	\begin{split}
		\mathcal{P}_{I, \mathbf{v}}[A] 
		&:= \{\mathcal{P}_I(\boldsymbol{\eta}) + 
		\ell_{\boldsymbol{\eta}, \phi_I(\mathbf{v}, \boldsymbol{\lambda})}
		[A] 
		\sqrt{1-||\mathbf{v}||_{L^2}^2} \boldsymbol{\lambda} \in \mathbb{R}^P: 
		\boldsymbol{\lambda} \in \mathcal{S}_+^{P-1}\},
	\end{split}
	\label{eqn:projected_pareto_front}
\end{align}
for any polar surface $A \in \mathbb{L}_{\boldsymbol{\eta}}$. In \cref{prop:projected_pareto_front}, immediately below, we show that the projection of any Pareto front surface $A^* \in \mathbb{Y}^*_{\boldsymbol{\eta}}$, that is $\mathcal{P}_{I, \mathbf{v}}[A^*] \in \mathbb{L}_{\mathcal{P}_I(\boldsymbol{\eta})}$, is a valid $P$-dimensional Pareto front surface with the reference vector $\mathcal{P}_I(\boldsymbol{\eta}) \in \mathbb{R}^P$. 
\begin{proposition}
	[Projected Pareto front surface] For any Pareto front surface $A^* \in \mathbb{Y}_{\boldsymbol{\eta}}^*$, any non-empty set of $P>0$ indices $I \subset [M]$, any vector $\mathbf{v} \in \mathcal{V}_+^{M-P}$, the corresponding projected Pareto front surface \eqref{eqn:projected_pareto_front} is a $P$-dimensional Pareto front surface with the reference vector $\mathcal{P}_I(\boldsymbol{\eta}) \in \mathbb{R}^P$, that is $\mathcal{P}_{I, \mathbf{v}}[A^*] \in \mathbb{Y}^*_{\mathcal{P}_I(\boldsymbol{\eta})}$.
	\label{prop:projected_pareto_front}
\end{proposition}
This result follows similarly in spirit to \cref{lemma:projected_slice} and the proof is presented in \cref{app:proofs:prop:projected_pareto_front}. In addition, we have also included a concrete example in \cref{fig:pareto_front_slices}, where we demonstrate how a three-dimensional Pareto front surface can be partitioned into a collection of two-dimensional slices.

\begin{remark}
	[Non-constant fixed vector] The projected Pareto front surface $\mathcal{P}_{I, \mathbf{v}}[A^*] \in \mathbb{Y}^*_{\mathcal{P}_I(\boldsymbol{\eta})}$ only considers the values indexed by the set $I$. It ignores the values at the other components $[M] \setminus I$. The values at these components were a constant for the spherical Pareto front surface $\mathcal{S}_+^{M-1}$, but there are not necessarily a constant for a general Pareto front surface $A^* \in \mathbb{Y}_{\boldsymbol{\eta}}^*$. Specifically, the projected vector at these indices are given by
	\begin{equation*}
		\mathcal{P}_{[M] \setminus I}(\boldsymbol{\eta}) + \ell_{\boldsymbol{\eta}, \phi_I(\mathbf{v}, \boldsymbol{\lambda})}[A^*] \mathbf{v} \in \mathbb{R}^{M-P}
	\end{equation*}
	for any $\boldsymbol{\lambda} \in \mathcal{S}_+^{P-1}$. Note that this vector is constant if and only if the projected lengths $\ell_{\boldsymbol{\eta}, \phi_I(\mathbf{v}, \boldsymbol{\lambda})}[A^*]$ are constant for all $\boldsymbol{\lambda} \in \mathcal{S}_+^{P-1}$. If this is not the case, then one must also be aware of this feature when examining the slices.
	\label{rem:non_constant}
\end{remark}

\subsubsection{Dashboard applications}
\label{sec:dashboard_applications}
\begin{figure}
	\includegraphics[width=0.8\linewidth]{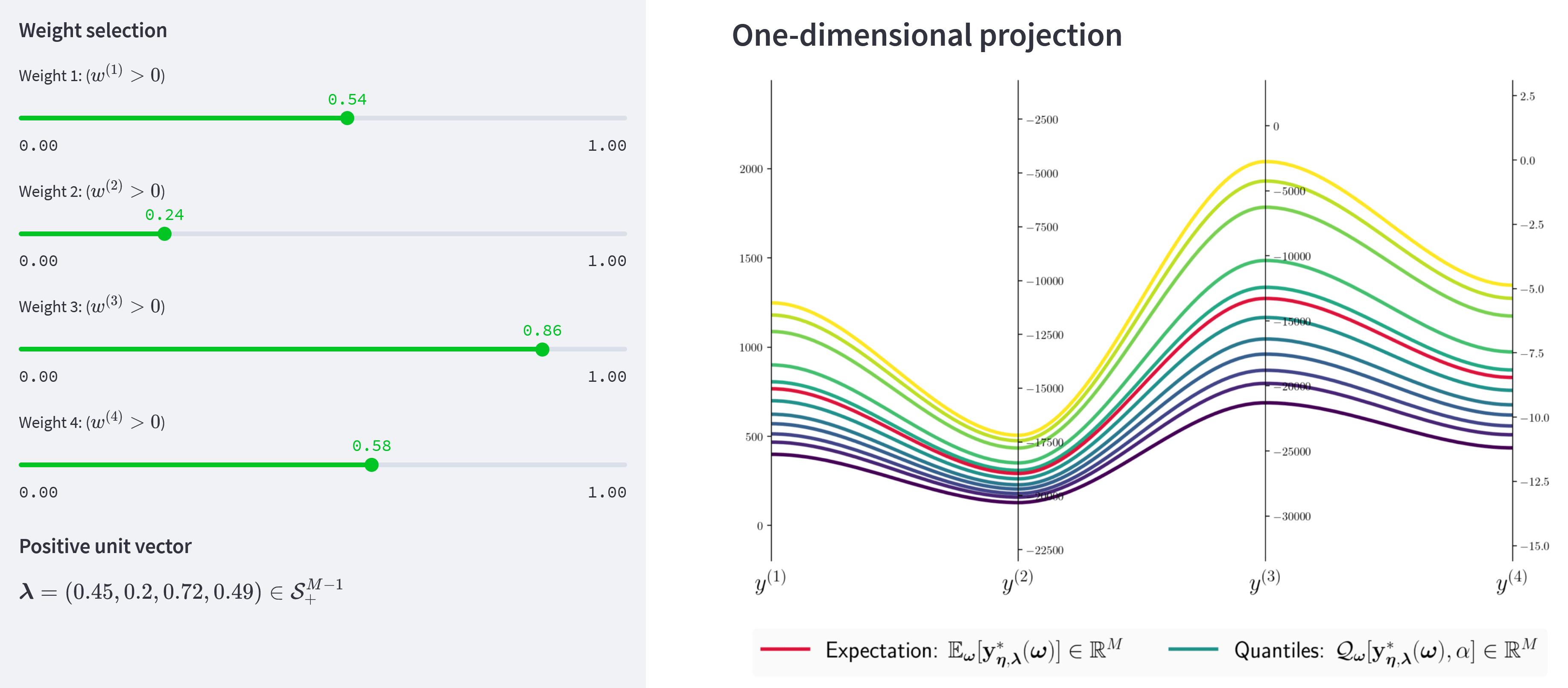}
	\centering
	\caption{An illustration of the dashboard application that we made in order to navigate the space of one-dimensional projections of a Pareto front surface.}
	\label{fig:projection_one}
\end{figure}

\begin{figure}
	\includegraphics[width=0.75\linewidth]{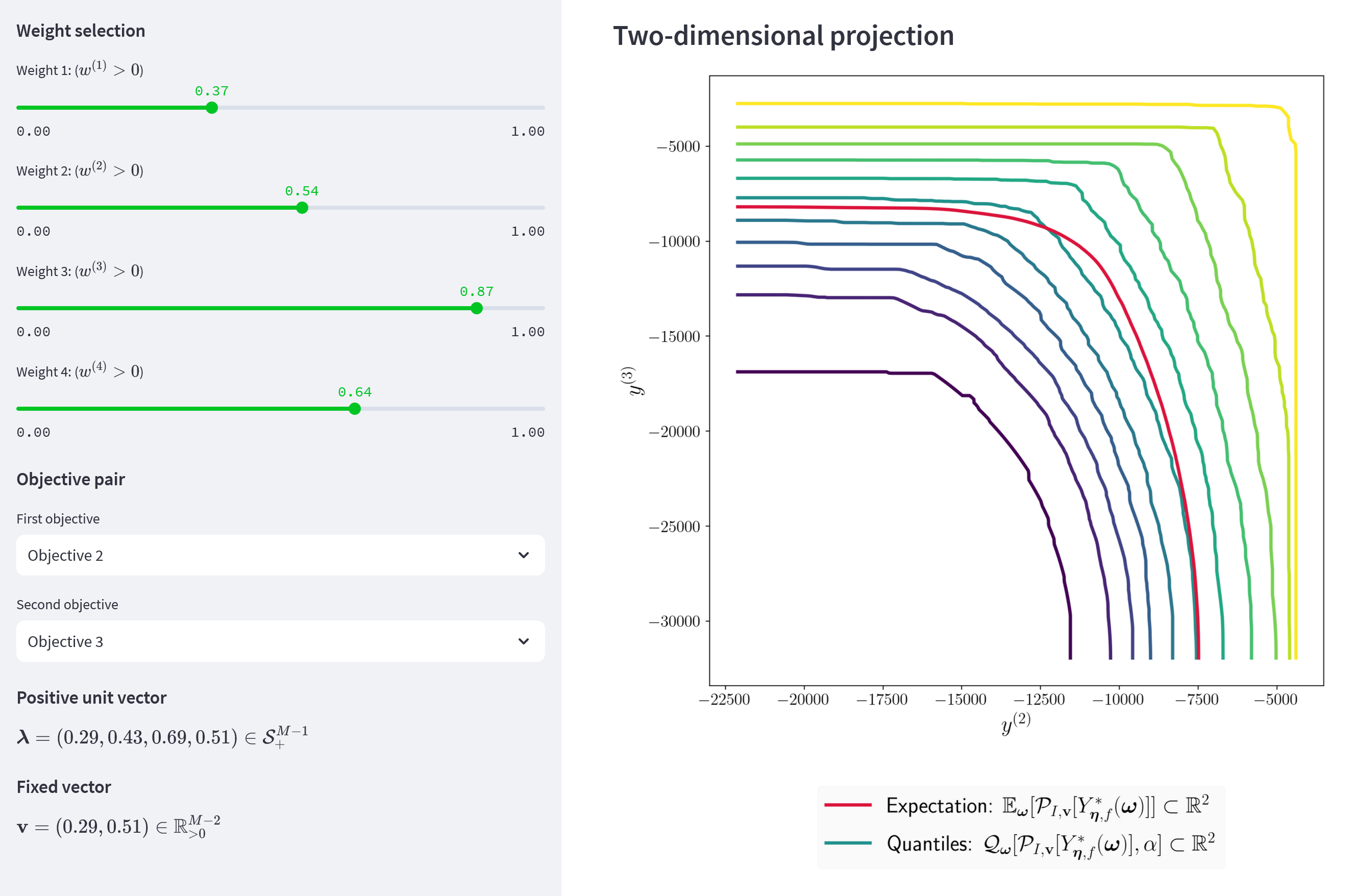}
	\centering
	\caption{An illustration of the dashboard application that we made in order to navigate the space of two-dimensional projections of a Pareto front surface.}
	\label{fig:projection_two}
\end{figure}

\cref{prop:projected_pareto_front} gives us a way to visualise an $M$-dimensional Pareto front surface distribution by navigating its lower-dimensional projections. To illustrate the utility and simplicity of this new concept, we have built two interactive dashboard applications which can be used to navigate the space of one or two dimensional slices, respectively. We present a snapshot of these applications in \cref{fig:projection_one} and \cref{fig:projection_two}, respectively, for a four-dimensional Pareto front surface distribution based on the reformulated bulk carrier design problem \citep{tanabe2020asc}. Below we elaborate on the construction of these applications in some more detail.

\paragraph{One-dimensional projection} To navigate the space of one-dimensional slices, we propose introducing $M$ different sliders that can be used to adjust the positive unit vectors $\boldsymbol{\lambda} \in \mathcal{S}^{M-1}_+$. Conceptually, each slider is associated with a weight lying in the open unit interval: $w^{(m)} \in (0, 1)$ for the objectives $m = 1,\dots,M$. The positive unit vector can then be constructed by normalising the weight vector appropriately: $\boldsymbol{\lambda} = \mathbf{w} / ||\mathbf{w}||_{L^2} \in \mathcal{S}_+^{M-1}$. 

Ideally, we want the weight vector $\mathbf{w} \in (0, 1)^M$ to denote the relative importance of each objective. For this reason, we apply a normalisation transformation $\tau: \mathbb{R}^M \rightarrow \mathbb{R}^M$ to the objective vectors: $\tau(\mathbf{y}) = (\mathbf{y} - \mathbf{l})/(\mathbf{u} - \mathbf{l})$ for any vector $\mathbf{y} \in \mathbb{R}^M$, where $\mathbf{l}, \mathbf{u} \in \mathbb{R}^M$ denote the estimates for the lower and upper bound of the objective vectors, respectively. In addition, we also use these bound estimates to set the reference vector\footnote{Following \cref{rem:reference_vector}, we set the reference vector to be slightly worse than the estimated nadir. Naturally, we could also allow the reference vector to be set dynamically by adding some more sliders into the application.}: $\boldsymbol{\eta} = \mathbf{l} - 0.2 (\mathbf{u} - \mathbf{l}) \in \mathbb{R}^M$. As described in \cref{rem:representation}, to reconstruct the Pareto front surface, we just have to invert this affine transformation, namely we set
\begin{align*}
	\mathbf{y}^*_{\boldsymbol{\eta}, \boldsymbol{\lambda}}
	&= \tau^{-1}( \tau(\boldsymbol{\eta}) + \ell_{\tau(\boldsymbol{\eta}), \boldsymbol{\lambda}}[\tau(A^*)] \boldsymbol{\lambda})
	= \boldsymbol{\eta} + \ell_{\boldsymbol{\eta}, \mathbf{r}}[A^*] \mathbf{r} \in \mathbb{R}^M
\end{align*}
for all $\boldsymbol{\lambda} \in \mathcal{S}_+^{M-1}$, where $A^* \in \mathbb{Y}^*_{\boldsymbol{\eta}}$ is the Pareto front surface of interest and $\mathbf{r}\in \mathcal{S}_+^{M-1}$ denotes the updated positive unit vector with $r^{(m)} \propto (u^{(m)} - l^{(m)}) \lambda^{(m)}$ for objectives $m=1,\dots,M$.

To visualise the one-dimensional projections of the Pareto front surface, we propose using a parallel coordinates plot, which is a common visualisation tool for high-dimensional Pareto fronts \citep{tusar2015itec}. The novelty of our work is that we can also view uncertainty information along each projection as well. For example, in our application we computed and visualised the empirical quantiles and the sample mean obtained using a finite set of samples $\{A^*(\boldsymbol{\omega}_n) \in \mathbb{Y}^*_{\boldsymbol{\eta}}\}_{n=1}^N$. Notably, this overall application is very lightweight to run because all of these statistics can computed and updated very quickly and cheaply on the fly. 

\paragraph{Two-dimensional projection} Our two-dimensional dashboard application works in very much the same way as our one-dimensional application described above. The only key difference now is the introduction of two additional drop-down lists, which are used to determine the indices of the set $I = \{i_1, i_2\}$. The corresponding fixed vector $\mathbf{v} \in \mathcal{V}_+^{M-2}$ is then determined by the remaining (normalised) weight sliders. 

To visualise the two-dimensional slices of the Pareto front surface distribution, we considered using a regular two-dimensional line plot. In our application, we included visuals on the empirical quantiles and the sample mean of the projected two-dimensional Pareto front surface. Naturally, other useful information could be included as well. For example, to address \cref{rem:non_constant}, it might be beneficial to also include a parallel coordinates plot of the fixed vector as well---similar to the one used in \cref{fig:projection_one}, but only for the fixed components. 

\paragraph{Higher dimensional projections} The one-dimensional slices gave us information about the marginal performance for each objective. Whilst the two-dimensional slices gave us information about the correlation between any two objectives. Similarly, in order to learn more about the $P$-th order interactions, we need to be able to visualise the $P$-dimensional slices of the Pareto front surface. For example, one might consider adapting one of the methods surveyed by \cite{tusar2015itec} in order to accomplish this. 
\subsection{Uncertainty quantification}
\label{sec:uncertainty_quantification}
The statistical concepts described in \cref{sec:statistics} gives us a tangible and useful way to analyse and quantify the uncertainty surrounding a distribution of random Pareto front surfaces. In this section, we highlight some potential use cases for this machinery in a Bayesian experimental design setting. Formally, we consider the problem of identifying the Pareto front surface associated with some bounded vector-valued black-box objective function $g: \mathbb{X} \rightarrow \mathbb{R}^M$. We suppose that we have executed an experimental design procedure and have observed a collection of potentially noisy data points $\mathcal{D}_T = \{(\mathbf{x}_1, \mathbf{y}_1), \dots, (\mathbf{x}_T, \mathbf{y}_T)\} \subset \mathbb{X} \times \mathbb{R}^M$. We then adopt a standard Bayesian set-up in which we assume a probabilistic prior on the objective function $p(g)$ and a likelihood on the observations $p(\mathcal{D}_T|g)$. These variables can then be used to compute a posterior distribution over the objective function, $p(g|\mathcal{D}_T) \propto p(g)p(\mathcal{D}_T|g)$, and consequently over the Pareto front surface, $Y_{\boldsymbol{\eta}, g}^* := \mathcal{Y}_{\boldsymbol{\eta}}^{\text{int}}[\{g(\mathbf{x})\}_{\mathbf{x} \in \mathbb{X}}]$, as summarised in the following flow diagram:
\begin{equation*}
	p(g), p(\mathcal{D}_T|g)
	\xrightarrow[\text{posterior}]{} p(g|\mathcal{D}_T) 
	\xrightarrow[\text{Pareto front}]{} p(Y_{\boldsymbol{\eta}, g}^*|\mathcal{D}_T).
\end{equation*}
We can then appeal to our earlier work in \cref{sec:statistics} in order to study this resulting Pareto front surface distribution. More precisely, we can associate the random function $f: \mathbb{X} \times \Omega \rightarrow \mathbb{R}^M$, described in \cref{sec:statistics}, with the sampling distribution induced by the latest posterior model: $f(\cdot, \boldsymbol{\omega}) \sim p(g(\cdot)|\mathcal{D}_T)$, where $\boldsymbol{\omega} \in \Omega$ is distributed according to $\mathbb{P}$. We can then analyse and evaluate the posterior Pareto front surface distribution $p(Y_{\boldsymbol{\eta}, g}^*|\mathcal{D}_T)$ by studying the corresponding polar parameterised random variable described in \eqref{eqn:stochastic_pareto_front}. To showcase how this overall routine works in practice, we present an illustrative example in \cref{sec:bayesian_optimisation}, where we visualise the evolution of a Pareto front surface distribution during different runs of Bayesian optimisation. Afterwards, in \cref{sec:input_decision}, we then demonstrate how this distributional information can be used in conjunction with the visualisation techniques described in \cref{sec:visualisation} in order to help us make final decisions. 

\subsubsection{Bayesian optimisation}
\label{sec:bayesian_optimisation}
Bayesian optimisation is a popular strategy for black-box optimisation---see the book by \cite{garnett2023} for a recent overview on this topic. Notably, this experimental design strategy takes advantage of a probabilistic surrogate model in order to determine the best inputs to evaluate. As described above, we can easily take advantage of this probabilistic model in order to compute any Pareto front surface statistic of interest. Practically speaking, we envision that these statistics might be valuable for an active decision maker who is interested in adapting the Bayesian optimisation run for their own purposes. For instance, one might use these statistics in conjunction with the visualisation ideas described in \cref{sec:visualisation} in order to visualise and better understand the evolution of the Pareto front surface distribution. Given this newfound understanding, a keen decision maker might then actively refine and reprogram\footnote{For instance, in the expected hypervolume acquisition criterion \citep{emmerich2006itec}, one might use this visual information in order to update the reference vector $\boldsymbol{\eta} \in \mathbb{R}^M$.} the acquisition procedure in order to target a specific region of interest. Alternatively, earlier work by \cite{binois2015ejoor} suggested that it might also be possible to use some Pareto front surface statistics, such as the Vorob'ev deviation \eqref{eqn:vorobev_deviation}, as a basis for a stopping criteria. 

\begin{figure}
	\includegraphics[width=0.9\linewidth]{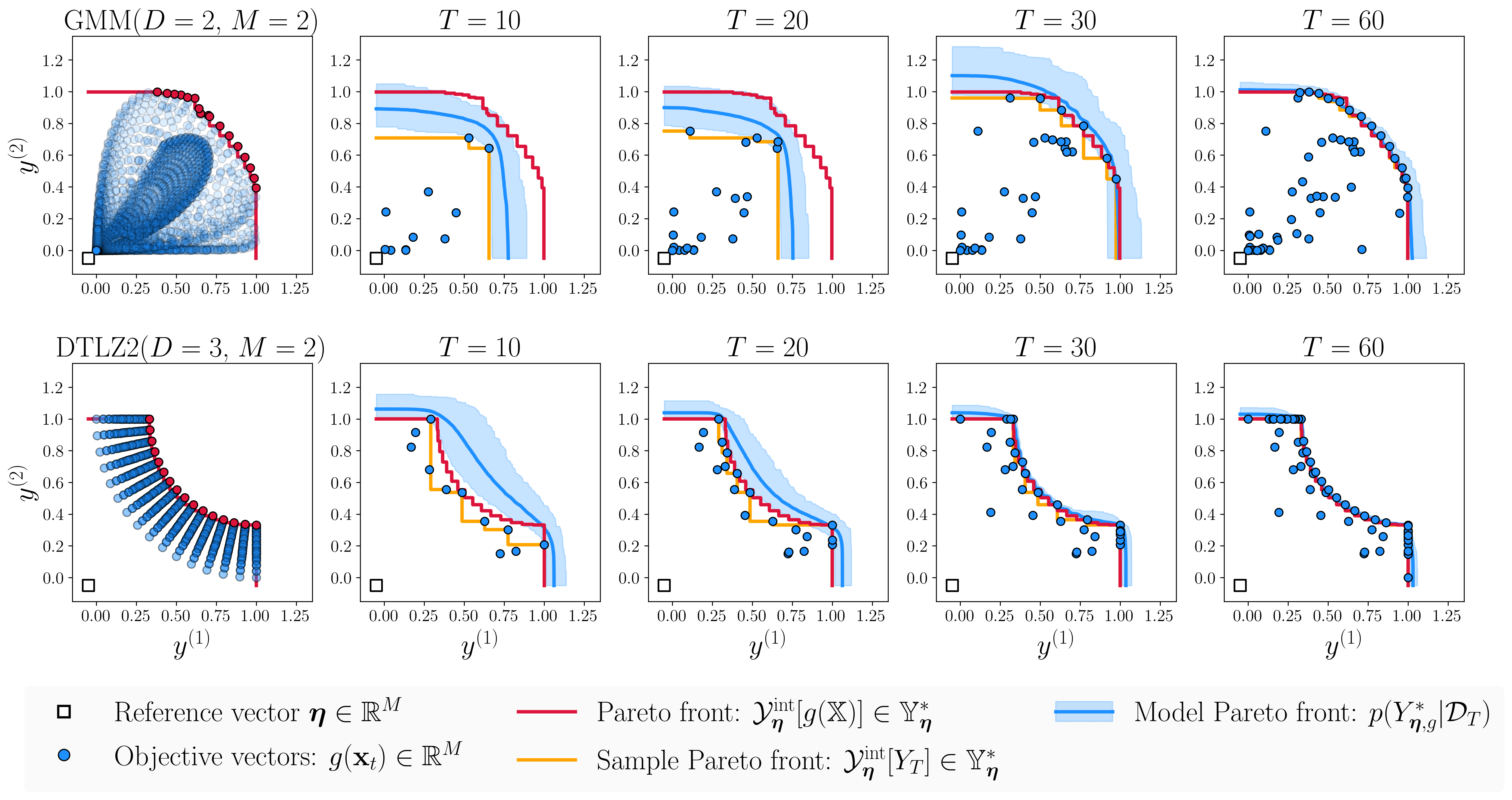}
	\centering
	\caption{An illustration of the change in the Pareto front surface distribution $p(Y_{\boldsymbol{\eta}, g}^*|\mathcal{D}_T)$ when we perform Bayesian optimisation on the normalised GMM and DTLZ2 benchmarks. In the plots, we draw the estimated mean front and highlight the region between the $5\%$ and $95\%$ estimated quantile fronts in blue.}
	\label{fig:bo_example}
\end{figure}

In \cref{fig:bo_example}, we illustrate an example of how the Pareto front surface distribution evolves over time during one run of the Bayesian optimisation algorithm applied on the Gaussian mixture model (GMM) \citep{daulton2022icml} and the DTLZ2 \citep{deb2002p2cecccn} benchmark problem. For the probabilistic model, we adopted a standard Gaussian process prior \citep{rasmussen2006} on our objective function and an independent Gaussian observation likelihood on the function observations. For the acquisition function, we used the expected hypervolume improvement \citep{emmerich2006itec}. For illustrative convenience, we discretised the input space in both of these benchmark problems to have $|\mathbb{X}| = 2^{12}$ points. This latter simplification is only to ensure that we can compute the Pareto front surface of the objective function and the model samples exactly using our polar parameterisation. 

There are a number of key observations that we see from the plots in \cref{fig:bo_example}. Firstly, we see that the Pareto front surface distribution does indeed slowly converge to the actual Pareto front surface when we observe more data points. This is clearly a desirable property and is something that is expected on these benchmark problems. Secondly, we see that the model Pareto front surface always dominates the sample Pareto front surface in these examples. This makes intuitive sense because the sample front considers only a finite number of $|Y_T|=T$ points. In contrast, the model Pareto front surface considers the objective values over the entire input space. Note however that this intuition only holds in our example because the observations were not contaminated with any output noise. Otherwise, when there is observation noise in our data, the sample Pareto front could easily dominate both the actual and model Pareto front surface. For this reason, it is common for practitioners to rely on model-based estimates of the Pareto front surface when the data is noisy. Thirdly, we see that the uncertainty of the Pareto front surface does not necessarily decrease in a monotonic fashion as we add more points. This feature naturally arises in our example because we did not fix the model hyperparameters in our Gaussian process prior. Instead, we followed standard practice and updated these hyperparameters in an online manner by always maximising the latest log marginal likelihood \citep{balandat2020anips}. As a result of this updating step, the overall uncertainty in the Pareto front surface occasionally increased when we incorporate more points---for instance we see this feature when we transition from $T=20$ to $T=30$ in the GMM problem. 

\subsubsection{Input decision}
\label{sec:input_decision}
\begin{figure}
	\includegraphics[width=0.9\linewidth]{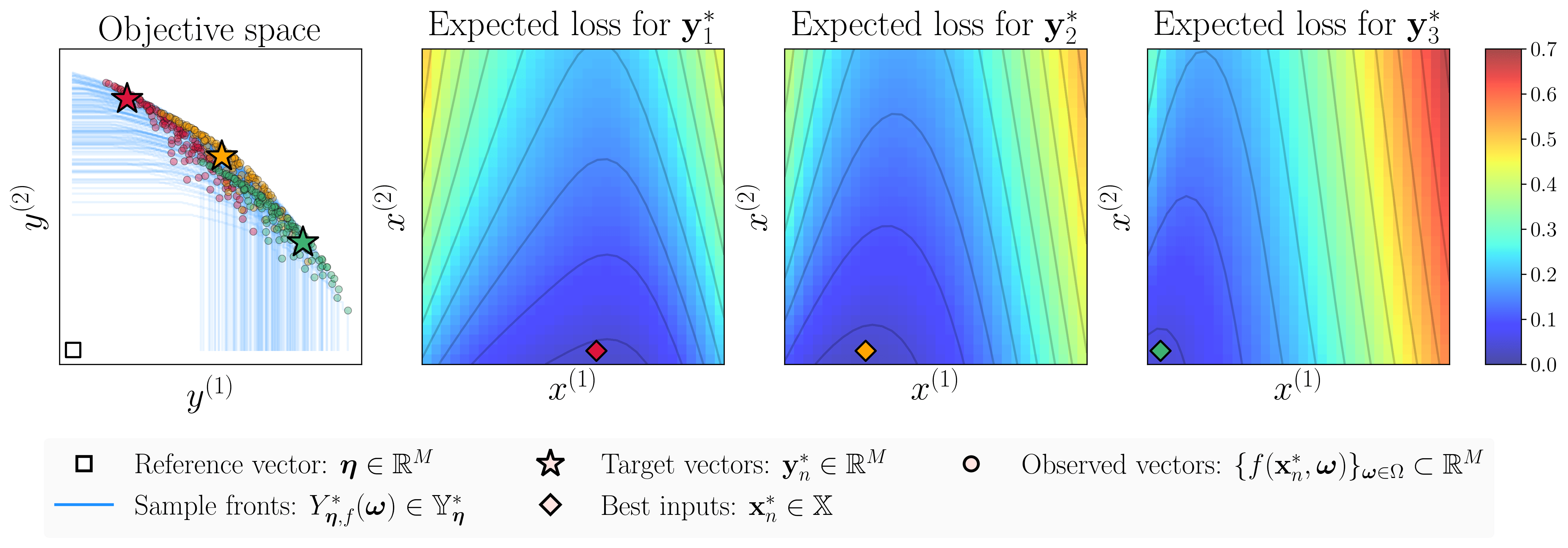}
	\centering
	\caption{An illustration of the best input attained by minimising the length based loss function \eqref{eqn:length_based_loss} with the squared error scoring function $S(x, y) = (x-y)^2$.}
	\label{fig:pareto_front_input_decision}
\end{figure}

The Pareto front surface distribution gives us information on how the optimal set of objective vectors are distributed. This valuable information can be used by the decision maker in order to guide them in their downstream decision making. For instance, this information can help them decide which feasible objective vectors $\mathbf{y}^* \in \mathbb{R}^M$ are the most desirable. On a practical level, this post-selection procedure is typically handled with the help of an interactive application which allows the decision maker to visualise and navigate the Pareto front surface. Notably, we envision that the ideas and tools that we introduced earlier in \cref{sec:visualisation} could be used in this interactive decision making procedure. Consequently, once the desirable vectors have been elicited, the decision maker would often be interested in identifying the inputs $\mathbf{x}^* \in \mathbb{X}$, which would most likely lead to these desirable vectors. To solve this problem, we propose adopting a decision-theoretic approach, where we select the best input as the one which minimise some $M$-dimensional loss function $L: \mathbb{R}^M \times \mathbb{R}^M \rightarrow \mathbb{R}$. That is, we propose minimising the expected empirical loss
\begin{equation*}
	\mathbf{x}^* \in \argmin_{\mathbf{x} \in \mathbb{X}} \frac{1}{N}\sum_{n=1}^N L(f(\mathbf{x}, \boldsymbol{\omega}_n), \mathbf{y}^*),
	\label{eqn:loss_function}
\end{equation*}
where $\{\boldsymbol{\omega}_n\}_{n=1}^N \in \Omega$ denotes i.i.d. samples of the random parameter and $\mathbf{y}^* \in \mathbb{R}^M$ denotes the target vector of interest. Clearly there are many potential loss functions that we can choose in practice. Motivated by the work so far, a natural candidate for the loss function would be a frontier loss function \eqref{eqn:frontier_distance_function}:
\begin{equation*}
	L(\mathbf{y}, \mathbf{y}^*) = \mathcal{D}_{\boldsymbol{\eta}, S}
	[
	\mathcal{Y}^{\textnormal{int}}_{\boldsymbol{\eta}}[\{\mathbf{y}\}], 
	\mathcal{Y}^{\textnormal{int}}_{\boldsymbol{\eta}}[\{\mathbf{y}^*\}]
	]
\end{equation*} 
for any two vectors $\mathbf{y}, \mathbf{y}^* \in \mathbb{R}^M$. A weakness of this loss function is that it can be very expensive to estimate and optimise in practice because it requires computing an integral over the space of positive unit vectors $\mathcal{S}_+^{M-1}$. Motivated by \cref{lemma:domination_equivalence}, we propose using a much cheaper simplification where we consider only computing the score along the optimal direction \eqref{eqn:optimal_weight}:
\begin{equation}
	L(\mathbf{y}, \mathbf{y}^*) 
	= S(s_{\boldsymbol{\eta}, \boldsymbol{\lambda}_{\boldsymbol{\eta}}^*(\mathbf{y}^*)}(\mathbf{y}), s_{\boldsymbol{\eta}, \boldsymbol{\lambda}_{\boldsymbol{\eta}}^*(\mathbf{y}^*)}(\mathbf{y}^*)).
	\label{eqn:length_based_loss}
\end{equation}
Geometrically, this loss function scores any vector $\mathbf{y} \in \mathbb{R}^M$ by computing its projected length along the reference line $L_{\boldsymbol{\eta}, \boldsymbol{\lambda}_{\boldsymbol{\eta}}^*(\mathbf{y}^*)} = \{\boldsymbol{\eta} + t  \boldsymbol{\lambda}_{\boldsymbol{\eta}}^*(\mathbf{y}^*): t \in \mathbb{R}\}$ and then comparing it with the desired projected length. We illustrate the efficacy of this strategy for a two-dimensional example in \cref{fig:pareto_front_input_decision} based on the four bar truss optimisation problem \citep{cheng1999eo,tanabe2020asc}. We see clearly in this example that the random vectors $f(\mathbf{x}^*, \boldsymbol{\omega}) \in \mathbb{R}^M$ associated with the best inputs $\mathbf{x}^* \in \mathbb{X}$ are indeed distributed close to the corresponding target vectors $\mathbf{y}^* \in \mathbb{R}^M$.

\begin{remark}
	[Scale sensitivity] Any $M$-dimensional loss function will naturally be sensitive to the scales of the different objectives. When using the length-based loss functions \eqref{eqn:length_based_loss}, we recommend normalising each objective by its range in order for each objective to have a similar influence on the projected length as illustrated in \cref{sec:dashboard_applications}. Evidently we can also take advantage of this sensitivity in order to inflate the importance of some objectives over others. That is, in order to up-weight the importance of an objective, we can increase its range and similarly in order to down-weight the importance of an objective, we can decrease its range.
\end{remark}

\subsection{Extreme value theory}
\label{sec:extreme_value_theory}
Extreme value theory is a well-established branch of statistics which studies the distribution of extreme values---see for instance \cite{embrechts1997,coles2001} or \cite{beirlant2004} for a background on this topic. In this section, we showcase how our polar parameterisation result can be used in order to generalise many existing ideas in this topic to the multivariate setting, where the maximum is defined using the Pareto partial ordering. To the best of our knowledge, the majority of existing work in multivariate extreme value theory has largely focussed on the marginal maximisation setting \citep{barnett1976jrsssg}, where the maximum operation on a set of vectors is defined component-wise. No work has focussed on the setting where the maximum is defined using the Pareto partial ordering. We attribute this lack of interest based on the fact that the Pareto maximum is very challenging to work with in practice. Despite this, there are clear benefits for adopting the latter approach over the former. Most notably, the Pareto approach is much more flexible because it can also accommodate for the scenario where the various component-wise maximums do not all take place simultaneously.

The most notable results from extreme value theory are the Fisher–Tippett–Gnedenko theorem \citep{fisher1928mpcps,gnedenko1943am} and the Balkema–de Haan–Pickands theorem \citep{balkema1974ap,pickands1975as}. The former result is concerned with the asymptotic distribution of the maximum order statistic associated with a collection of independent and identically distributed univariate random variables. In words, it states that the distribution of the maximum, upon proper normalisation, can only converge in distribution to either a Gumbel, Fréchet or Weibull distribution. In contrast, the latter result is concerned with the limiting distribution of the corresponding conditional excess distribution (\cref{def:excess}). Conceptually, this result tells us that the distribution of the tail events, pass some threshold, can be closely approximated by a generalised Pareto distribution. For reference purposes, we recall both of these results in the Appendix---namely, \cref{thm:gev} and \cref{thm:excess}, respectively.

\paragraph{Component-wise maximum} The existing work on multivariate extreme value theory has largely focussed on the marginal maximisation setting where the maximum of a collection of $N$ independent and identically-distributed vectors $Y_1, \dots, Y_N \in \mathbb{R}^M$ are defined component-wise. The traditional goal of interest is then to study the multivariate generalisation of the maximum domain of attraction (MDA) for the resulting multivariate distribution function $F: \mathbb{R}^M \rightarrow \mathbb{R}$,
\begin{equation*}
	F(\mathbf{x}) = \mathbb{P}[\max(\{Y^{(1)}_1, \dots, Y^{(1)}_N\}) \leq x^{(1)}, \dots, \max(\{Y^{(M)}_1, \dots, Y^{(M)}_N\}) \leq x^{(M)}]
\end{equation*}
for $\mathbf{x} \in \mathbb{R}^M$---see \citet[Chapter 8]{beirlant2004} for an overview on this topic. The primary benefit of using this component-wise definition is that we can immediately apply both \cref{thm:gev} and \cref{thm:excess} in order to determine the asymptotic properties of the corresponding marginal distributions. The remaining challenge is then to identify the corresponding dependence structure between the components. This latter problem turns out to be a major hurdle in multivariate extreme value theory because this dependency structure cannot necessarily be described using a finitely parameterised model. Notably, the estimation and analysis of this dependency structure is still an area of active research---see \citet[Chapter 8]{beirlant2004} for an in-depth discussion.

\begin{figure}
	\includegraphics[width=1\linewidth]{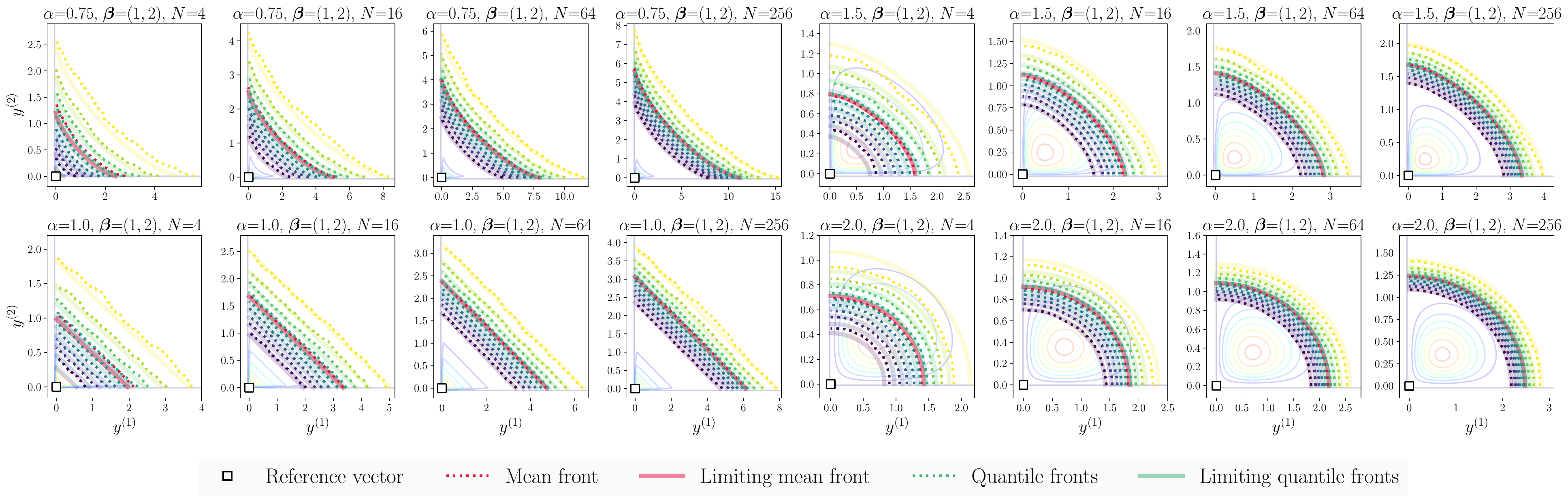}
	\centering
	\caption{An illustration of the Pareto front surface statistics associated with a collection of independent Weibull distributions (\cref{prop:weibull_distribution}). In the dotted lines, we plot the empirical estimates obtained using $N\in\{4, 16, 64, 256\}$ random samples, whereas in the shaded lines we plot the predicted mean and quantiles given by the limiting distribution. As a reference, we also plot the contour of the corresponding bivariate probability density function as well.}
	\label{fig:weibull_pareto_fronts}
\end{figure}

\paragraph{Pareto maximum} One notable outcome of our work is that it manages to connect the definition of the Pareto maximum with the definition of the component-wise maximum. Specifically, our polar parameterisation result tells us that the Pareto maximum can be completely characterised by its projected lengths \eqref{eqn:length_process}, which is defined as an infinite collection of scalarised maximums. To put it more concretely, if we had a collection of identically-distributed vectors $Y_1, \dots, Y_N \in \mathbb{R}^M$, then the corresponding Pareto maximum is governed by the following infinite-dimensional random variable
\begin{equation*}
	\mathcal{L}_{\boldsymbol{\eta}} = \{
	\max(\{ s_{\boldsymbol{\eta}, \boldsymbol{\lambda}}(Y_1), \dots,  s_{\boldsymbol{\eta}, \boldsymbol{\lambda}}(Y_N) \}) \geq 0: 
	\boldsymbol{\lambda} \in \mathcal{S}^{M-1}_+\},
\end{equation*}
for any reference vector $\boldsymbol{\eta} \in \mathbb{R}^M$. As a consequence of this observation, many of the results that are known for the component-wise maximum can now be adapted to the Pareto maximum setting. For example, we can use \cref{thm:gev} to determine the possible asymptotic distributions of the projected length process along each positive direction. Similarly, we can use \cref{thm:excess} in order to approximate the distribution of the tails of the projected length process conditional on the fact that it dominates some specified polar surface. That is, instead of setting just a single threshold value of $u \in \mathbb{R}$, we set the threshold to be a polar surface defined by some set $\mathcal{U} = \{u_{\boldsymbol{\lambda}}\geq 0: \boldsymbol{\lambda} \in \mathcal{S}^{M-1}_+\}$.

As a proof of concept for these general ideas, we present a simple example in the proposition below, where we study the asymptotic distribution of a Pareto front surface constructed using a collection of independent Weibull distributed vectors---the proof of this result is presented in \cref{app:proofs:prop:weibull_distribution}. 
\begin{proposition}
	[Weibull distributed vectors] Let the reference vector be set to zero $\boldsymbol{\eta} = \mathbf{0}_M \in \mathbb{R}^M$ and the vectors $Y_n \in \mathbb{R}^M$ be distributed according to $M > 0$ independent Weibull distributions,
	\begin{equation*}
		Y_n^{(m)} \sim \textnormal{Weibull}(\alpha, \beta^{(m)}),
	\end{equation*}
	with the cumulative distribution function $\mathbb{P}[Y^{(m)} \leq x] = 1 - \exp(-(\beta^{{(m)}}x)^\alpha)$, where $\alpha > 0$ and $\beta^{(m)} > 0$ denotes the corresponding the shape and rate parameter for $m=1,\dots,M$, respectively. Then, upon proper normalisation, the projected lengths of $\mathcal{Y}_{\boldsymbol{\eta}}^{\textnormal{int}}[\{Y_1,\dots, Y_N\}]$, along any positive direction $\boldsymbol{\lambda} \in \mathcal{S}^{M-1}_+$, converges to a standard Gumbel distribution:
	\begin{equation*}
		\lim_{N \rightarrow \infty} 
		\mathbb{P}\biggl[\frac{
			\ell_{\boldsymbol{\eta}, \boldsymbol{\lambda}}[\mathcal{Y}_{\boldsymbol{\eta}}^{\textnormal{int}}[\{Y_1,\dots, Y_N\}]]
			- b_{\boldsymbol{\eta}, \boldsymbol{\lambda}, N}}{a_{\boldsymbol{\eta}, \boldsymbol{\lambda}, N}}  \leq x \biggr]
		= \exp(-\exp(-x))
	\end{equation*}
	where $a_{\boldsymbol{\eta}, \boldsymbol{\lambda}, N} = \log(N)^{1/\alpha - 1} / (\alpha k_{\boldsymbol{\lambda}})$, $b_{\boldsymbol{\eta}, \boldsymbol{\lambda}, N} = \log(N)^{1/\alpha} / k_{\boldsymbol{\lambda}}$ and $k_{\boldsymbol{\lambda}} = (\sum_{m=1}^M (\beta^{{(m)}} \lambda^{(m)})^\alpha )^{1/\alpha}$.
	\label{prop:weibull_distribution}
\end{proposition}
To illustrate this convergence, we present some visual two-dimensional examples in \cref{fig:weibull_pareto_fronts}. In these plots, we fixed\footnote{The rate parameter only controls the relative scaling of each objective and therefore it does not play a major role in the assessment of the empirical result.} the rate parameter $\beta^{(m)} > 0$ and varied the choice of shape parameter $\alpha > 0$. On the whole, we see that the empirical and limiting distributions are already quite similar even at small sample sizes such as $N=64$. Conceptually, we see that varying the shape parameter $\alpha > 0$ amounts to changing from a concave Pareto front surface distribution when $\alpha\in(0,1)$ to a convex Pareto front surface distribution when $\alpha>1$. 
\begin{figure}
	\includegraphics[width=0.9\linewidth]{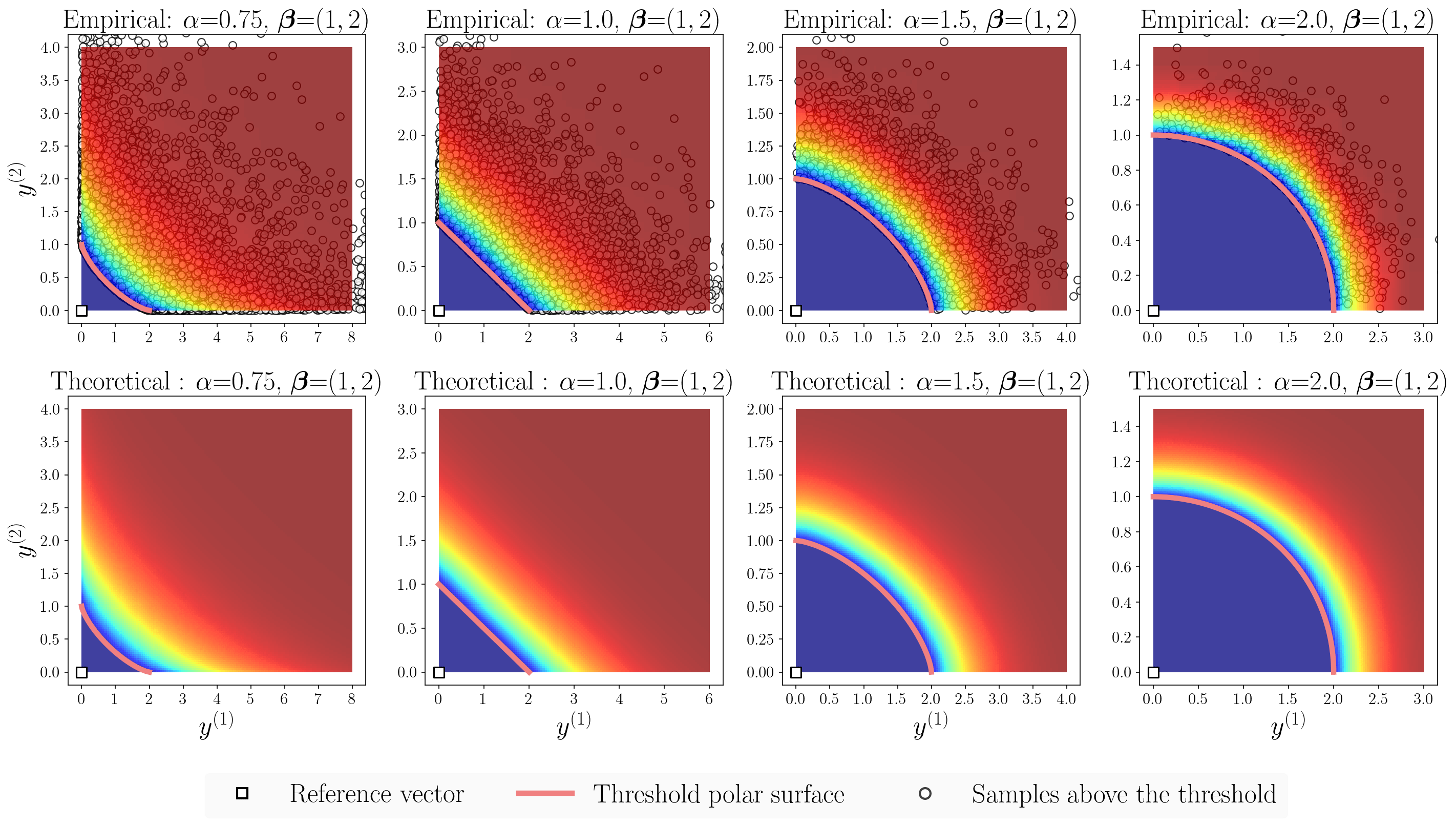}
	\centering
	\caption{A comparison between the empirical and theoretical approximation of the conditional excess probabilities \eqref{eqn:polar_excess_probabilities} associated with the two-dimensional Pareto front surface distributions described in \cref{fig:weibull_pareto_fronts}. To set the threshold polar surface, we used an upper quantile of the original Pareto front surface distribution.} 
	\label{fig:weibull_pareto_fronts_excess}
\end{figure}

\cref{prop:weibull_distribution} can also be used in conjunction with \cref{thm:excess}. In words, this result tells us that the tails of the corresponding projected length process are approximately distributed according to an exponential distribution. To demonstrate this property, we present an illustrative example in \cref{fig:weibull_pareto_fronts_excess}, where we study the conditional excess probabilities associated with the $M=2$ dimensional Pareto front surface distributions described in \cref{fig:weibull_pareto_fronts}. That is, we plot the probabilities
\begin{equation}
	F_{\mathcal{U}}(\mathbf{z}) := \mathbb{P}[s_{\boldsymbol{\eta}, \boldsymbol{\lambda}_{\boldsymbol{\eta}}^*(\mathbf{z})}(Y) - u_{\boldsymbol{\lambda}_{\boldsymbol{\eta}}^*(\mathbf{z})}
	\leq s_{\boldsymbol{\eta}, \boldsymbol{\lambda}_{\boldsymbol{\eta}}^*(\mathbf{z})}(\mathbf{z})| 
	s_{\boldsymbol{\eta}, \boldsymbol{\lambda}_{\boldsymbol{\eta}}^*(\mathbf{z})}(Y) > u_{\boldsymbol{\lambda}_{\boldsymbol{\eta}}^*(\mathbf{z})}]
	\label{eqn:polar_excess_probabilities}
\end{equation}
for any $\mathbf{z} \in \mathbb{D}_{\succsucc}[\{\boldsymbol{\eta}\}]$, where $\boldsymbol{\eta} = \mathbf{0}_M$ is the reference vector, $Y \in \mathbb{R}^M$ is the random vector of interest and $\mathcal{U} = \{u_{\boldsymbol{\lambda}}\geq 0: \boldsymbol{\lambda} \in \mathcal{S}^{M-1}_+\}$ is the projected length process which defines the threshold polar surface. For convenience, in these examples, we set the threshold polar surface to be one of upper quantiles. Nevertheless, we note that any polar surface, whose projected length process is sufficiently long, can be used in practice. Overall we see that these empirical estimates are indeed close to their theoretical approximate values.
\subsection{Air pollution example}
\label{sec:air_pollution}

\begin{figure}
	\includegraphics[width=0.9\linewidth]{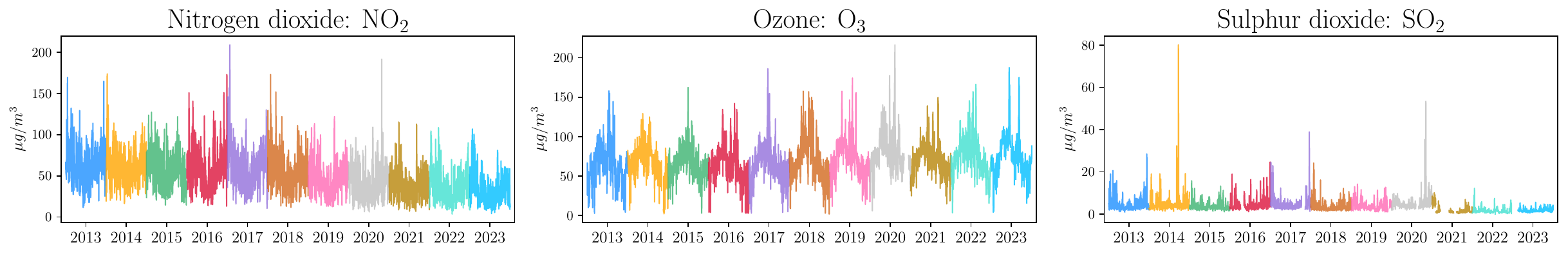}
	\centering
	\caption{An illustration of the daily maximum registered pollutants at the North Kensington monitoring station (UKA00253) over the period of 2013 to 2023.} 
\label{fig:air_pollution_time_series}
\end{figure}
\begin{figure}
	\includegraphics[width=0.78\linewidth]{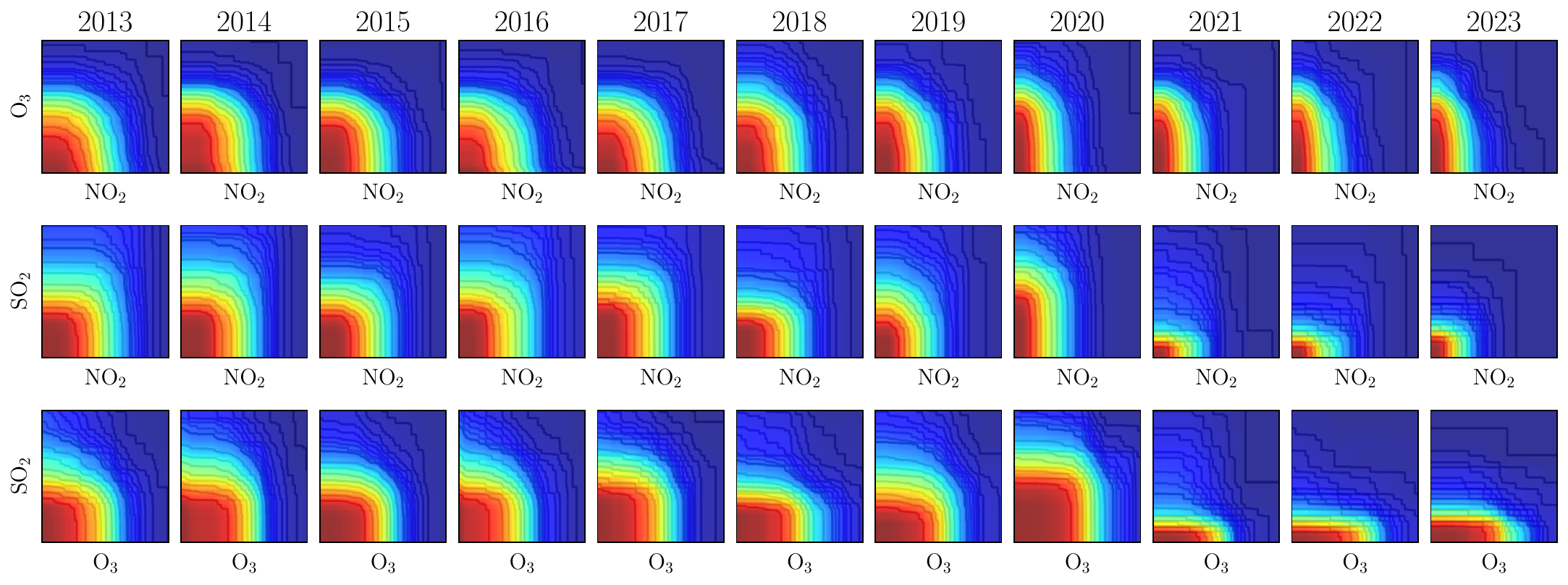}
	\centering
	\caption{An illustration of the change in the domination probabilities for the pairwise Pareto front surfaces associated with the air pollution data (\cref{fig:air_pollution_time_series}).} 
	\label{fig:air_pollution_pareto_fronts}
\end{figure}

All of the ideas that we have presented so far can also accommodate for dynamic problems, where the Pareto front surface distribution can vary in time. As an illustrative example, we will now consider how these ideas can be used on a real-world time series data set. Precisely, we study the air pollution data obtained at the North Kensington (UKA00253) air monitoring station\footnote{We have curated this data from the UK-AIR website (\url{https://uk-air.defra.gov.uk/}), which is a publicly accessible domain that is ran by the Department for Environment, Food \& Rural Affairs (DEFRA) in the United Kingdom.} in west London. 

\paragraph{Data cleaning} The air monitoring station at North Kensington measures many different pollutants every day at different frequencies. We focus our attention on the measurements on the three key pollutants in this area: Nitrogen dioxide $(\text{NO}_2)$, Ozone $(\text{O}_3)$ and Sulphur dioxide $(\text{SO}_2)$. The Nitrogen dioxide and Ozone are measured at a rate of once every $60$ minutes, whilst the Sulphur dioxide is measured at a rate of once every $15$ minutes. In our work, we use the daily maximum of the measured observations as our statistic of interest. Sometimes there are periods where some subset of the measurement devices are offline and therefore no readings can be made. The daily maximums are then computed by using the partial observations that were recorded. On the days where no observations were made at all, we omit the day entirely. In \cref{fig:air_pollution_time_series}, we plot the resulting time series of these daily maximums over the period of 2013 to 2023.

\begin{figure}
	\includegraphics[width=0.78\linewidth]{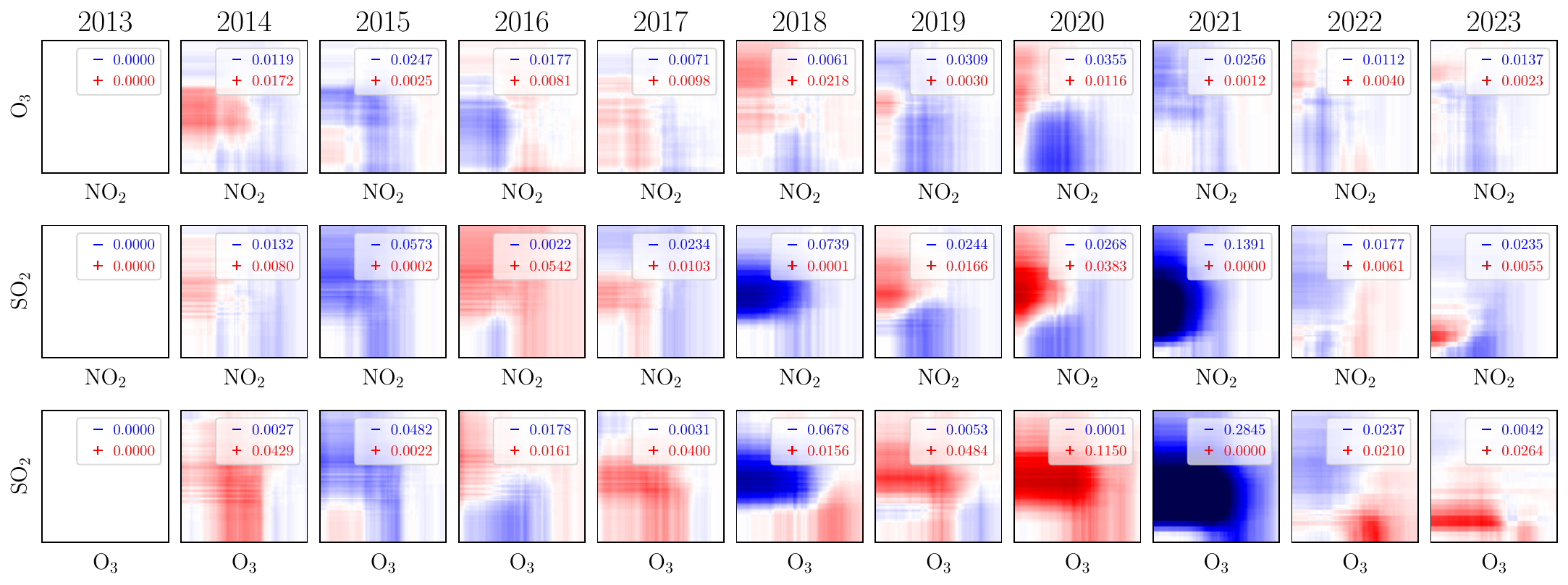}
	\centering
	\caption{An illustration of the signed yearly changes in the domination probabilities for the pairwise Pareto fronts described in \cref{fig:air_pollution_pareto_fronts}. The average yearly negative and positive changes are depicted in the legends. The total average yearly change can be computed by summing the absolute values of these signed changes.}
	\label{fig:air_pollution_deviations}
\end{figure}

\paragraph{Pareto front surface distributions} In this working example, we assume that the goal of interest is to assess and better understand the change in the daily worst-case Pareto front surface of pollution in the North Kensington area over the last decade. That is, we want to see the change in the Pareto fronts of the daily maximum pollutants over the different years. Practically speaking, we envision that this information might be useful for policy makers who are interested in assessing the effectiveness of new pollution reducing initiatives in London such as the Ultra Low Emission Zone (ULEZ) or the lowering of the local speed limits. In \cref{fig:air_pollution_pareto_fronts}, we take advantage of some of our Pareto front machinery in order to compute and plot the corresponding domination probabilities for the different pairwise Pareto front surfaces. We see clearly that there is a general positive trend after 2020, where the Pareto front has been progressing downwards. This indicates a general reduction in the daily maximum pollution. Evidently this time period also aligns with many events such as: the Coronovirus pandemic, the introduction of ULEZ and the lowering of local speed limits. Cumulatively, all of these effects seem to have had a noticeably positive impact on reducing the daily maximum detected Nitrogen dioxide and Sulphur dioxide in North Kensington. Nevertheless, the maximum amount of Ozone is still at a comparative level to the earlier years. To further illustrate the reduction of these domination probabilities more explicitly, we plot and calculate the corresponding average yearly signed changes in \cref{fig:air_pollution_deviations}. That is, we compare every year's Pareto front surface distribution with the previous year and see where the relative positive and negative changes occur. On the whole, we see that there is indeed a quantifiable reduction happening over the later years as these new initiatives were being introduced.

\section{Conclusion}
\label{sec:conclusion}
In this work we presented a novel result which showed that any Pareto front surface can be explicitly parameterised using polar coordinates. We exploited this polar parameterisation result in order to define and compute statistics for random Pareto front surfaces. We then demonstrated the usefulness of these new probabilistic ideas on some interesting applications such as: visualisation, uncertainty quantification and extreme value theory. In the following paragraphs, we highlight some interesting directions for future research. 

\paragraph{Other interpolation schemes} Throughout this work, we have focussed our attention on the interpolated Pareto front (\cref{def:pareto_front_surface}), which was defined using the weak Pareto partial ordering. This type of interpolation is sensible because it is consistent with the weak Pareto partial ordering on sets (\cref{def:set_domination}):  $\mathcal{Y}_{\boldsymbol{\eta}}^{\text{int}}[A] \preceq \mathcal{Y}_{\boldsymbol{\eta}}^{\text{int}}[B]$ if $A \subseteq B \subseteq \mathbb{D}_{\succsucc}[\{\boldsymbol{\eta}\}]$. In practice, one might also consider using alternate interpolation (or approximation) schemes such as the ones based on Delaunay triangulation \citep{hartikainen2012coa}, sandwiching \citep{bokrantz2013ijc} or hyperboxing \citep{dachert2020a,eichfelder2022jgo}. Notably these other schemes might be beneficial when we are interested in approximating a Pareto front of a continuous set $Y \subset \mathbb{R}^M$ using a finite subset $\hat{Y} \subseteq Y$. In this setting, the interpolated weak Pareto front of $\hat{Y}$ would always give us a lower bound estimate of the actual front. In contrast, a different interpolation scheme might be able to give us a more accurate approximation of the target front, with respect to some frontier loss (\cref{sec:loss}). Naturally, many of the results which we have shown here continues to hold even when we change the interpolation scheme. The only notable difference is that the corresponding projected length function $\ell_{\boldsymbol{\eta}, \boldsymbol{\lambda}}$, which arises in the polar parameterisation (\cref{thm:polar_parameterisation}), might no longer be written in terms of the length scalarisation function \eqref{eqn:length_scalarisation}. Instead, the new projected length function would likely depend on a different scalarisation function---one which accurately caters for the interpolation scheme at hand. We leave the analysis of these other interpolation schemes for future research. 

\paragraph{Modelling the Pareto front surface} Our polar parameterisation result implies that any Pareto front surface is completely characterised by its projected length function. \cref{prop:pareto_front_conditions} gives two necessary and sufficient conditions that a length function must satisfy in order for it to induce a valid Pareto front surface. An interesting direction for future work would be focussed on learning this projected length function directly from observational data. Note that the first constraint \eqref{eqn:pareto_condition_1}, positivity, can be easily handled by just considering the logarithm of the length function instead. Whilst the second constraint \eqref{eqn:pareto_condition_2}, the maximum ratio condition, is much less trivial to worth with. Specifically, by taking the logarithm and rearranging the terms, we see that this latter condition is very similar to a Lipschitz type constraint, which can be a difficult constraint to enforce in practice \citep{virmaux2018anips}.


\section*{Acknowledgements}
Ben Tu was supported by the EPSRC StatML CDT programme EP/S023151/1 and BASF SE, Ludwigshafen am Rhein. Nikolas Kantas was partially funded by JPMorgan Chase \& Co. under J.P. Morgan A.I. Faculty Research Awards 2021.
\newpage
\appendix
\section{Proofs}
\label{app:proofs}
We now prove all the results discussed in the main paper. Some of these proofs rely on the monotonicity of the length scalarisation function \eqref{eqn:length_scalarisation}, which we define below. The proof of these monotonicity results follow immediately from the definition of the length scalarisation function for a fixed reference vector $\boldsymbol{\eta} \in \mathbb{R}^M$.

\begin{lemma}
	[Monotonically increasing] The length scalarisation function is a monotonically increasing function over the whole of $\mathbb{R}^M$, that is $\mathbf{y} \succeq \mathbf{y}' \implies s_{\boldsymbol{\eta}, \boldsymbol{\lambda}}(\mathbf{y}) \geq s_{\boldsymbol{\eta}, \boldsymbol{\lambda}}(\mathbf{y}')$ for all vectors $\mathbf{y}, \mathbf{y}' \in \mathbb{R}^M$ and any positive unit vector $\boldsymbol{\lambda} \in \mathcal{S}_+^{M-1}$. 
	\label{lemma:monotonicity}
\end{lemma}

\begin{lemma}
	[Strongly monotonically increasing] The length scalarisation function is a strongly monotonically increasing function over the truncated space, that is $\mathbf{y} \succsucc \mathbf{y}' \implies s_{\boldsymbol{\eta}, \boldsymbol{\lambda}}(\mathbf{y}) > s_{\boldsymbol{\eta}, \boldsymbol{\lambda}}(\mathbf{y}')$ for all vectors $\mathbf{y}, \mathbf{y}' \in \mathbb{D}_{\succsucc}[\{\boldsymbol{\eta}\}]$ and any positive unit vector $\boldsymbol{\lambda} \in \mathcal{S}_+^{M-1}$.
	\label{lemma:strong_monotonicity}
\end{lemma}
\subsection{Proof of \cref{thm:polar_parameterisation}}
\label{app:proofs:thm:polar_parameterisation}
Consider a bounded set of vectors $A \subset \mathbb{R}^M$ and reference vector $\boldsymbol{\eta} \in \mathbb{R}^M$ which admits a non-empty Pareto front surface $\mathcal{Y}_{\boldsymbol{\eta}}^{\textnormal{int}}[A] \neq \emptyset$. Let
\begin{equation*}
	A^* = \{ \boldsymbol{\eta} + \sup_{\mathbf{y} \in A} s_{\boldsymbol{\eta}, \boldsymbol{\lambda}}(\mathbf{y}) \boldsymbol{\lambda} \in \mathbb{R}^M: \boldsymbol{\lambda} \in \mathcal{S}_+^{M-1}\} \in \mathbb{L}_{\boldsymbol{\eta}}.
\end{equation*}
We will first show that this polar surface contains the Pareto front surface: $\mathcal{Y}_{\boldsymbol{\eta}}^{\textnormal{int}}[A] \subseteq A^*$. To prove this result, we take advantage of the well-known fact that every Pareto optimal point $\mathbf{y}^* \in \mathcal{Y}_{\boldsymbol{\eta}}^{\textnormal{int}}[A]$ can be obtained by solving the following optimisation problem \cite[Part 2, Theorem 3.4.5]{miettinen1998}:
\begin{equation}
	\mathbf{y}^* \in \argmax_{\mathbf{y} \in \mathcal{Y}_{\boldsymbol{\eta}}^{\textnormal{int}}[A]} s^{\textnormal{Len}}_{(\boldsymbol{\eta}, \boldsymbol{\lambda}^*_{\boldsymbol{\eta}}(\mathbf{y}^*))}(\mathbf{y})
	= \argmax_{\mathbf{y} \in \mathcal{Y}_{\boldsymbol{\eta}}^{\textnormal{int}}[A]} \min_{m=1,\dots,M}\frac{y^{(m)} - \eta^{(m)}}{(y^*)^{(m)} - \eta^{(m)}}.
	\label{eqn:chebyshev_implication}
\end{equation}
From the monotonicity of the length scalarisation function (\cref{lemma:monotonicity}), we have that
\begin{equation*}
	\max_{\mathbf{y} \in \mathcal{Y}_{\boldsymbol{\eta}}^{\textnormal{int}}[A]} s_{\boldsymbol{\eta}, \boldsymbol{\lambda}}(\mathbf{y})
	= \sup_{\mathbf{y} \in \mathbb{D}_{\preceq,\boldsymbol{\eta}}[A]} s_{\boldsymbol{\eta}, \boldsymbol{\lambda}}(\mathbf{y})
	= \sup_{\mathbf{y} \in A} s_{\boldsymbol{\eta}, \boldsymbol{\lambda}}(\mathbf{y})
\end{equation*}
for any $\boldsymbol{\lambda} \in \mathcal{S}_+^{M-1}$. Combining this result with \eqref{eqn:chebyshev_implication}, we find that
\begin{align*}
	\mathbf{y}^* 
	&= \mathcal{T}^{-1}_{\boldsymbol{\eta}}((\boldsymbol{\lambda}^*_{\boldsymbol{\eta}}(\mathbf{y}^*), s_{\boldsymbol{\eta}, \boldsymbol{\lambda}_{\boldsymbol{\eta}}^*(\mathbf{y}^*)}(\mathbf{y}^*)))
	= \boldsymbol{\eta} + \sup_{\mathbf{y} \in A} s_{\boldsymbol{\eta}, \boldsymbol{\lambda}_{\boldsymbol{\eta}}^*(\mathbf{y}^*)}(\mathbf{y}) \boldsymbol{\lambda}^*_{\boldsymbol{\eta}}(\mathbf{y}^*) \in A^*
\end{align*}
for all $\mathbf{y}^* \in \mathcal{Y}_{\boldsymbol{\eta}}^{\textnormal{int}}[A]$, which implies the desired result: $\mathcal{Y}_{\boldsymbol{\eta}}^{\textnormal{int}}[A] \subseteq A^*$. We will now show by a simple proof by contradiction that the other containment also holds: $\mathcal{Y}_{\boldsymbol{\eta}}^{\textnormal{int}}[A] \supseteq A^*$. Suppose for a contradiction that there exists a vector $\mathbf{a}^* \in A^*$ which is not Pareto optimal, that is $\mathbf{a}^* \notin \mathcal{Y}_{\boldsymbol{\eta}}^{\textnormal{int}}[A]$. Then by the definition of weak Pareto optimality (\cref{def:pareto_front_surface}), there must exist an element $\mathbf{y}^* \in A$ which strongly dominates it. As the length scalarisation function is strongly monotonically increasing over the truncated space (\cref{lemma:strong_monotonicity}), this implies that $s_{\boldsymbol{\eta}, \boldsymbol{\lambda}}(\mathbf{y}^*) > s_{\boldsymbol{\eta}, \boldsymbol{\lambda}}(\mathbf{a}^*)$ for all $\boldsymbol{\lambda} \in \mathcal{S}_+^{M-1}$. This is a contradiction because $\mathbf{a}^* \in A^*$ and therefore it must achieve the largest projected length for at least one positive direction $\boldsymbol{\lambda}' \in \mathcal{S}_+^{M-1}$, that is $s_{\boldsymbol{\eta}, \boldsymbol{\lambda}'}(\mathbf{y}^*) \leq s_{\boldsymbol{\eta}, \boldsymbol{\lambda}'}(\mathbf{a}^*)$ for all $\mathbf{y}^* \in A$.

\begin{flushright}
	$\blacksquare$
\end{flushright}
\subsection{Proof of \cref{lemma:domination_equivalence}}
\label{app:proofs:lemma:domination_equivalence}
Consider any Pareto front surface $A^* \in \mathbb{Y}_{\boldsymbol{\eta}}^*$ and any vector $\mathbf{y} \in \mathbb{D}_{\succsucc}[\{\boldsymbol{\eta}\}]$. We will start by proving the first statement. For the right implication, suppose that $\mathbf{y} \in \mathbb{D}_{\preceq, \boldsymbol{\eta}}[A^*]$. Let $\mathbf{a} \in A^*$ be any vector such that $\mathbf{a} \succeq \mathbf{y}$. By the monotonicity of the length scalarisation function (\cref{lemma:monotonicity}), we have that $s_{\boldsymbol{\eta}, \boldsymbol{\lambda}}(\mathbf{y})
\leq s_{\boldsymbol{\eta}, \boldsymbol{\lambda}}(\mathbf{a}) \leq \ell_{\boldsymbol{\eta}, \boldsymbol{\lambda}} [A^*]$ for all positive directions $\boldsymbol{\lambda} \in \mathcal{S}_+^{M-1}$ and especially for $\boldsymbol{\lambda}_{\boldsymbol{\eta}}^*(\mathbf{y}) \in \mathcal{S}_+^{M-1}$. Now we will prove the left implication. Suppose that $s_{\boldsymbol{\eta}, \boldsymbol{\lambda}_{\boldsymbol{\eta}}^*(\mathbf{y})}(\mathbf{y})
\leq \ell_{\boldsymbol{\eta}, \boldsymbol{\lambda}_{\boldsymbol{\eta}}^*(\mathbf{y})} [A^*]$, then this implies the existence of a vector $\mathbf{a} \in A^*$ such that
\begin{equation*}
	1 \leq 
	\frac{\ell_{\boldsymbol{\eta}, \boldsymbol{\lambda}_{\boldsymbol{\eta}}^*(\mathbf{y})} [A^*]}
	{s_{\boldsymbol{\eta}, \boldsymbol{\lambda}_{\boldsymbol{\eta}}^*(\mathbf{y})}(\mathbf{y})}
	= \frac{s_{\boldsymbol{\eta}, \boldsymbol{\lambda}_{\boldsymbol{\eta}}^*(\mathbf{y})}(\mathbf{a})}
	{s_{\boldsymbol{\eta}, \boldsymbol{\lambda}_{\boldsymbol{\eta}}^*(\mathbf{y})}(\mathbf{y})}
	= \min_{m=1,\dots,M} \frac{a^{(m)} - \eta^{(m)}}{y^{(m)} - \eta^{(m)}},
	\label{eqn:domination_ratio}
\end{equation*}
which implies $\mathbf{a} \succeq \mathbf{y}$ and therefore $\mathbf{y} \in \mathbb{D}_{\preceq, \boldsymbol{\eta}}[A^*]$. The proof of the second statement follows in the same way to the proof of the first statement if we replace some instances of $\preceq$ and $\leq$ with $\precprec$ and $<$, respectively. The third and fourth statements are just the the contrapositive of the first and second statements, respectively, because $\mathbb{D}_{\succsucc, \boldsymbol{\eta}}[A^*] = \mathbb{D}_{\succsucc}[\{\boldsymbol{\eta}\}] \setminus \mathbb{D}_{\preceq, \boldsymbol{\eta}}[A^*]$ and $\mathbb{D}_{\succeq, \boldsymbol{\eta}}[A^*] = \mathbb{D}_{\succsucc}[\{\boldsymbol{\eta}\}] \setminus \mathbb{D}_{\precprec, \boldsymbol{\eta}}[A^*]$.

\begin{flushright}
	$\blacksquare$
\end{flushright}
\subsection{Proof of \cref{prop:pareto_front_conditions}}
\label{app:proofs:prop:pareto_front_conditions}
Consider a polar surface $A \in \mathbb{L}_{\boldsymbol{\eta}}$. We will first prove the sufficiency of these two conditions. Suppose that \ref{eqn:pareto_condition_1} and \ref{eqn:pareto_condition_2} holds, then we will show that $\mathcal{Y}^{\textnormal{int}}_{\boldsymbol{\eta}}[A]$ is non-empty and $\mathcal{Y}^{\textnormal{int}}_{\boldsymbol{\eta}}[A] = A$. Firstly, we see that \ref{eqn:pareto_condition_1} ensures that the set $\mathcal{Y}^{\textnormal{int}}_{\boldsymbol{\eta}}[A]$ is non-empty. Whilst secondly, we see that \ref{eqn:pareto_condition_2} implies that the set $A$ is indeed a Pareto front surface because we cannot find any two vectors in this set such that one strongly dominates the other. In particular, assume for a contradiction that there exists positive directions $\boldsymbol{\lambda}, \boldsymbol{\upsilon} \in \mathcal{S}_+^{M-1}$ such that $\boldsymbol{\eta} + \ell_{\boldsymbol{\eta}, \boldsymbol{\lambda}}[A]\boldsymbol{\lambda} \precprec \boldsymbol{\eta} + \ell_{\boldsymbol{\eta}, \boldsymbol{\upsilon}}[A]\boldsymbol{\upsilon}$. This implies that $\eta^{(m)} + \ell_{\boldsymbol{\eta}, \boldsymbol{\lambda}}[A]\lambda^{(m)} < \eta^{(m)} + \ell_{\boldsymbol{\eta}, \boldsymbol{\upsilon}}[A]\upsilon^{(m)}$, for all objectives $m=1,\dots, M$ and therefore
\begin{equation*}
	\frac{\ell_{\boldsymbol{\eta}, \boldsymbol{\upsilon}}[A] \upsilon^{(m)}}{\ell_{\boldsymbol{\eta}, \boldsymbol{\lambda}}[A] \lambda^{(m)}} > 1
\end{equation*}
for all objectives $m=1,\dots,M$. The above expression contradicts \ref{eqn:pareto_condition_2} because
\begin{equation*}
	\max_{m=1,\dots,M} \frac{\ell_{\boldsymbol{\eta}, \boldsymbol{\lambda}}[A] \lambda^{(m)}}
	{\ell_{\boldsymbol{\eta}, \boldsymbol{\upsilon}}[A] \upsilon^{(m)}}
	 \geq 1,
\end{equation*}
which implies that there exist at least one objective $m$ where the previous statement fails to hold. We will now prove the necessity of these two conditions. Assume that $A$ is a Pareto front surface. Firstly, \ref{eqn:pareto_condition_1} has to hold because the Pareto front surface is a subset of the truncated space $\mathbb{D}_{\succsucc}[\{\boldsymbol{\eta}\}]$. To see that \ref{eqn:pareto_condition_2} has to hold, we reuse the arguments in the proof of \cref{lemma:domination_equivalence}. Consider the vectors $\mathbf{y}_{\boldsymbol{\eta}, \boldsymbol{\lambda}} = \boldsymbol{\eta} + \ell_{\boldsymbol{\eta}, \boldsymbol{\lambda}}[A] \boldsymbol{\lambda} \in A$ and $\mathbf{y}_{\boldsymbol{\eta}, \boldsymbol{\upsilon}} = \boldsymbol{\eta} + \ell_{\boldsymbol{\eta}, \boldsymbol{\upsilon}}[A] \boldsymbol{\upsilon} \in A$ for any two positive directions $\boldsymbol{\lambda}, \boldsymbol{\upsilon} \in \mathcal{S}_+^{M-1}$. As $A$ is Pareto front surface, neither one of these vectors strongly dominate the other. This implies that $\mathbf{y}_{\boldsymbol{\eta}, \boldsymbol{\upsilon}} - \boldsymbol{\eta}$ does not strongly dominate $\mathbf{y}_{\boldsymbol{\eta}, \boldsymbol{\lambda}} - \boldsymbol{\eta}$, which implies \ref{eqn:pareto_condition_2}:
\begin{align*}
	1 \leq \max_{m=1,\dots,M} \frac{y_{\boldsymbol{\eta}, \boldsymbol{\lambda}}^{(m)} - \eta^{(m)}}{y_{\boldsymbol{\eta}, \boldsymbol{\upsilon}}^{(m)} - \eta^{(m)}}
	= \frac{s_{\boldsymbol{\eta}, \boldsymbol{\lambda}}(\mathbf{y}_{\boldsymbol{\eta}, \boldsymbol{\lambda}})}
	{s_{\boldsymbol{\eta}, \boldsymbol{\lambda}}(\mathbf{y}_{\boldsymbol{\eta}, \boldsymbol{\upsilon}})}
	=
	\frac{\ell_{\boldsymbol{\eta}, \boldsymbol{\lambda}}[A]}
	{\ell_{\boldsymbol{\eta}, \boldsymbol{\upsilon}}[A]} \max_{m=1,\dots,M} \frac{\lambda^{(m)}} {\upsilon^{(m)}}
\end{align*}
for any $\boldsymbol{\lambda}, \boldsymbol{\upsilon} \in \mathcal{S}_+^{M-1}$.

\begin{flushright}
	$\blacksquare$
\end{flushright}
\subsection{Proof of \cref{prop:strict_pareto_compliancy}}
\label{app:proofs:prop:strict_pareto_compliancy}
Consider any reference vector $\boldsymbol{\eta} \in \mathbb{R}^M$ and any strictly monotonically increasing transformation $\tau: \mathbb{R}_{\geq 0} \rightarrow \mathbb{R}$. Let $A, B \subset \mathbb{D}_{\succsucc}[\{\boldsymbol{\eta}\}]$ denote two finite sets lying in the truncated space with $A \succ B$. We will show that the length-based utility \eqref{eqn:length_based_utility} satisfies the strict Pareto compliancy property: $U_{\boldsymbol{\eta}, \tau}[A] > U_{\boldsymbol{\eta}, \tau}[B]$. For notational convenience, we begin by defining the set function
\begin{equation*}
	S_{\boldsymbol{\lambda}, \tau}[Y] := \max_{\mathbf{y} \in Y} \tau(s_{\boldsymbol{\eta}, \boldsymbol{\lambda}}(\mathbf{y})).
\end{equation*}
By the monotonicity of the transformation $\tau$, we have that $A \succ B \implies S_{\boldsymbol{\lambda}, \tau}[A] \geq S_{\boldsymbol{\lambda}, \tau}[B]$ for all $\boldsymbol{\lambda} \in \mathcal{S}_+^{M-1}$. By the strict monotonicity of $\tau$ and $\tau^{\text{HV}}$, we have that
\begin{align}
	\begin{split}
		S_{\boldsymbol{\lambda}, \tau^{\text{HV}}}[A] > 
		S_{\boldsymbol{\lambda}, \tau^{\text{HV}}}[B]
		&\iff
		S_{\boldsymbol{\lambda}, \tau}[A] > 
		S_{\boldsymbol{\lambda}, \tau}[B],
		\\
		S_{\boldsymbol{\lambda}, \tau^{\text{HV}}}[A] = 
		S_{\boldsymbol{\lambda}, \tau^{\text{HV}}}[B]
		&\iff
		S_{\boldsymbol{\lambda}, \tau}[A] = 
		S_{\boldsymbol{\lambda}, \tau}[B].
	\end{split}
	\label{eqn:monotonic_condition}
\end{align}
As the hypervolume indicator \eqref{eqn:hypervolume_indicator} is strictly Pareto compliant \citep{zitzler2003itec}, we have that $U_{\boldsymbol{\eta}}^{\text{HV}}[A] > U_{\boldsymbol{\eta}}^{\text{HV}}[B]$, which implies that
\begin{align*}
	0 < U_{\boldsymbol{\eta}}^{\text{HV}}[A] - U_{\boldsymbol{\eta}}^{\text{HV}}[B]
	&= \mathbb{E}_{\boldsymbol{\lambda} \sim \text{Uniform}(\mathcal{S}_+^{M-1})}
	[
	\mathbbm{1}[\boldsymbol{\lambda} \in \Lambda]
	(S_{\boldsymbol{\lambda}, \tau^{\text{HV}}}[A]
	- S_{\boldsymbol{\lambda}, \tau^{\text{HV}}}[B])
	],
\end{align*}
where $\Lambda \subseteq \mathcal{S}_+^{M-1}$ denotes the measurable subset of positive unit vectors where $S_{\boldsymbol{\lambda}, \tau^{\text{HV}}}[A] > S_{\boldsymbol{\lambda}, \tau^{\text{HV}}}[B]$ and $\nu[\Lambda] > 0$. Combining this result with the implications in \eqref{eqn:monotonic_condition}, we obtain the desired result: $U_{\boldsymbol{\eta}, \tau}[A] > U_{\boldsymbol{\eta}, \tau}[B]$.

\begin{flushright}
	$\blacksquare$
\end{flushright}
\subsection{Proof of \cref{prop:expected_front}}
\label{app:proofs:prop:expected_front}
Equipped with \cref{ass:positive_lengths,ass:bounded_lengths}, we will show that the expectation in \eqref{eqn:expectation} satisfies the two conditions in \cref{prop:pareto_front_conditions} and therefore is a valid Pareto front surface. Firstly, \ref{eqn:pareto_condition_1} is satisfied because the lengths are assumed to be positive almost surely, which implies that $\mathbb{E}_{\boldsymbol{\omega}}[\ell_{\boldsymbol{\eta}, \boldsymbol{\lambda}}[Y_{\boldsymbol{\eta}, f}^*(\boldsymbol{\omega})]] > 0$ for all $\boldsymbol{\lambda} \in \mathbb{R}^M$. Secondly, \ref{eqn:pareto_condition_2} is satisfied because $Y_{\boldsymbol{\eta}, f}^*(\boldsymbol{\omega}) \in \mathbb{L}_{\boldsymbol{\eta}}$ is a Pareto front surface almost surely and therefore it satisfies \ref{eqn:pareto_condition_2} almost surely,
\begin{equation*}
	\ell_{\boldsymbol{\eta}, \boldsymbol{\lambda}}[Y_{\boldsymbol{\eta}, f}^*(\boldsymbol{\omega})]
	\max_{m=1,\dots,M} \frac{\lambda^{(m)}} {\upsilon^{(m)}} \geq 
	\ell_{\boldsymbol{\eta}, \boldsymbol{\upsilon}}[Y_{\boldsymbol{\eta}, f}^*(\boldsymbol{\omega})],
\end{equation*}
which implies that the expectation also satisfies \ref{eqn:pareto_condition_2},
\begin{equation*}
	\mathbb{E}_{\boldsymbol{\omega}}[\ell_{\boldsymbol{\eta}, \boldsymbol{\lambda}}[Y_{\boldsymbol{\eta}, f}^*(\boldsymbol{\omega})]]
	\max_{m=1,\dots,M} \frac{\lambda^{(m)}} {\upsilon^{(m)}} \geq 
	\mathbb{E}_{\boldsymbol{\omega}}[\ell_{\boldsymbol{\eta}, \boldsymbol{\upsilon}}[Y_{\boldsymbol{\eta}, f}^*(\boldsymbol{\omega})]],
\end{equation*}
for any two positive directions $\boldsymbol{\lambda}, \boldsymbol{\upsilon} \in \mathcal{S}_+^{M-1}$.

\begin{flushright}
	$\blacksquare$
\end{flushright}
\subsection{Proof of \cref{prop:quantile_front}}
\label{app:proofs:prop:quantile_front}
To prove that the $\alpha$-quantile \eqref{eqn:quantile} is a valid Pareto front surface under the given assumptions, we can show that it satisfies the two conditions in \cref{prop:pareto_front_conditions}. This exercise is simply a repeat of the arguments in \cref{app:proofs:prop:expected_front}, with every instance of the expectation $\mathbb{E}_{\boldsymbol{\omega}}[\cdot]$ being replaced with the $\alpha$-quantile $\mathcal{Q}_{\boldsymbol{\omega}}[\cdot, \alpha]$ instead.

\begin{flushright}
	$\blacksquare$
\end{flushright}
\subsection{Proof of \cref{lemma:projected_slice}}
\label{app:proofs:lemma:projected_slice}
Consider a non-empty set of indices $I \subset [M]$ with $|I|=P$ and a vector $\mathbf{v} \in \mathcal{V}_+^{M-P}$. We have the following equivalence:
\begin{align*}
	\mathcal{P}_{I, \mathbf{v}}[\mathcal{S}^{M-1}_+] 
	&= \{\mathcal{P}_I(\boldsymbol{\lambda}) \in \mathbb{R}^P: \boldsymbol{\lambda} \in \mathcal{S}_+^{M-1} \text{ and }
	\mathcal{P}_{[M] \setminus I}(\boldsymbol{\lambda}) = \mathbf{v}\}
	\\
	&=
	\{\sqrt{1-||\mathbf{v}||^2_{L^2}} \boldsymbol{\lambda} \in \mathbb{R}^P: \boldsymbol{\lambda} \in \mathcal{S}_+^{P-1}\}
	\\
	&= \sqrt{(1-||\mathbf{v}||^2_{L^2})} \odot \mathcal{S}_+^{P-1}.
\end{align*}
The fact that this final set is a $P$-dimensional Pareto front surface with the zero reference vector follows from \cref{eg:scalar_multiplication} and the fact that $\mathcal{S}_+^{P-1} \in \mathbb{Y}_{\mathbf{0}_P}^*$ and $\sqrt{(1-||\mathbf{v}||^2_{L^2})} > 0$.

\begin{flushright}
	$\blacksquare$
\end{flushright}
\subsection{Proof of \cref{prop:projected_pareto_front}}
\label{app:proofs:prop:projected_pareto_front}
Consider a Pareto front surface $A^* \in \mathbb{Y}_{\boldsymbol{\eta}}^*$, any non-empty set of indices $I \subset [M]$ with $|I| = P$, any vector $\mathbf{v} \in \mathcal{V}_+^{M-P}$, we will show that the projected Pareto front surface $\mathcal{P}_{I, \mathbf{v}}[A^*] \in \mathbb{L}_{\mathcal{P}_I(\boldsymbol{\eta})}$ is a $P$-dimensional Pareto front surface with the reference vector $\mathcal{P}_I(\boldsymbol{\eta}) \in \mathbb{R}^P$. To accomplish this, we will show that this set satisfies the two conditions in \cref{prop:pareto_front_conditions}. Firstly, \ref{eqn:pareto_condition_1} is satisfied because the projected lengths of $A^*$ are positive and $\sqrt{1-||\mathbf{v}||^2_{L^2}} > 0$. Secondly, to see that \ref{eqn:pareto_condition_2} is satisfied, we begin by noting that $A^*$ satisfies \ref{eqn:pareto_condition_2} because it is a valid Pareto front surface, which implies
\begin{equation}
	\frac{\ell_{\boldsymbol{\eta}, \phi_I(\mathbf{v}, \boldsymbol{\lambda})}[A^*]}{\ell_{\boldsymbol{\eta}, \phi_I(\mathbf{v}, \boldsymbol{\upsilon})}[A^*]}
	\max_{m=1,\dots,M} \frac{\phi^{(m)}_I(\mathbf{v}, \boldsymbol{\lambda})} {\phi_I^{(m)}(\mathbf{v}, \boldsymbol{\upsilon})} \geq 1
	\label{eqn:projected_maximum_ratio_a}
\end{equation}
for any $\boldsymbol{\lambda}, \boldsymbol{\upsilon} \in \mathcal{S}_+^{P-1}$. As the vectors on the indices of $[M] \setminus I$ are fixed to be $\mathbf{v} \in \mathcal{V}_+^{M-P}$ and $\mathcal{S}_+^{P-1}$ is a Pareto front surface, we see that the maximum ratio on the set of indices $[M]$ is achieved on the subset $I$:
\begin{equation}
	\max_{m=1,\dots,M} \frac{\phi^{(m)}_I(\mathbf{v}, \boldsymbol{\lambda})} {\phi_I^{(m)}(\mathbf{v}, \boldsymbol{\upsilon})}
	=
	\max_{m \in I} \frac{\phi^{(m)}_I(\mathbf{v}, \boldsymbol{\lambda})} {\phi_I^{(m)}(\mathbf{v}, \boldsymbol{\upsilon})}.
	\label{eqn:projected_maximum_ratio_b}
\end{equation}
By substituting \eqref{eqn:projected_maximum_ratio_b} into \eqref{eqn:projected_maximum_ratio_a} and using the definition of the reconstruction function in \eqref{eqn:reconstruction_function}, we see that \ref{eqn:pareto_condition_2} does indeed hold for the projected Pareto front surface:
\begin{equation*}
	\frac{\sqrt{(1-||\mathbf{v}||^2_{L^2})}\ell_{\boldsymbol{\eta}, \phi_I(\mathbf{v}, \boldsymbol{\lambda})}[A^*]}
	{\sqrt{(1-||\mathbf{v}||^2_{L^2})}\ell_{\boldsymbol{\eta}, \phi_I(\mathbf{v}, \boldsymbol{\upsilon})}[A^*]}
	\max_{m = 1,\dots,P} \frac{\lambda^{(m)}}{\upsilon^{(m)}} \geq 1
\end{equation*}
for any $\boldsymbol{\lambda}, \boldsymbol{\upsilon} \in \mathcal{S}_+^{P-1}$.

\begin{flushright}
	$\blacksquare$
\end{flushright}
\subsection{Extreme value theory}
We begin this subsection by first recalling the statement of Fisher–Tippett–Gnedenko theorem \citep{fisher1928mpcps,gnedenko1943am} and the Balkema–de Haan–Pickands theorem \citep{balkema1974ap,pickands1975as} in \cref{thm:gev} and \cref{thm:excess}, respectively. The proof of these two results can be found in standard textbooks; for example, see the proofs presented by \citet[Theorem 1.4.2]{leadbetter1983} and \citet[Theorem 1.6.2]{leadbetter1983}, respectively. Afterwards, we move on to \cref{app:proofs:prop:weibull_distribution}, where we present the proof of \cref{prop:weibull_distribution}.

\begin{definition}
	[Maximum domain of attraction] A cumulative distribution function $F: \mathbb{R} \rightarrow \mathbb{R}$ is in the maximum domain of attraction (MDA) of a cumulative distribution function $H: \mathbb{R} \rightarrow \mathbb{R}$, denoted by $F \in \text{MDA}(H)$, if there exist two sequences of real numbers $a_N>0$ and $b_N \in \mathbb{R}$ such that
	\begin{equation*}
		\lim_{N \rightarrow \infty} \mathbb{P}\biggl[\frac{\max(\{Y_1, \dots, Y_N\}) - b_N}{a_N}  \leq x \biggr] 
		= \lim_{N \rightarrow \infty} F(a_N x + b_N)^N
		=  H(x)
	\end{equation*}
	for $x \in \mathbb{R}$, where $Y, Y_1, \dots, Y_N \in \mathbb{R}$ denotes a collection of independent and identically distributed random samples from the distribution $F$.
\end{definition}

\begin{theorem}
	[Extreme value distribution] \citep{fisher1928mpcps,gnedenko1943am} If the cumulative distribution function $F: \mathbb{R} \rightarrow \mathbb{R}$ is in the MDA of a non-degenerate distribution function $H: \mathbb{R} \rightarrow \mathbb{R}$, that is $F \in \text{MDA}(H)$, then $H(x) = \exp(-(1+\xi (\frac{x-\mu}{\sigma}) )_+^{-1/\xi})$ is a generalised extreme value distribution,
	where $\xi \in \mathbb{R}$, $\mu \in \mathbb{R}$ and $\sigma > 0$, denotes the shape, location and scale parameter, respectively. 
	\label{thm:gev}
\end{theorem}

\begin{definition}
	[Conditional excess distribution] Let $Y \in \mathbb{R}$ denote a random variable with a distribution function $F: \mathbb{R} \rightarrow \mathbb{R}$. The corresponding conditional excess distribution function at some level $u \in \mathbb{R}$ is given by
	\begin{equation*}
		F_{u}(x) := \mathbb{P}[Y - u \leq x| Y > u] = \frac{F(u + x) - F(u)}{1 - F(u)},
	\end{equation*}
	for any $x \in [0, y_{F} - u]$, where $y_{F} := \sup\{x \in \mathbb{R}: F(x) < 1\}$ denotes the finite or infinite right endpoint of the distribution function $F$.
	\label{def:excess}
\end{definition}

\begin{theorem}
	[Generalised Pareto distribution] \citep{balkema1974ap,pickands1975as} If the cumulative distribution function $F: \mathbb{R} \rightarrow \mathbb{R}$ is in the MDA of a generalised extreme value distribution, $F \in \text{MDA}(H_{\xi, \mu, \sigma})$, then there exists a positive, measurable function\footnote{Suppose that $F(x)^N \approx H_{\xi, \mu, \sigma}(x)$ for large $N$, then \citet[Theorem 4.1]{coles2001} suggest that we can set $\beta(u) = \sigma + \xi(u - \mu) > 0$ for sufficiently large $u \in \mathbb{R}$.} $\beta: \mathbb{R} \rightarrow \mathbb{R}_{>0}$ such that the following limit holds:
	\begin{equation*}
		\lim_{u \rightarrow y_{F}} \sup_{x \in [0, y_{F} - u]}|F_u(x) - G_{\xi, \beta(u)}(x)| = 0
	\end{equation*}
	where $G_{\xi, \beta}(x) = 1- (1 + \frac{\xi x}{\beta})_+^{-1/\xi}$ denotes the distribution function of a generalised Pareto distribution with shape parameter $\xi \in \mathbb{R}$ and rate parameter $\beta > 0$.
	\label{thm:excess}
\end{theorem}

\subsubsection{Proof of \cref{prop:weibull_distribution}}
\label{app:proofs:prop:weibull_distribution}
As described in \cref{sec:applications}, the projected lengths along any positive direction $\boldsymbol{\lambda} \in \mathcal{S}^{M-1}_+$ is given by the maximum of the length scalarised values
\begin{equation}
	\ell_{\boldsymbol{\eta}, \boldsymbol{\lambda}}[\mathcal{Y}_{\boldsymbol{\eta}}^{\textnormal{int}}[\{Y_1,\dots, Y_N\}]]
	= \max(\{ s_{\boldsymbol{\eta}, \boldsymbol{\lambda}}(Y_1), \dots,  s_{\boldsymbol{\eta}, \boldsymbol{\lambda}}(Y_N) \}).
	\label{eqn:projected_length_distribution}
\end{equation}
As a Weibull distributed random variable is non-negative and $\boldsymbol{\eta} = \mathbf{0}_M \in \mathbb{R}^M$, we have that $s_{\boldsymbol{\eta}, \boldsymbol{\lambda}}(Y_n) = \min_{m=1,\dots,M} Y^{(m)}/\lambda^{(m)}$ for $n=1,\dots,N$. By a standard calculation, we see that the Weibull distribution is closed under scaling,
\begin{equation*}
	\mathbb{P}\biggl[ \frac{Y^{(m)}}{\lambda^{(m)}} \leq x\biggr] = 1 - \exp(-(\beta^{{(m)}} \lambda^{(m)}x)^\alpha),
\end{equation*}
which implies $Y^{(m)}/\lambda^{(m)} \sim \text{Weibull}(\alpha, \beta^{(m)} \lambda^{(m)})$. Similarly, we can show that the minimum of a collection of independent Weibull distributed random variables is also Weibull distributed:
\begin{align*}
	\mathbb{P}\biggl[ \min_{m=1,\dots,M} \frac{Y^{(m)}}{\lambda^{(m)}} \leq x \biggr] 
	&= 1 - \prod_{m=1}^M \biggl(1 - \mathbb{P}\biggl[ \frac{Y^{(m)}}{\lambda^{(m)}} \leq x\biggr] \biggr)
	= 1 - \exp\biggl(- \sum_{m=1}^M (\beta^{{(m)}} \lambda^{(m)})^\alpha x^\alpha \biggr),
\end{align*}
which implies that $s_{\boldsymbol{\eta}, \boldsymbol{\lambda}}(Y_n) \sim \text{Weibull}(\alpha,  (\sum_{m=1}^M (\beta^{{(m)}} \lambda^{(m)})^\alpha )^{1/\alpha})$ for $n=1,\dots,N$. This implies that the distribution of the projected lengths \eqref{eqn:projected_length_distribution} is equivalent to the distribution of the maximum of a collection of independent and identically distributed Weibull distributions. The final result is then obtained by repeating a standard calculation---confer with \citet[Table 3.4.4]{embrechts1997}.

\begin{flushright}
	$\blacksquare$
\end{flushright}
\newpage
\bibliographystyle{plainnat}
\bibliography{ms}
\end{document}